%% file: main.tex
\documentclass{article}
\usepackage[margin=1in]{geometry}

\usepackage[utf8]{inputenc} % allow utf-8 input
\usepackage[T1]{fontenc}    % use 8-bit T1 fonts
\usepackage[colorlinks, linkcolor=orange, anchorcolor=blue, citecolor=blue]{hyperref}
\usepackage{url}            % simple URL typesetting
\usepackage{booktabs}       % professional-quality tables
\usepackage{amsfonts}       % blackboard math symbols
\usepackage{nicefrac}       % compact symbols for 1/2, etc.
\usepackage{microtype}      % microtypography
\usepackage{xcolor}         % colors
\usepackage{comment}
\usepackage{caption}
\usepackage{subcaption}
\usepackage{amsmath}
\usepackage{amssymb}
\usepackage{mathtools}
\usepackage{amsthm}
\usepackage{smile}
\usepackage{natbib}
\newcommand{\diff}{\mathrm{d}}

\newcommand{\Xyt}{X_{t}^y}

\newcommand{\ttXytb}{\tilde{X}_{t}^{t, \Leftarrow}}
\newcommand{\px}{x_{\parallel}}
\newcommand{\ox}{x_{\perp}}
\newcommand{\subangle}[2]{\angle({#1}, {#2})}

\newcommand{\Clip}{C_{\rm lip}}
\newcommand{\clip}{c_{\rm lip}}
\newcommand{\dtv}{\texttt{d}_{\rm TV}}
\newcommand{\dlabel}{\cD_{\rm label}}
\newcommand{\dunlabel}{\cD_{\rm unlabel}}
\newcommand{\subopt}{\texttt{SubOpt}}
\newcommand{\dshift}{\texttt{DistroShift}}

\newcommand{\re}[1]{{\textcolor{black}{ \rm{#1}}}}

\def\P{{\mathbb P}}

\def\Hcal{{\mathcal H}}

\title{\bf Diffusion Model for Data-Driven Black-Box Optimization\thanks{$\diamond$ denotes equal contribution listed in alphabetical order. $\ddagger$ denotes co-correspondence.}\thanks{This paper is a journal version of \cite{yuan2023reward}.}}
\author{Zihao Li$^{1, \diamond}$ \quad Hui Yuan$^{1, \diamond}$ \quad Kaixuan Huang$^1$ \quad Chengzhuo Ni$^1$ \\ Yinyu Ye$^2$ \quad Minshuo Chen$^{1, \ddagger}$ \quad Mengdi Wang$^{1, \ddagger}$ \\
$^1$Princeton University \quad $^2$Stanford University}
\begin{document}

\maketitle
\input{sections/abstract}
\input{sections/intro}
\input{sections/related_work}
\input{sections/pre.tex}
\input{sections/method.tex}

\input{sections/casestudy.tex}

\input{sections/experiment.tex}

\bibliography{citation, ref}
\bibliographystyle{plainnat}

\appendix
\newpage
\input{appendix/appdx}
\input{appendix/experiment}
\input{appendix/nonparametric}

\end{document}

%% file: sections/abstract.tex
\begin{abstract}
Generative AI has redefined artificial intelligence, enabling the creation of innovative content and customized solutions that drive business practices into a new era of efficiency and creativity. In this paper, we focus on diffusion models, a powerful generative AI technology, and investigate their potential for black-box optimization over complex structured variables. Consider the practical scenario where one wants to optimize some structured design in a high-dimensional space, based on massive unlabeled data (representing design variables) and a small labeled dataset. We study two practical types of labels: 1) noisy measurements of a real-valued reward function and 2) human preference based on pairwise comparisons. The goal is to generate new designs that are near-optimal and preserve the designed latent structures. Our proposed method reformulates the design optimization problem into a conditional sampling problem, which allows us to leverage the power of diffusion models for modeling complex distributions. In particular, we propose a reward-directed conditional diffusion model, to be trained on the mixed data, for sampling a near-optimal solution conditioned on high predicted rewards. Theoretically, we establish sub-optimality error bounds for the generated designs. The sub-optimality gap nearly matches the optimal guarantee in off-policy bandits, demonstrating the efficiency of reward-directed diffusion models for black-box optimization. Moreover, when the data admits a low-dimensional latent subspace structure, our model efficiently generates high-fidelity designs that closely respect the latent structure. We provide empirical experiments validating our model in decision-making and content-creation tasks.
\end{abstract}

%% file: sections/intro.tex
\section{Introduction}\label{sec:intro}

Generative AI has revolutionized the field of artificial intelligence by introducing capabilities that allow for the creation of new, original content — from realistic images and music to sophisticated text and code — paving the way for unprecedented innovations across industries such as entertainment, marketing, and business management \citep{kshetri2023generative, ooi2023potential}. In particular, diffusion models stand out as a powerful and universal generative AI technology that serves as the backbone for Stable Diffusion, DALLE 3, Sora, etc. \citep{ho2020denoising, song2020denoising,rombach2022high,liu2024sora,esser2024scaling}. Diffusion models, known as a family of score-matching generative models,  have demonstrated the state-of-the-art performance in various domains, such as image generation \citep{rombach2022high, ramesh2022hierarchical, balaji2022ediffi} and audio generation, with fascinating potentials in broader domains, including text modeling \citep{austin2021structured, li2022diffusion}, reinforcement learning \citep{janner2022diffuser, ajay2023is, pearce2023imitating, liang2023adaptdiffuser} and protein structure modeling \citep{LeeProtein}.
Diffusion models are trained to sample from the training data's distribution and generate a new sample through a sequential denoising process.
The denoising network (a.k.a. score network) $s(x,t)$ approximates the score function $\nabla \log p_t(x)$  \citep{song2020sliced, song2020score}, and controls the behavior of diffusion models. People can incorporate any control information $c$ as an additional input to the score network $s(x,c, t)$ during the training and inference \citep{ramesh2022hierarchical, zhang2023adding}.

In this paper, we focus on capitalizing diffusion models for \textit{data-driven black-box optimization}, where the goal is to generate new solutions that optimize an unknown objective function. Black-box optimization problems, also known as model-based optimization in machine learning, cover various application domains including computational biology \citep{watson2023novo, guo2023diffusion}, reinforcement learning \citep{ajay2022conditional, levine2020offline, jin2021pessimism}, and management science, e.g., pricing \citep{bu2023offline,homburg2019multichannel,wang2021online} and recommendation systems \citep{pan2020novel,pan2019user}. In these applications, the data is high dimensional and the unknown objective function is nonconvex, posing challenges to search for good solutions. To further complicate matters, plausible solutions, e.g., molecule structures in drug discovery \citep{barrera2016survey}, are often high-dimensional with latent geometric structures -- they lie on a \textit{low-dimensional manifold} \citep{kumar2020model}. This additionally poses a critical requirment for black-box optimization: we need to capture data geometric structures to avoid suggesting unrealistic solutions that deviate severely from the original data domain.

\begin{figure}[htb!]
    \centering
    \includegraphics[width =0.8\textwidth]{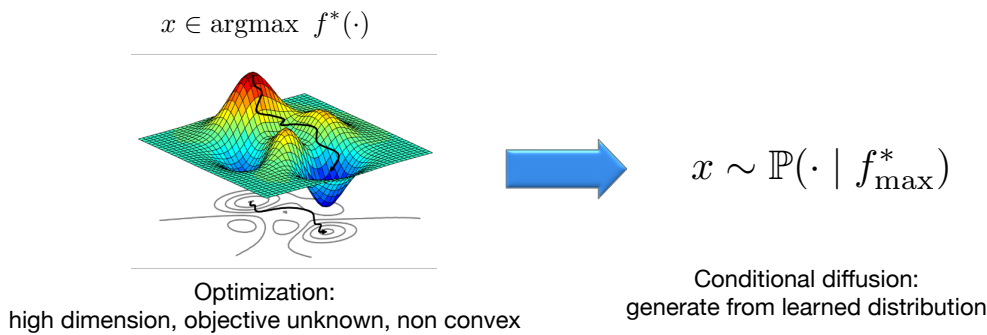} 
    \caption{{\bf Generative AI for black-optimization via conditional sampling.} We convert the problem of black-box optimization into the problem of sampling from a conditional distribution learned from a pre-collected dataset. }
\label{fig:modelbased_opt}
\end{figure}

Data-driven black-box optimization differentiates from conventional optimization problems, as it exposes the knowledge of the objective function only via a pre-collected dataset \citep{kumar2020model, brookes2019conditioning,fu2021offline}, while additional interactions with the objective function are prohibitive. This excludes the possibility of sequentially searching for good solutions, e.g., gradient ascent, aided by feedback from the objective function at each iteration. Instead, we reformulate the data-driven black-box optimization as sampling from a conditional distribution. This novel perspective is demonstrated in Figure~\ref{fig:modelbased_opt}, and the objective function value is the conditioning in the conditional distribution. We learn such a conditional distribution from the pre-collected dataset using diffusion models, and optimal solutions are then generated by conditioning on large objective values. Meanwhile, the generated solutions are expected to respect the data latent structures, since the data distribution encodes the corresponding data geometry.
 
\begin{figure}[htb!]
    \centering
    \includegraphics[width = 1\textwidth]{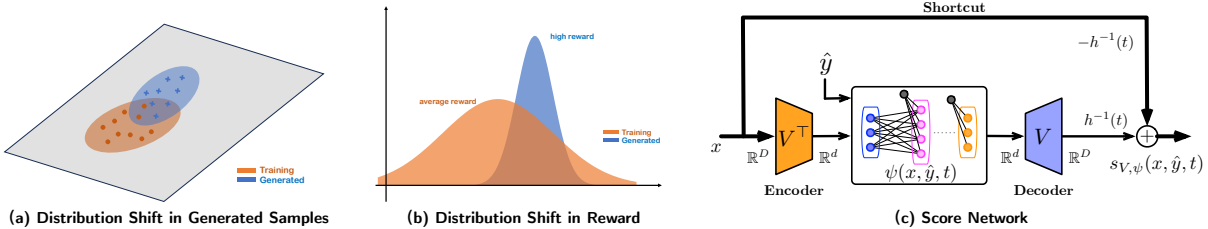}
    \caption{\textbf{Illustration of distribution shifts in samples and reward, as well as encoder-decoder score networks.} When performing reward-directed conditional generation, \textbf{(a)} the distribution of the generated data shifts, but still stays close to the feasible data support; \textbf{(b)} the distribution of the rewards for the generated data shifts and the average reward improves. \textbf{(c)}. The score network for reward-directed conditioned diffusion adopts an encoder-decoder structure.}
\label{fig:3-figure}
\end{figure}

The subtlety of this method, however, lies in that the new solution generation potentially conflicts with the training process: diffusion models are learned to generate solutions \textit{similar to} the training distribution, however, optimizing the objective function drives the model to \textit{deviate from} the training distribution. In other words, the model needs to both ``interpolate" and ``extrapolate". A higher value of the conditioning provides a stronger signal that guides the diffusion model towards higher objective function values, while the increasing distribution shift (i.e., the mismatch between the training data and the targeted generated data; see Figure~\ref{fig:3-figure}(a)(b)) may hurt the generated data' quality.

In this paper, we term various objective functions as reward functions and we introduce reward-directed conditional diffusion models for data-driven black-box optimization, with both the real-valued rewards and human preference. We show that reward-directed diffusion models enjoy strong theoretical guarantees and appealing empirical implementations. The theoretical crux to the success of our method centers around the following questions:
\begin{center}
    \textit{How to provably estimate the reward-conditioned distribution in black-box optimization via diffusion? \\
    How to properly choose high reward guidance in diffusion models for data generation, so as to ensure reward improvement among generated data?}
\end{center}
\paragraph{Our Approach.} To answer the questions above, we consider a semi-supervised learning setting, where we are given a small annotated dataset $\cD_{\rm label}$, and a massive unlabeled dataset $\cD_{\rm unlabel}$. \re{We study two forms of rewards in $\cD_{\rm label}$:
\begin{itemize}
\item {\bf (Real-valued reward)} The dataset consists of data and reward pairs, i.e., the reward is a real-valued noise-perturbed version of the underlying ground-truth reward;
\item {\bf (Human preference)} The dataset consists of triples taking two comparable data points and a binary preference label. The preference label indicates that the corresponding data point is likely to have an edge in the underlying reward over the other one.
\end{itemize}
Our semi-supervised formulation spans a wide range of applications. For example, in protein design problems, there is a large collection of unlabeled structures, while only a small fraction underwent expensive wet-lab tests. We remark that our method can also be applied to learning scenarios, where only labeled data is available.}

Our proposed method begins with estimating the reward function using $\dlabel$ and leveraging the estimator for pseudo-labeling on $\dunlabel$. Then we train a reward-conditioned diffusion model using the pseudo-labeled data.
% Our approach is illustrated in Figure~\ref{fig:approach}.
In real-world applications, there are other ways to incorporate the knowledge from the massive dataset $\dunlabel$, e.g., finetuning from a pre-trained model \citep{ouyang2022training, zhang2023adding}. We focus on the pseudo-labeling approach, as it provides a cleaner formulation and exposes the error dependency on data size and distribution shift. The intuition behind and the message are applicable to other semi-supervised approaches; see experiments in Section~\ref{sec:experiment:2}.

To model the data intrinsic low-dimension structures from a theoretical standpoint, we consider data $x$ to have a latent linear representation. Specifically, we assume $x = Az$ for some unknown matrix $A$ with orthonormal columns and $z$ being a latent variable. The latent variable often has a smaller dimension, reflecting the fact that practical datasets are often low-dimensional \citep{gong2019intrinsic, tenenbaum2000global, pope2021intrinsic}. The representation matrix $A$ should be learned to promote sample efficiency and generation quality \citep{chen2023score}. 
Our theoretical analysis reveals an intricate interplay between reward guidance, distribution shift, and implicit representation learning.
\paragraph{Contributions.} Our results are summarized as follows. 
\begin{itemize}
    \item[(1)]We show that the reward-conditioned diffusion model implicitly learns the latent subspace representation of $x$. Consequently, the model provably generates high-fidelity data that stays close to the subspace (Theorem \ref{thm:fidelity}).

\item [(2)] Given a target reward value, we analyze the sub-optimality of reward-directed generation, measured by the difference between the target value and the average reward of the generated data. In the case of a linear reward model, we show that the regret mimics the off-policy regret of linear bandits with full knowledge of the subspace feature. In other words, the reward-conditioned generation can be viewed as a form of off-policy bandit learning in the latent feature space. We further instantiate the \textit{real-valued reward} and \textit{human preference} setups, and prove that our algorithm achieves a converging sub-optimality under both cases (Theorem \ref{thm:parametric} and \ref{thm:rlhf-parametric}). 

\item[(3)] We further extend our theory to nonparametric settings where reward prediction and the training of diffusion models are parameterized by general function classes, which covers the wildly adopted ReLU neural networks in real-world implementations (Section \ref{sec:nonparametric}). 
\vspace{-0.2em}

\item[(4)] We provide numerical experiments on both synthetic data, text-to-image generation and offline reinforcement learning to support our theory (Section~\ref{sec:experiment}).
\vspace{-0.2em}
\end{itemize}
To our best knowledge, our results present the first statistical theory for conditioned diffusion models and provably reward improvement guarantees for reward-directed generation.

%% file: sections/related_work.tex
\section{Related Work}
\label{sec:related}
Our work lies in the intersection of three different streams: guided diffusion, theory for diffusion model, and black-box optimization.
\subsection{Guided Diffusion Models}
Diffusion models successfully incorporate diverse guidance in practical applications. For image generation, guiding the backward diffusion process towards higher log probabilities predicted by a classifier (which can be viewed as the reward signal) leads to improved sample quality, where the classifier can either be externally trained, i.e., classifier guidance \citep{dhariwal2021diffusion} or implicitly specified by a conditioned diffusion model, i.e., classifier-free guidance \citep{ho2022classifier}. Classifier-free guidance has become a standard technique in the state-of-the-art text-to-image diffusion models \citep{rombach2022high, ramesh2022hierarchical, balaji2022ediffi}. Other types of guidance are also explored in \citet{nichol2021glide, graikos2022diffusion, bansal2023universal}.

Similar ideas have been explored in sequence modeling problems. In offline reinforcement learning, Decision Diffuser~\citep{ajay2023is} is a diffusion model trained on offline trajectories and can be conditioned to generate new trajectories with high returns, satisfying certain safety constraints, or composing skills. For discrete generations, Diffusion LM~\citep{li2022diffusion} manages to train diffusion models on discrete text space with an additional embedding layer and a rounding step. The authors further show that gradients of any classifier can be incorporated to control and guide the text generation. These appealing empirical performance raises condensed curiosity of their theoretical underpinnings.

\subsection{Theory of Diffusion Models}
The theoretical foundation of diffusion models are gradually developing in the recent two years. A line of work studies diffusion models from a sampling perspective \citep{de2021diffusion, albergo2023stochastic, block2020generative, lee2022convergencea, chen2022sampling, lee2022convergenceb, chen2023restoration, chen2023probability, benton2023linear, montanari2023posterior}. When assuming access to a score function that can accurately approximate the ground truth score function in $L^\infty$ or $L^2$ norm, \cite{chen2022sampling, lee2023convergence} provide polynomial convergence guarantees of score-based diffusion models. \cite{de2022convergence} further studies diffusion models under the manifold hypothesis.

Recently, \cite{chen2023score, oko2023diffusion, mei2023deep, wibisono2024optimal} provide an end-to-end analysis of diffusion models, leveraging both sampling and statistical tools. In particular, they develop score estimation and distribution estimation guarantees of diffusion models. Statistical sample complexities are established and explain the success of diffusion models by their minimax optimality and adaptivity to data intrinsic structures. These results largely motivate our theory, whereas, we are the first to consider conditional score matching and statistical analysis of conditional diffusion models.

\subsection{Black-Box Optimization}
In this paper, we concentrate on black-box optimization using a pre-collected dataset, while additional first/second-order information cannot be accessed beyond the given dataset. This is sometimes termed the offline black-box or zero-order optimization \citep{fu2021offline,shahriari2015taking,luenberger1984linear}, and is different from the widely studied setting where active interactions with the objective function are allowed \citep{peters2007reinforcement,snoek2014input,hebbal2019bayesian,snoek2012practical}. To solve the black-box optimization without active function interactions, \cite{kumar2020model} propose a model inversion network that learns a stochastic inverse mapping from a reward value to the data instances. Then they generate new instances corresponding to the best possible reward. Such a method demonstrates empirical success, leaving theoretical understandings unexplored. Notably, \cite{krishnamoorthy2023diffusion} also empirically study offline black-box optimization with diffusion model with classifier-free guidance.

Our proposed method and analysis is related the offline bandit/RL theory \citep{munos2008finite, liu2018breaking, chen2019information, fan2020theoretical, jin2021pessimism, nguyen2021offline, brandfonbrener2021offline}. In particular, our theory extensively deals with distribution shift in the pre-collected dataset by class restricted divergence measures, which are commonly adopted in offline RL \citep{duan2020minimax, xie2021bellman, ji2022sample}. Moreover, our established sub-optimality bound of the generated data consists of an error term that coincides with off-policy linear bandits. However, our analysis goes far beyond the scope of bandit/RL.

%% file: sections/pre.tex
\section{Problem Setup}\label{sec:prob-setup}
Recall that we are presented with a pre-collected dataset for solving data-driven black-box optimization. The massive unlabeled dataset $\cD_{\rm unlabel}$ contains only feature vectors $\{x_i\}_{i=1}^{n_1}$, where $n_1$ denotes the sample size in $\cD_{\rm unlabel}$. For the labeled dataset $\cD_{\rm label}$, we study real-valued rewards and human preferences. In both cases, there exists a ground-truth reward function $f^*$, evaluating the quality of any data point $x$.

\noindent $\bullet$ {\bf Real-valued reward}. In this setting, we assume an access to label $y_i$ for each $x_i$. In such a case, $$\dlabel = \{(x_i, y_i)\}_{i=1}^{n_2},$$ where $n_2$ is the sample size, $x_i$ is randomly sampled and $y_i$ is a noisy observation of the reward:
\begin{equation}\label{eq:label-gen}
    y_i = f^*(x_i) + \xi_i \quad \text{for} \quad \xi_i \sim {\sf N}(0, \sigma^2).
\end{equation}
Here, $\sigma > 0$ is the variance of the observation noise. Such a setting corresponds to the scenario when we have a noisy labeler, who can ``inform'' us how good the generated data point is.

\noindent $\bullet$ {\bf Human preference}. In many real-life applications, a direct evaluation of reward can be impractical and highly biased, e.g., $\EE[\xi_i] \neq 0$ in \eqref{eq:label-gen}. Existing methods such as \citet{Schlett_2022} and \citet{babnik2023diffiqa} resort to biological or visual expressions to assess the data quality. Nonetheless, these methods are limited when it comes to aligning the generation process with human desired properties such as aesthetic quality and stylish requirements \citep{wang2023stylediffusion}. Thus, in many of these applications, extensive reward engineering or case-by-case training is required, which tends to be costly and time-consuming. Fortunately, recent revolutionary success in language models suggests that {\it human preference} is a sensible resource for handling hard-to-perceive abstract rewards \citep{ouyang2022training, lee2021pebble, wang2022skill,wu2021recursively,christiano2017deep}. In such scenarios, a human expert makes his/her preference among a pair of randomly sampled data points. The human preference reflects the ground-truth reward value, yet circumvents direct queries on the absolute reward values. Formally, our labeled dataset becomes
\begin{align*}
\dlabel = \{(x_i^{(1)}, x_i^{(2)}, u_i)\}_{i=1}^{n_2},
\end{align*}
where $x^{(1)}_i$ and $x^{(2)}_i$ are i.i.d. randomly sampled and $u_i \in \{x_i^{(1)}, x_i^{(2)}\}$ is the preferred instance. 

We model the preference as a probabilistic outcome dependent on the ground-truth reward of data $x$. In particular, we consider the widely adopted model Bradley-Terry model \citep{zhu2023principled, ouyang2022training}, where we have
\begin{align}\label{eq:human-pref}
\P(u \mid x^{(1)}, x^{(2)}) = \frac{\exp(f^*(u))}{\exp(f^*(x^{(1)})) + \exp(f^*(x^{(2)}))}.
\end{align}
Based on the preference information, we can estimate the underlying reward function and further utilize the estimator for inference. Since there is no need to perform real-valued reward engineering, using preference data reduces the reward tuning efforts and leads to outstanding empirical performance \citep{ouyang2022training,zhu2023principled,li2023reinforcement,stiennon2020learning,glaese2022improving,zhang2023high}.

We assume without loss of generality that $\dlabel$ and $\dunlabel$ are independent. In both datasets, $x$ is i.i.d. sampled from an unknown population distribution $P_x$.
In our subsequent analysis, we focus on the case where $P_x$ is supported on a latent subspace, meaning that the raw data $x$ admits a low-dimensional representation as in the following assumption.

\begin{assumption}
\label{assumption:subspace}
Data sampling distribution $P_x$ is supported on a low-dimensional linear subspace, i.e., 
$x = Az$ for an unknown $A \in \RR^{D \times d}$ with orthonormal columns and $z \in \RR^d$ is a latent variable. 
\end{assumption}
Note that our setup covers the full-dimensional setting as a special case when $d = D$. Yet the case of $d < D$ is much more interesting, as practical datasets are rich in intrinsic geometric structures \citep{gong2019intrinsic, pope2021intrinsic, tenenbaum2000global,kumar2020model}. 
Furthermore, the representation matrix $A$ may encode critical constraints on the generated data. For example, in protein design, the generated samples need to be similar to natural proteins and abide rules of biology, otherwise they easily fail to stay stable, leading to substantial reward decay \citep{LeeProtein}. In those applications, off-support data may be risky and suffer from a large degradation of reward. Base on this motivation, we further assume that the ground-truth reward has the following decomposition. 
\begin{assumption}\label{assumption:rew-decomp}
    We assume that $f^*(x) = g^*(\px) + h^*(\|\ox\|_2)$. Here $h^*$ is always nonpositive and $x = \px + \ox$, where $\px$ is the projection of $x$ onto subspace spanned by columns of $A$, and $\ox$ is the orthogonal complement.
\end{assumption}
Assumption \ref{assumption:rew-decomp} assumes that the reward can be decomposed into two terms: 1) a positive function that only concerns the projection of $x$ on the underlying representation $A$, and 2) a penalty term that measures how the data deviates from the subspace spanned by matrix $A$.

\section{Reward-Directed Generation via Conditional Diffusion Models}
\label{sec:alg}
In this section, we develop a conditioned diffusion model-based method to generate high-fidelity samples with desired properties. In real-world applications such as image/text generation and protein design, one often has access to abundant unlabeled data, but relatively limited number of labeled data. This motivates us to consider a semi-supervised learning setting.

\noindent {\bf Notation}: 
$P_{xy}$ denotes ground truth joint distribution of $x$ and its label $y$; $P_x$ is the marginal distribution of $x$. Any pair of data in $\dlabel$ follows $P_{xy}$ and any data in $\dunlabel$ follows $P_x$. $P$ is used to denote a distribution and $p$ denotes its corresponding density. $P(x \mid y=a)$ and $P(x, y=a)$ are the conditionals of $P_{xy}$
Similarly, we also use the notation $P_{x \hat{y}}$, $P(x \mid \hat{y}=a)$ for the joint and conditional of $(x, \hat{y})$, where the learned reward model predicts $\hat y$. Also, denote a generated distribution using diffusion by $\hat{P}$ (density $\hat{p}$) followed by the same argument in parentheses as the true distribution it approximates, e.g. $\hat{P}(x \mid y=a )$ is generated as an approximation of $P(x \mid y=a )$.

\subsection{Meta Algorithm}
\begin{figure}[htb!]
    \centering
    \includegraphics[width = 1\textwidth]{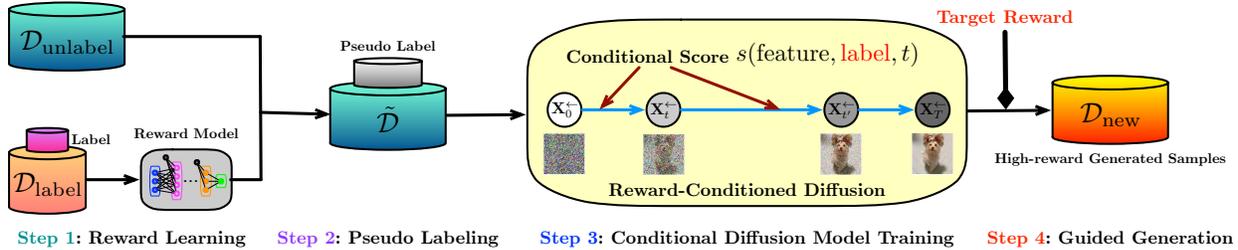} 
    \caption{Overview of solving black-box optimization using reward-directed conditional diffusion models. We estimate the reward function from the labeled dataset. Then we compute the estimated reward for each instance in the unlabeled dataset. Next, we train a reward-conditioned diffusion model using the pseudo-labeled data, and generate high reward new instances under proper target reward values.}
\label{fig:approach}
\end{figure}

\begin{algorithm}[ht]
\caption{Reward-Conditioned Generation via Diffusion Model (RCGDM)}
\label{alg:cdm}
\begin{algorithmic}[1]
\STATE {\bf Input}: Datasets $\dunlabel$, $\dlabel$, target reward value $a$, early-stopping time $t_0$, noise level $\nu$.\\
(Note: in the following psuedo-code, $\phi_t(x)$ is the Gaussian density and $\eta$ is the step size of discrete backward SDE, see Section \ref{sec:cond_diff} for elaborations on conditional diffusion)
\STATE {\bf Reward Learning}: Estimate the reward function by 
\begin{align}\label{eq:reward_regression}
\hat{f} \in \argmin_{f \in \cF} {L(f, \dlabel)},
\end{align}
where $\ell$ is a loss that depends on the dataset setup, see \S\ref{sec:prob-setup}, and $\cF$ is a function class. \label{step-rew-reg}
\STATE {\bf Pseudo labeling}: Use the learned function $\hat{f}$ to evaluate unlabeled data $\dunlabel$ and augment it with pseudo labeles: $\tilde{\cD} = \{(x_j, \hat{y}_j) = \hat{f}(x_j) + \xi_j \}_{j=1}^{n_1}$ for $\xi_j \overset{\text{i.i.d.}}{\sim} {\sf N}(0, \nu^2)$.
\STATE {\bf Conditional score matching}: \label{line_mtc} Minimize over $s \in \cS$ ($\cS$ constructed as \ref{equ:function_class}) on data set $\tilde{\cD}$ via %\mw{score class undefined}
\begin{align}
\label{equ:scr_mtc}
\hat{s} \in \argmin_{s \in \cS} \int_{t_0}^T \hat{\EE}_{(x, \hat{y}) \in \tilde{\cD}} \EE_{x^{\prime} \sim {\sf N}(\alpha(t)x, h(t)I_D)}\left[\norm{\nabla_{x^{\prime}} \log \phi_t(x^{\prime} | x) - s(x^{\prime}, \hat{y}, t)}_2^2\right] \diff t.
\end{align}
\STATE {\bf Conditioned generation}: Use the estimated score $\hat{s}(\cdot, a, \cdot)$ to sample from the backward SDE: \label{line_bcw}
\begin{align}
\label{eq:backward_discrete}
\diff \ttXytb = \left[\frac{1}{2} \tilde{X}_{t}^{y, \Leftarrow} + \hat{s}(\tilde{X}_{k\eta}^{y, \Leftarrow}, a, T - k\eta) \right] \diff t + \diff \overline{W}_t \quad \text{for} \quad t \in [k\eta, (k+1)\eta].
\end{align}
\STATE {\bf Return}: Generated population $\hat{P}(\cdot | \hat{y} = a)$, learned subspace representation $V$ contained in $\hat{s}$.
\end{algorithmic}
\end{algorithm}
We present our main algorithm in Algorithm \ref{alg:cdm} and also see a graphical illustration in Figure~\ref{fig:approach}. In order to generate novel samples with both high fidelity and high rewards, we propose Reward-Conditioned Generation via Diffusion Models (RCGDM). \re{Specifically, we learn the reward function $\hat{f}$ from the dataset $\dlabel$ via regression, where $L(f, \dlabel)$ is the corresponding loss function. In the real-valued reward setting, we specify $$L(f, \dlabel) = \sum_{i=1}^{n_2} (y_i - f(x_i))^2 + \lambda \rho(f),$$ where $\rho(f)$ is a regularization term. When $f$ is a linear function of $x$, we are actually doing ridge regression. In the human preference setting, we specify $$L(f, \dlabel) = \sum_{i=1}^{n_2} \{f(u_i) - \log\big(\exp(f(x_i^{(1)})) + \exp\big(f(x_i^{(2)})\big)\big)\},$$ which is essentially an MLE objective.}  We then use $\hat{f}$ to augment the unlabeled data $\dunlabel$ with ``pseudo labeling" and additive noise, i.e., $\tilde{\cD} = \{(x_j, \hat{y}_j = \hat{f}(x_j)+\xi_j)\}_{j=1}^{n_1}$ with $\xi_j \sim {\sf N}(0, \nu^2)$ of a small variance $\nu^2$. Here, we add noise $\xi_j$ merely for technical reasons in the proof. We denote the joint distribution of $(x, \hat{y})$ as $P_{x\hat{y}}$. Next, we train a conditional diffusion model using the augmented dataset $\tilde{D}$. If we specify a target value of the reward, for example letting $\hat{y}=a$, we can generate conditioned samples from the distribution $\hat P(x| \hat y =a) $ by backward diffusion. 

\subsection{Training of Conditional Diffusion Model}
\label{sec:cond_diff}
We provide details about the training and sampling of conditioned diffusion in Algorithm \ref{alg:cdm} (Line \ref{line_mtc}: conditional score matching and Line \ref{line_bcw}: conditional generation). 
In Algorithm \ref{alg:cdm}, conditional diffusion model is learned with $\tilde{\cD} = \{(x_j, \hat{y}_j = \hat{f}(x_j)+\xi_j)\}_{j=1}^{n_1}$, where $(x, \hat{y}) \sim P_{x \hat{y}}$. For simplicity, till the end of this section we use $y$ instead of $\hat{y}$ to denote the condition variable. The diffusion model is to approximate the conditional probability $P(x \mid \hat{y})$.

\subsection{Conditional Score Matching} 
The working flow of conditional diffusion models is nearly identical to that of unconditioned diffusion models reviewed in \cite{chen2023score}. A major difference is we learn a conditional score $\nabla \log p_t(x | y)$ instead of the unconditional one. Here $p_t$ denotes the marginal density function at time $t$ of the following forward O-U process,
\begin{align}\label{eq:forward}
\diff X_t^y = -\frac{1}{2} g(t) X_t^y \diff t + \sqrt{g(t)} \diff W_t \quad \text{with} \quad X_0^y \sim P_0(x | y) ~\text{and}~ t \in (0, T],
\end{align}
where similarly $T$ is a terminal time, $(W_t)_{t \geq 0}$ is a Wiener process, and the initial distribution $P_0(x | y)$ is induced by the $(x ,\hat{y})$-pair distribution $P_{x \hat{y}}$. Note here the noise is only added on $x$ but not on $y$. Throughout the paper, we consider $g(t) = 1$ for simplicity. We denote by $P_t(x_t | y)$ the distribution of $\Xyt$ and let $p_t(x_t | y)$ be its density and $P_t(x_t, y)$ be the corresponding joint, shorthanded as $P_t$. A key step is to estimate the unknown $\nabla \log p_t(x_t | y)$ through denoising score matching \citep{song2020score}. A conceptual way is to minimize the following quadratic loss with $\cS$, a concept class:
\begin{equation}\label{eq1}
    \argmin_{s \in \cS} \int_{0}^T \EE_{(x_t, y) \sim P_t} \left[\norm{\nabla \log p_t(x_t | y) - s(x_t, y, t)}_2^2\right] \diff t,
\end{equation}
Unfortunately, the loss in \eqref{eq1} is intractable since $\nabla \log p_t(x_t | y)$ is unknown. Inspired by \cite{hyvarinen2005estimation} and \cite{vincent2011connection}, we choose a new objective \eqref{equ:scr_mtc} and show their equivalence in the following Proposition. The proof is provided in Appendix~\ref{pf:equivalent_score_matching}. 
\begin{proposition}[{Score Matching Objective for Implementation}]
\label{prop:equivalent_score_matching}
For any $t > 0$ and score estimator $s$, there exists a constant $C_t$ independent of $s$ such that 
\begin{align}\label{eq:equivalent_score_matching}
& \EE_{(x_t, y) \sim P_t} \left[\norm{\nabla \log p_t(x_t | y) - s(x_t, y, t)}_2^2\right] \nonumber \\
& =\EE_{(x, y) \sim P_{x \hat{y}}} \EE_{x^{\prime} \sim {\sf N}(\alpha(t)x, h(t)I_D)}\left[\norm{\nabla_{x^{\prime}} \log \phi_t(x^{\prime} | x) - s(x^{\prime}, y, t)}_2^2\right] + C_t,
\end{align}
where $\nabla_{x'} \log \phi_t(x' | x) = -\frac{x' - \alpha(t)x}{h(t)}$, where $\phi_t(x' | x)$ is the density of ${\sf N}(\alpha(t)x, h(t)I_D)$ with $\alpha(t) = \exp(- t/2)$ and $h(t) = 1 - \exp(-t)$.
\end{proposition}
The proof of Proposition \ref{prop:equivalent_score_matching} can be found in Appendix \ref{pf:equivalent_score_matching}. Here Equation~\eqref{eq:equivalent_score_matching} allows an efficient implementation, since $P_{x \hat{y}}$ can be approximated by the empirical data distribution in $\tilde{\mathcal{D}}$ and $x'$ is easy to sample. Integrating \eqref{eq:equivalent_score_matching} over time $t$ leads to a practical conditional score matching object
\begin{align}\label{eq:score_matching_practical}
\argmin_{s \in \cS} \int_{t_0}^T \hat{\EE}_{(x, y) \sim P_{x \hat{y}}} \EE_{x^{\prime} \sim {\sf N}(\alpha(t)x, h(t)I_D)}\left[\norm{\nabla_{x^{\prime}} \log \phi_t(x^{\prime} | x) - s(x^{\prime}, y, t)}_2^2\right] \diff t,
\end{align}
where $t_0 > 0$ is an early-stopping time to stabilize the training \citep{song2020improved, vahdat2021score} and $\hat{\EE}$ denotes the empirical distribution.

Constructing a function class $\cS$ adaptive to data structure is beneficial for learning the conditional score. In the same spirit of \cite{chen2023score}, we propose the score network architecture (see Figure \ref{fig:3-figure}(c) for an illustration): 
{\small
\begin{align}
\label{equ:function_class}
\cS = \bigg\{\sbb_{V, \psi}(x, y, t) = \frac{1}{h(t)} (V \cdot \psi (V^\top x, y, t) - x) & :~ V \in \RR^{D \times d}, ~\psi \in \Psi:\RR^{d+1} \times [t_0, T] \to \RR^d~\bigg\},
\end{align}}
with $V$ being any $D \times d$ matirx with orthonormal columns and $\Phi$ a customizable function class. This design has a linear encoder-decoder structure, catering to the latent subspace structure in data. Also $-\frac{1}{h(t)} x$ is included as a shortcut connection.

\subsection{Conditioned Generation}
Sampling from the model is realized by running a discretized backward process with step size $\eta > 0$ described as follows:
\begin{align}
\tag{Eqn. \eqref{eq:backward_discrete} revisited}
\diff \ttXytb = \left[\frac{1}{2} \tilde{X}_{t}^{y, \Leftarrow} + \hat{s}(\tilde{X}_{k\eta}^{y, \Leftarrow}, y, T - k\eta) \right] \diff t + \diff \overline{W}_t \quad \text{for} \quad t \in [k\eta, (k+1)\eta].
\end{align}
initialized with $\ttXytb \sim {\sf N}(0, I_D)$ and $\overline{W}_t$ is a reversed Wiener process. Note that in \eqref{eq:backward_discrete}, the unknown conditional score $\nabla p_t(x | y)$ is substituted by $\hat{s}(x, y, t)$. 
\remark[Alternative methods] \re{In Algorithm~\ref{alg:cdm} we train the conditional diffusion model via conditional score matching (Line \ref{line_mtc}). This approach is suitable when we have access to the unlabeled dataset and need to train a brand-new diffusion model from scratch. Practically, we can utilize a pre-trained diffusion model of the unlabeled data directly and incorporate the reward information to the pre-trained model. Popular methods in this direction include classifier-based guidance~\citep{dhariwal2021diffusion}, fine-tuning \citep{zhang2023adding}, and self-distillation \citep{song2023consistency}; all sharing a similar spirit with Algorithm \ref{alg:cdm}.}

%% file: sections/method.tex
\section{Statistical Analysis of Reward-Directed Condition Diffusion}\label{sec:linear}
In this section, let us analyze the reward-conditioned diffusion model give by Algorithm~\ref{alg:cdm} for data-driven optimization. In particular, we are interested in two properties of the generated samples: 1) the reward levels of these new samples and 2) their level of fidelity -- how much do new samples deviate from the latent subspace. To measure these properties, we adopt two sets of metrics. Firstly, the sub-optimality gap quantifies the average reward of the generated samples to the target reward value. Secondly, the subspace angle and an expected $l_2$ deviation quantify the sample fidelity. In the sequel, we analyze these performance metrics, relating them to the reward learning and diffusion model training steps in Algorithm~\ref{alg:cdm}.
\subsection{Sub-Optimality Decomposition}
We introduce the performance metric for a generated distribution with respect to a certain target reward value. Let $y^*$ be a target reward value and $P$ be a distribution learned by Algorithm~\ref{alg:cdm} (we omit the hat notation on $P$ for simplicity). We define the sub-optimality of $P$ as
$$\subopt(P; y^*) = y^*-  \EE_{x \sim P}[f^*(x)],$$
which measures the gap between the expected reward of $x \sim P$ and the target value $y^*$. In the language of bandit learning, this gap can be viewed as a form of {\it off-policy sub-optimality}.

We introduce the following proposition, which decomposes the sub-optimality $\subopt(P; y^*)$. Specifically, we show that $\subopt(P; y^* = a)$ comprises of three components: off-policy bandit regret which comes from the estimation error of the reward function, on-support and off-support errors coming from estimating conditional distributions with diffusion.
\begin{proposition}\label{prop:decom}
    Set the target reward value $y^* = a$. For a generated distribution $P$ of Algorithm \ref{alg:cdm}, it holds that
\begin{align*}
    \subopt(P ; y^* = a) 
    \leq \underbrace{\EE_{x \sim P_a} \left[ \left|\hat{f}(x) - g^*(x)\right| \right]}_{\cE_1:\textrm{off-policy bandit regret}} + \underbrace{\left| \EE_{x \sim P_a} [g^*(\px)] - \EE_{x \sim P} [g^*(\px)]\right|}_{\cE_2: \textrm{on-support diffusion error}} +  \underbrace{\EE_{x \sim P} [h^*(\|\ox\|_2)]}_{\cE_3: \textrm{off-support diffusion error}}.
\end{align*}
\end{proposition}
The proof is provided in Appendix \ref{sec:dcp}. Error $\cE_1$ corresponds to the reward function estimation error, errors $\cE_2$ and $\cE_3$ are induced by the training of diffusion model.
\subsection{Sample Fidelity}
In this section, we analyze the fidelity of the conditional generation process specified by Algorithm \ref{alg:cdm}. Specifically, we assess its ability to capture the underlying representation, and how the data generated by the conditional diffusion model deviates the ground truth representation.  
Recall that Algorithm \ref{alg:cdm} has two outputs: the generated distribution $\hat{P}(\cdot | \hat{y} = a)$ and the learned representation matrix $V$. We use notations $\hat{P}_a:= \hat{P}(\cdot | \hat{y} = a)$ (generated distribution) and $P_a:= P(\cdot | \hat{y} = a)$ (target distribution) for better clarity in the rest of the presentation. We define two metrics
\begin{align}
\subangle{V}{A} = \norm{VV^\top - AA^\top}_{\rm F}^2 \text{\quad and \quad} \EE_{x \sim \hat{P}_a} [\|\ox\|_2].
\end{align}
Here, $\subangle{V}{A}$ is defined for matrices $V$ and $A$, where $A$ is the matrix encoding the ground truth subspace. Clearly, $\subangle{V}{A}$ measures the difference in the column span of $V$ and $A$, which is also known as the subspace angle. Term $\EE_{x \sim \hat{P}_a} [\|\ox\|_2]$ is defined as the expected $l_2$ distance between $x$ and the subspace.

We impose the following commonly adopted light-tail assumption \citep{wainwright2019high} on the latent distribution $z$ for quantifying sample fidelity.
\begin{assumption}
\label{asmp:tail}
The latent variable $z$ follows distribution $P_{z}$ with density $p_z$, such that there exist constants $B, C_1, C_2$ verifying $p_{z}(z) \leq (2\pi)^{-d/2} C_1 \exp\left(-C_2 \norm{z}_2^2 / 2\right)$ whenever $\norm{z}_2 > B$. Moreover, the ground truth score is realizable, i.e., $\nabla \log p_t(x \mid \hat y) \in \cS$.
\end{assumption}
We remark that the realizability condition is for technical convenience, which can be removed in our nonparametric extension in Section~\ref{sec:nonparametric}. Assumption~\ref{asmp:tail} is also mild and covers arbitrary sub-Gaussian latent data distributions, which include the following Gaussian assumption as a special case.
\begin{assumption}\label{assumption:gaussian_design}
The latent variable $z \sim {\sf N}(0, \Sigma)$ with a covariance matrix $\Sigma$ satisfying $\lambda_{\min}I_d \preceq \Sigma \preceq \lambda_{\max}I_d$ for $\lambda_{\min} \leq \lambda_{\max} \leq 1$ and $\lambda_{\min}$ being positive.
\end{assumption}

Now we present sample fidelity guarantees.
\begin{theorem}[{Subspace Fidelity of Generated Data}] 
\label{thm:fidelity}
Under Assumption \ref{assumption:subspace}, if Assumption \ref{asmp:tail} holds with $c_0 I_d \preceq \EE_{z \sim P_z} \left[ z z^{\top}\right]$ for some constant $c_0 > 0$, then with high probability on data, it holds that
\begin{equation}
\label{eq:subspave_covering}
    \subangle{V}{A} = \tilde{\cO} \left( \frac{1}{c_0} \sqrt{\frac{\cN(\cS, 1/n_1) D} {n_1}} \right)
\end{equation}
with $\cN(\cS, 1/n_1)$ being the log covering number of function class $\cS$ as in Equation \eqref{equ:function_class}. Moreover, with Assumption  \ref{assumption:gaussian_design} and $D \geq d^2$, we have $\cN(\cS, 1/n_1) = \cO((d^2 + Dd)\log (D d n_1))$  and thus $\subangle{V}{A} = \tilde{\cO} ( \frac{1}{\lambda_{\min}} \sqrt{\frac{Dd^2 + D^2d} {n_1}})$.
\end{theorem}
The proof is provided in Appendix \ref{pf:fidelity-opr}. In Theorem \ref{thm:fidelity}, we provide a fidelity guarantee for the learned subspace $V$. Specifically, when the covariance of the underlying representation $z$ is invertible, we prove that the error in subspace learning converges in a parametric rate.
\begin{remark} We make the following remarks for Theorem \ref{thm:fidelity}.
\begin{enumerate}
\item Guarantee \eqref{eq:subspave_covering} applies to general distributions with light tail as assumed in Assumption \ref{asmp:tail}.
\item Guarantee \eqref{eq:off_norm} guarantees high fidelity of generated data in terms of staying in the subspace when we have access to a large unlabeled dataset.
\item Guarantee \eqref{eq:off_norm} shows that $\EE_{x \sim \hat{P}_a} [\|\ox\|_2]$ scales up when $t_0$ goes up, which aligns with the dynamic in backward process that samples are concentrating to the learned subspace as $t_0$ goes to $0$. Taking $t_0 \to 0$, $\EE_{x \sim \hat{P}_a} [\|\ox\|]$ has the decay in $O(n_1^{-\frac{1}{4}})$. However, taking $t_0 \to 0$ is not ideal for the sake of high reward of $x$, we take the best trade-off of $t_0$ later in Theorem \ref{thm:parametric}.
\end{enumerate}
\end{remark}

%% file: sections/casestudy.tex
\section{Optimization Theory}
In this section, we provide optimization guarantees by instantiating our performance metrics. We consider two theoretical settings: the parametric and nonparametric settings. In the parametric setting, we focus on a linear reward model with off-support penalties. The reward observation in $\dlabel$ covers both the real-valued and human preference cases. Through detailed analysis, we establish handy guidance on how to choose a proper target reward value for ensuring quality improvement. We then extend our results to nonparametric settings, where a ReLU neural network is used for reward estimation. We then present the sub-optimality guarantee in the corresponding context.

\subsection{Results under Real-Valued Reward}
In this subsection, we consider the case when we have access to a noisy reward label for every $x_i$ in $\dlabel$, i.e. $\dlabel = \{(x_i. y_i)\}_{i\in[n_2]}$, where $y_i$ is sampled according to \eqref{eq:label-gen}.
Recall that in Assumption \ref{assumption:rew-decomp}, the ground truth reward has the decomposition $f^*(x) = g^*(\px) + h^*(\ox)$, where $\px$ is the projection of $x$ on subspace $A$, and $\ox$ is the part orthogonal to $A$. We consider two setups for $g^*(\px)$: 1) the \textit{parametric setting}, where $g^*(\px) $ is a linear function of $\px$; 2) the \textit{nonparametric setting}, where $g^*(\px)$ is a general H\"{o}lder continuous function, and we use a deep ReLU neural network for reward estimation.
\paragraph{Parametric Setting.}
For the parametric setup, we make the following assumption. Such an assumption assumes that the reward function consists of two parts: a linear function of $x$ and a penalty term that increases in magnitude when $x$ deviates from the representation subspace.  
\begin{assumption}
\label{assumption:linear_reward}
 We assume that $g^*(\px) = (\theta^*)^\top \px$ where $\theta^* = A \beta^*$ for some $\beta^* \in \RR^d$ and $\norm{\theta^*}_2 = \norm{\beta^*}_2 = 1$. The penalty $h^*(\ox)$ is always nonpositive and decreases with respect to $\norm{\ox}_2$ with $h^*(0) = 0$.
\end{assumption}
Assumption \ref{assumption:linear_reward} adopts a simple linear reward model, which will be generalized in Section~\ref{sec:nonparametric}. \re{ We now present Theorem \ref{thm:opr}, which is a corollary of Theorem \ref{thm:fidelity} and characterizes the deviation of the generated samples from the ground-truth representation subspace. }

\begin{theorem}\label{thm:opr}
Assume that the conditions in Theorem \ref{thm:fidelity} hold. Further under Assumption \ref{assumption:linear_reward}, it holds that
\begin{equation}
\label{eq:off_norm}
    \EE_{x \sim \hat{P}_a} [\|\ox\|_2] = \cO \left( \sqrt{t_0D} + \sqrt{\subangle{V}{A}}  \cdot \sqrt{\frac{a^2}{\|\beta^*\|_{\Sigma}} + d} \right),
\end{equation}
where $\beta^*$ is the ground-truth parameter of linear model and $t_0$ is the early-stopping time in diffusio models.
\end{theorem}
The proof is provided in Appendix \ref{pf:fidelity-opr}. We observe that larger target reward $a$ leads to more severe subspace deviation, since aggressive extrapolation is needed to match the target reward. Therefore, to maintain high sample fidelity prevents us from setting $a$ too large.

\re{On the other hand, we examine the sub-optimality bound. We begin with the real-valued rewards, where a labeled dataset $\dlabel = \{(x_i, y _i)\}_{i=1}^{n_2}$ is given. In this case, we perform ridge regression to estimate the linear reward function in Step \ref{step-rew-reg} of Algorithm \ref{alg:cdm}, i.e., $$\argmin_{\theta}~ \sum_{i=1}^{n_2} (\theta^\top x_i - y _i)^2 + \lambda \|\theta\|_2^2$$ for a positive coefficient $\lambda$. We now characterize the sub-optimality bound in the following theorem.}
\begin{theorem}[{Off-policy Regret of Generated Samples}]\label{thm:parametric}
Suppose Assumptions~\ref{assumption:subspace}, \ref{assumption:gaussian_design} and \ref{assumption:linear_reward} hold. We choose $\lambda = 1$,
$t_0 = \left((Dd^2 + D^2d) / n_1\right)^{1/6}$ and $\nu = 1/\sqrt{D}$. With high probability, running Algorithm~\ref{alg:cdm} with a target reward value $a$ gives rise to 
\begin{align}\label{equ:subopt}
& \quad \subopt(\hat{P}_a ; y^* = a) \nonumber \\
& \leq\underbrace{\sqrt{\operatorname{Tr}(\hat{\Sigma}_{\lambda}^{-1}\Sigma_{P_a})} \cdot \cO \left(\sqrt{\frac{d \log n_2}{n_2}} \right)}_{\cE_1:\textrm{off-policy bandit regret}} + \underbrace{{\tt DistroShift}(a) \cdot \left(d^2D + D^2 d\right)^{1/6} {n_1}^{-1/6} \cdot a}_{\cE_2:\textrm{on-support diffusion error}} \nonumber \\
& \quad + \underbrace{\EE_{\hat{P}_a} [h^*(\ox)]}_{\cE_3:\textrm{off-support diffusion error}},
\end{align}
where $\hat{\Sigma}_{\lambda}:= \frac{1}{n_2}(X^{\top} X+\lambda I_D)$ where $X$ is the stack matrix of $\dlabel$ and $\Sigma_{P_a} = \EE_{P_a} [xx^{\top}]$. 
\end{theorem}
The proof is provided in Appendix \ref{pf:parametric}.
\paragraph{Implications and Discussions:} 

\noindent {1).} Equation \eqref{equ:subopt} and \eqref{equ:hf-subopt} decompose the suboptimality gap into two separate parts of error: error from reward learning ($\cE_1$) and error coming from diffusion ($\cE_2$ and $\cE_3$).

\noindent {2).}  $\cE_1$ depending on $d$ shows diffusion model learns a low-dimensional representation of $x$, reducing $D$ to smaller latent dimension $d$. It can be seen from $\operatorname{Tr}( \hat{\Sigma}_{\lambda}^{-1} \Sigma_{p_{q}}) \leq \cO \left(\frac{ a^2}{ \|{\beta}^* \|_{\Sigma}} + d \right)$ when $n_2 = \Omega(\frac{1}{\lambda_{\min}})$. \re{Similarly, we can prove that $\operatorname{Tr}( \tilde{\Sigma}_{\lambda}^{-1} \Sigma_{P_a})\leq \cO \left(\frac{ a^2}{ \|{\beta}^* \|_{\Sigma}} + d \right)$ when $n_2 = \Omega(\frac{1}{\lambda_{\min}})$. Such a result shows that even without an observable reward, we can still obtain an off-policy bandit regret that matches the case with an observable reward dataset. }

\noindent {3).} If we ignore the diffusion errors, the suboptimality gap resembles the standard regret of off-policy bandit learning in $d$-dimensional feature subspace \citep[Section 3.2]{jin2021pessimism}, \citep{nguyen2021offline, brandfonbrener2021offline}. 

\noindent {4).} It is also worth mentioning that $\cE_2$ and $\cE_3$ depend on $t_0$ and that by taking $t_0 =\left((Dd^2 + D^2d) / n_1\right)^{1/6}$ one gets a good trade-off in $\cE_2$.

\noindent {5).} On-support diffusion error entangles with distribution shift in complicated ways. We show
\begin{align*}
\cE_2 = \left({\tt DistroShift}(a) \cdot \left(d^2D + D^2 d\right)^{1/6} {n_1}^{-1/6} \cdot a\right),
\end{align*}
where ${\tt DistroShift}(a)$ quantifies the distribution shift depending on different reward values. In the special case of the latent covariance matrix $\Sigma$ is known, we can quantify the distribution shift as ${\tt DistroShift}(a) = \cO(a \vee d)$.
We observe an interesting phase shift. When $a < d$, the training data have sufficient coverage with respect to the generated distribution $\hat{P}_a$. Therefore, the on-support diffusion error has a lenient linear dependence on $a$. However, when $a > d$, the data coverage is very poor and $\cE_2$ becomes quadratic in $a$, which quickly amplifies.

\noindent {6).} When generated samples deviate away from the latent space, the reward may substantially degrade as determined by the nature of $h$.

\subsection{Extension to Nonparametric Setting}\label{sec:nonparametric}
Our theoretical analysis, in its full generality, extends to using general nonparametric function approximation for both the reward and score functions. Built upon the insights from the linear setting, we provide an analysis of the nonparametric reward and general data distribution setting. We generalize Assumption~\ref{assumption:linear_reward} to the following.
\begin{assumption}\label{assumption:nonparametric}
We assume that $g^*(\px)$ is $\alpha$-H\"{o}lder continuous for $\alpha \geq 1$. Moreover, $g^*$ has a bounded H\"{o}lder norm, i.e., $\norm{g^*}_{\cH^{\alpha}} \leq 1$. 
\end{assumption}
H\"{o}lder continuity is widely studied in nonparametric statistics literature \citep{gyorfi2002distribution, tsybakov2008intro}.  Under Assumption~\ref{assumption:nonparametric}, we use nonparametric regression for estimating $f^*$. Specifically, we specialize \eqref{eq:reward_regression} in Algorithm~\ref{alg:cdm} to
\begin{align*}
\hat{f}_\theta \in \argmin_{f_\theta \in \cF} \frac{1}{2n} \sum_{i=1}^{n_1} (f_\theta(x_i) - y_i)^2,
\end{align*}
where $\cF $ is chosen to be a class of neural networks. Hyperparameters in $\cF$ will be chosen properly in Theorem~\ref{thm:nonparametric}.

Our theory also considers generic sampling distributions on $x$. Since $x$ lies in a low-dimensional subspace, this translates to a sampling distribution assumption on latent variable $z$.
\begin{assumption}
\label{assumption:tail_non}
The latent variable $z$ follows a distribution $P_{z}$ with density $p_z$, such that there exist constants $B, C_1, C_2$ verifying $p_{z}(z) \leq (2\pi)^{-d/2} C_1 \exp\left(-C_2 \norm{z}_2^2 / 2\right)$ whenever $\norm{z}_2 > B$. Moreover, there exists a positive constant $c_0$ such that $c_0 I_d \preceq \EE_{z \sim P_z} \left[ z z^{\top}\right]$.
\end{assumption}
Assumption~\ref{assumption:tail_non} says $P_z$ has a light tail and also encodes distributions with compact support. Furthermore, we assume that the curated data $(x, \hat{y})$ induces Lipschitz conditional scores. Motivated by \cite{chen2023score}, we show that the linear subspace structure in $x$ leads to a similar conditional score decomposition $\nabla \log p_t(x | \hat{y}) = s_{\parallel}(x, \hat{y}, t) + s_{\perp}(x, \hat{y}, t)$, where $s_{\parallel}$ is the on-support score and $s_{\perp}$ is the orthogonal score. The formal statement and proof for this decomposition are deferred to Appendix \ref{sec:score_matching_err}. We make the following assumption on $s_{\parallel}$.
\begin{assumption}
\label{asmp:lipschitz}
The on-support conditional score function $s_{\parallel}(x, \hat{y}, t)$ is Lipschitz with respect to $x, \hat{y}$ for any $t \in (0, T]$, i.e., there exists a constant $\Clip$, such that for any $x, \hat{y}$ and $x^{\prime}, \hat{y}^{\prime}$, it holds
\begin{align*}
\norm{s_{\parallel}(x, \hat{y}, t) - s_{\parallel}(x', \hat{y}', t)}_2 \leq \Clip \norm{x - x^{\prime}}_2 + \Clip |\hat{y} - \hat{y}^{\prime}|.
\end{align*}
\end{assumption}
Lipschitz score is commonly adopted in existing works \citep{chen2022sampling, lee2023convergence}. Yet Assumption~\ref{asmp:lipschitz} only requires the Lipschitz continuity of
the on-support score, which matches the weak regularity conditions in \cite{lee2023convergence, chen2023score}. We then choose the score network architecture similar to that in the linear reward setting, except we replace $m$ by a nonlinear network. Recall the linear encoder and decoder estimate the representation matrix $A$.

We consider feedforward networks with ReLU activation functions as concept classes $\cF$ and $\cS$ for nonparametric regression and conditional score matching. Given an input $x$, neural networks compute
\begin{align}\label{eq:fnn}
f_{\rm NN}(x) = W_L \sigma( \dots \sigma(W_1 x + b_1) \dots ) + b_{L},
\end{align}
where $W_i$ and $b_i$ are weight matrices and intercepts, respectively.
We then define a class of neural networks as
\begin{align*}
& {\rm NN}(L, M, J, K, \kappa) = \Big \{f: f~\text{in the form of \eqref{eq:fnn} with $L$ layers and width bounded by~}M, \\
& \sup_{x} \norm{f(x)}_2 \leq K, \max\{\norm{b_i}_\infty, \norm{W_i}_\infty\} \leq \kappa ~\text{for}~i = 1, \dots, L, ~\text{and}~ \sum_{i=1}^L \big( \norm{W_i}_0 + \norm{b_i}_0 \big) \leq J \Big\}.
\end{align*}
For the conditional score network, we will additionally impose some Lipschitz continuity requirement, i.e., $\norm{f(x) -f(y)}_2 \leq \clip\norm{x - y}_2$ for some Lipschitz coefficient $\clip$. Now we are ready to bound $\subopt(\hat{P}_a ; y^* = a)$ in Theorem~\ref{thm:nonparametric} in terms of non-parametric regression error, score matching error and distribution shifts in both regression and score matching.

\begin{theorem}\label{thm:nonparametric}
Suppose Assumptions~\ref{assumption:subspace}, \ref{assumption:nonparametric}, \ref{assumption:tail_non} and \ref{asmp:lipschitz} hold. Let $\delta(n) = \frac{d\log\log n}{\log n}$. Properly chosen $\cF$ and $\cS$, with high probability, running Algorithm~\ref{alg:cdm} with a target reward value $a$ and early-stopping time $ t_0 = \left(n_1^{-\frac{2 - 2\delta(n_1)}{d+6}} + Dn_1^{-\frac{d+4}{d+6}}\right)^{\frac{1}{3}}$ gives rise to $\subangle{V}{A} \leq \tilde{\cO}\left( \frac{1}{c_0} \left(n_1^{-\frac{2 - 2\delta(n_1)}{d+6}} + Dn_1^{-\frac{d+4}{d+6}} \right) \right)$ and
\begin{align*}
& \quad \subopt(\hat{P}_a ; y^* = a) \\
& \leq \underbrace{\sqrt{\cT(P(x | \hat{y} = a), P_{x}; \bar{\cF})}  \cdot \tilde{\cO}\left(n_2^{-\frac{\alpha - \delta(n_2)}{2\alpha + d}} + D / n_2\right)}_{\cE_1}\\
& \quad + \underbrace{ \left(\sqrt{\frac{\cT(P(x, \hat{y} = a), P_{x\hat{y}}; \bar{\cS})}{c_0}} \cdot \|g^*\|_{\infty} +\sqrt{M(a)}\right) \cdot \tilde \cO \left( \left(n_1^{-\frac{2 - 2\delta(n_1)}{d+6}} + Dn_1^{-\frac{d+4}{d+6}}\right)^{\frac{1}{3}}\right)}_{\cE_2}\\
& \quad + \underbrace{\EE_{x \sim \hat{P}_a} [h^*(\ox)]}_{\cE_3},
\end{align*}
where $M(a): = \EE_{z \sim \mathbb{P} ( a)} [\|z\|^2_2]$ and
\begin{align*}
   \bar{\cF}:= \{|f^*(x) - f(x)|^2: f \in \cF\}, \quad 
   \bar{\cS} = \left\{\frac{1}{T - t_0} \int_{t_0}^T \EE_{x_t \mid x} \norm{\nabla \log p_t(x_t \mid y) - s(x_t,  y, t)}_2^2 \diff t : s \in \cS\right\},
\end{align*}
$\cE_3$ penalizes the component in $\hat{P}_a$ that is off the subspace. The function classes $\cF$ and $\cS$ are chosen as $\cF = {\rm NN}(L_f, M_f, J_f, K_f, \kappa_f)$ with
\begin{align*}
& L_f = \cO(\log n_2), \ M_f = \cO\left(n_2^{-\frac{d}{d+2\alpha}} (\log n_2)^{d/2} \right), \ J_f = \cO\left(n_2^{-\frac{d}{d+2\alpha}} (\log n_2)^{d/2+1} \right) \\
& \hspace{1.8in} K_f = 1, \ \kappa_f = \cO\left(\sqrt{\log n_2}\right)
\end{align*}
and $\cS = {\rm NN}(L_s, M_s, J_s, K_s, \kappa_s)$ with
\begin{align*}
& L_s = \cO(\log n_1 + d), \ M_s = \cO\left(d^{d/2} n_1^{-\frac{d+2}{d+6}} (\log n_1)^{d/2} \right), \ J_s = \cO\left(d^{d/2} n_1^{-\frac{d+2}{d+6}} (\log n_1)^{d/2+1} \right) \\
& \hspace{1.5in} K_s = \cO\left(d\log (dn_1) \right), \ \kappa_s = \cO\left(\sqrt{d \log (n_1 d)}\right).
\end{align*}
Moreover, $\cS$ is also Lipschitz with respect to $(x, y)$ and the Lipschitz coefficient is $\clip = \cO\left(10d \Clip \right)$.
\end{theorem}
The proof is provided in Appendix~\ref{pf:nonparametric}. Here the $\delta(n)$ term accounts for the unbounded domain of $x$, which is negligible when $n$ is large. As a consequency, we prove that under a nonparametric setting, Algorithm \ref{alg:cdm} enjoys a $\tilde{\cO}\big(\dshift(a) \cdot \big(n_2^{-\frac{\alpha}{2\alpha+d}} + n_1^{-\frac{2}{3(d+6)}}\big)\big)$ suboptimality when the neural network is properly chosen.

\subsection{Results under Human Preference}
In this section, we study the setup where real-valued reward $y$ is not observable, and the labeled dataset consists of pairwise comparisons, i.e., $\dlabel = \{x_i^{(1)}, x_i^{(2)}, u_i\}_{i=1}^{n_2}$, where $u_i$ is the human preference between $\{x_i^{(1)}, x_i^{(2)}\}$ for every $i \in [n_2]$. Recall that we model human preference by the Bradley-Terry model, and the choice probability is explicitly given in \eqref{eq:human-pref}:
\begin{align*}
\P(u \mid x^{(1)}, x^{(2)}) = \frac{\exp(f^*(u))}{\exp(f^*(x^{(1)})) + \exp(f^*(x^{(2)}))}.
\end{align*}
Under such a model, changing the ground truth $f^*$ by a constant does not change the human choice probability. Thus the value of $f^*$ is in general unidentifiable. For identifiability, we make the following assumption: 
\begin{assumption}\label{assumption:iden-rew}
Suppose the ground truth reward $f^*$ is linear as in Assumption~\ref{assumption:linear_reward}. To identify $\theta^*$ with human preference, we assume that $\sum_{i=1}^D \theta_i^* = 0$. 
\end{assumption}
In the rest of this section, we denote the space of parameter $\theta$ as $\Theta = \{\theta \in \RR^D \mid \|\theta\|_2\leq 1, \sum_{i=1}^D \theta_i = 0\}$. We remark that Assumption~\ref{assumption:iden-rew} is imposed only for technical reasons and does not compromise the generality of our analysis. Such identification assumptions are widely made in previous works on choice models \citep{zhu2023principled,aguirregabiria2010dynamic,bajari2015identification,chernozhukov2022locally,fan2023spectral}.

We learn the underlying ground truth reward function by \textit{maximum likelihood estiomation} under linear function approximation. Specifically, by the human choice model \eqref{eq:human-pref}, we apply the following minimization in Step \ref{step-rew-reg} of Algorithm \ref{alg:cdm}: \begin{align*}
\argmin_{\theta \in \Theta}~ -\frac{1}{n_2}\sum_{i=1}^{n_2} \bigg\{ \theta^\top u_i - \log\bigg(\exp( \theta^\top x_i^{(1)} ) + \exp(\theta^\top x_i^{(2)})\bigg) \bigg\}.
\end{align*} 
Then we establish the following sub-optimality guarantee with human preference.
\begin{theorem}\label{thm:rlhf-parametric}
Suppose Assumptions \ref{assumption:subspace}, \ref{assumption:gaussian_design}, \ref{assumption:linear_reward} and \ref{assumption:iden-rew} hold. We set $t_0 =\big( (Dd^2 + D^2d)/n_1 \big)^{1/6}$ and $\nu = 1/\sqrt{D}$. Then for an arbitrary $\lambda > 0 $, with high probability, running Algorithm \ref{alg:cdm} with a target reward value $a$, we have
\begin{align}\label{equ:hf-subopt}
\subopt(\hat{P}_a ; y^* = a) &
\leq\underbrace{\sqrt{\operatorname{Tr}(\tilde{\Sigma}_{\lambda}^{-1}\Sigma_{P_a})} \cdot \cO \left(\sqrt{\frac{d + \log n_2+\lambda}{n_2}} \right)}_{\cE_1:\textrm{off-policy bandit regret}} \nonumber \\
& \quad + \underbrace{{\tt DistroShift}(a) \cdot \left(d^2D + D^2 d\right)^{1/6} {n_1}^{-1/6} \cdot a}_{\cE_2:\textrm{on-support diffusion error}} + \underbrace{\EE_{\hat{P}_a} [h^*(\|\ox\|)]}_{\cE_3:\textrm{off-support diffusion error}},
\end{align}
where $\Sigma_{P_a} := \EE_{P_a}[xx^\top]$ and $\tilde{\Sigma}_{\lambda}  := \frac{1}{n_2} \big(\sum_{i=1}^{n_2} (x_i^{(1)} - x_i^{(2)})(x_i^{(1)} - x_i^{(2)})^\top + \lambda I_D\big) $.
\end{theorem}
The proof is provided in Appendix~\ref{pf:parametric}.
\paragraph{Implications and Discussions:} 
\noindent {1).}  Similar to the real-valued labeled setup, $\cE_1$ depending on $d$ shows diffusion model learns a low-dimensional representation of $x$, reducing $D$ to smaller latent dimension $d$. \re{We can prove that $\operatorname{Tr}( \tilde{\Sigma}_{\lambda}^{-1} \Sigma_{P_a})\leq \cO \left(\frac{ a^2}{ \|{\beta}^* \|_{\Sigma}} + d \right)$ when $n_2 = \Omega(\frac{1}{\lambda_{\min}})$. Such a result shows that even without an observable reward, we can still obtain an off-policy bandit regret that matches the case with an observable reward dataset. \\
\noindent {2).} By setting $t_0 =\left((Dd^2 + D^2d) / n_1\right)^{1/6}$,  we still have 
\begin{align*}
\cE_2 = \left({\tt DistroShift}(a) \cdot \left(d^2D + D^2 d\right)^{1/6} {n_1}^{-1/6} \cdot a\right),
\end{align*}
where ${\tt DistroShift}(a)$ quantifies the distribution shift depending on different reward values. In the special case of the latent covariance matrix $\Sigma$ is known, we can quantify the distribution shift as ${\tt DistroShift}(a) = \cO(a \vee d)$.
\noindent {3).} Learning from human preferences is \textit{almost as good as} learning from an explicitly labeled dataset. Specifically, our results in human preference setup match those in the setup with the observable real-valued label.}
We also claim that Algorithm \ref{alg:cdm} can provably improve the reward on the basis of dataset samples. When the latent covariance matrix $\Sigma$ is known, we can quantify the distribution shift as ${\tt DistroShift}(a) = \cO(a \vee d)$. If  $n_2 = o( n_1^{1/3})$ and the off-support diffusion error is negligible, if we denote the average reward in $\dlabel$ with $s$, by selecting $a =  O(s+\frac{d}{n_2}\sqrt{\log n_2}) $, we have $\EE_{\hat{P}_a}[f^*(x)] \geq s$ with high probability.  To the authors' best knowledge, this is the first theoretical attempt to understand reward improvement of conditional diffusion. These results imply a potential connection between diffusion theory and off-policy bandit learning, which is interesting for more future research. See proofs in Appendix \ref{pf:parametric}.

%% file: sections/experiment.tex
\section{Numerical Experiments}\label{sec:experiment}
We now present numerical results in both simulated and real data environments to showcase the performance and validate our established theories.
\subsection{Synthetic Data}
We consider a data-driven product design optimization problem. The goal is to generate new designs that attain high rewards from customers, based on pre-collected customer feedback. Let $x \in \RR^D$ denote the product design. We consider $x$ having a linear subspace structure, i.e., $x = Az$ for some unknown matrix $A$ and latent variable $z \in \RR^d$. We set $D=64$ and $d=16$. The representation matrix $A$ is randomly sampled with orthonormal columns. The latent variable $z$ follows the standard Gaussian distribution $\sf{N}(0, I_d)$. For each product design, we associate a reward function in the form of Assumption~\ref{assumption:linear_reward}. In particular, for the linear part, we generate $\beta^*$ by uniformly sampling from the unit sphere and then fix it. The penalty term is chosen as $h^*(x)=-5\Vert x\Vert_2^2$.

To implement Algorithm~\ref{alg:cdm}, we use ridge regression to estimate the reward function and train a diffusion model utilizing a $1$-dimensional version of the UNet \citep{ronneberger2015u} for learning the score function. More training configurations and code are deferred to Appendix \ref{appendix:exp}.

\begin{figure}[htb!]
    \centering
    \includegraphics[width = \textwidth]{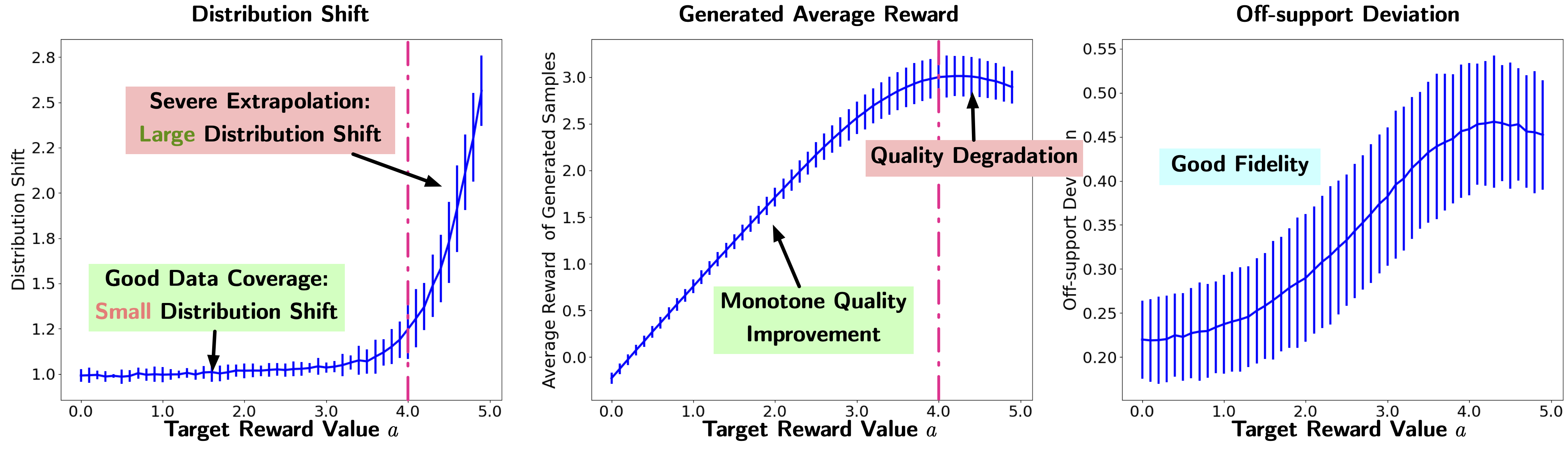}
    \caption{\textbf{Quality of generated samples as target reward value increases.}
    Left: Average reward of the generation; Middle: Distribution shift; Right: Off-support deviation. The errorbar is computed by $2$ times the standard deviation over $5$ runs.}
    \label{fig:simu_curves}
    \vspace{-12pt}
\end{figure}

\begin{figure}[htb!]
    \centering
    \includegraphics[width=0.24\linewidth]{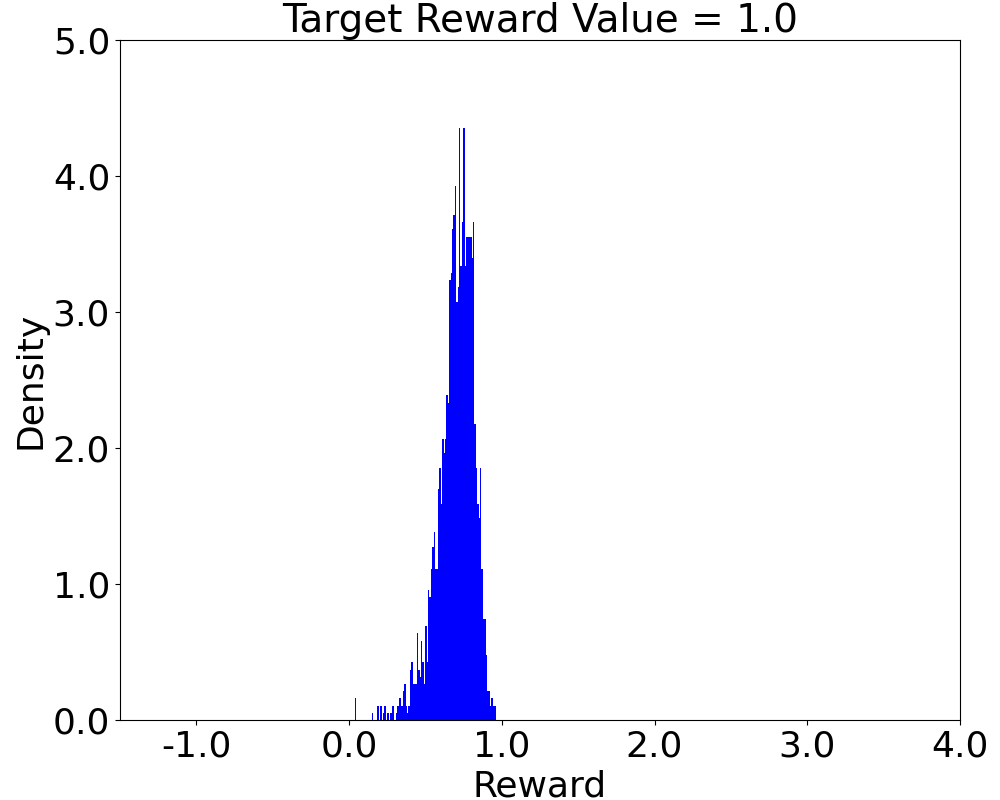}
    \includegraphics[width=0.24\linewidth]{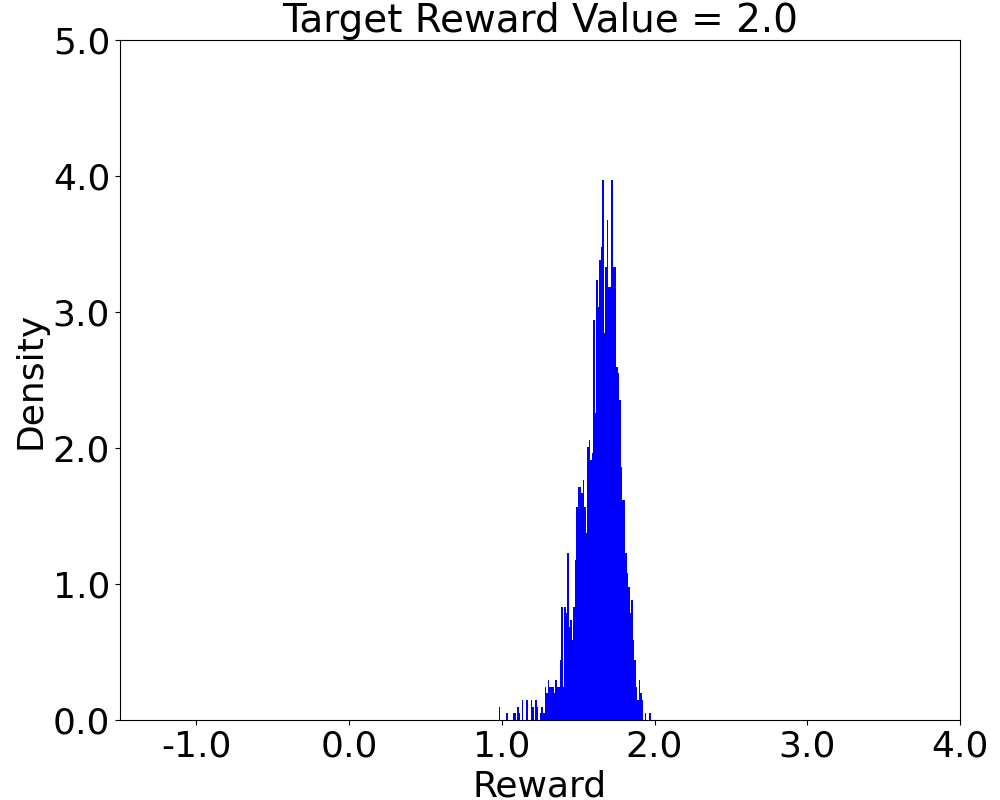}
    \includegraphics[width=0.24\linewidth]{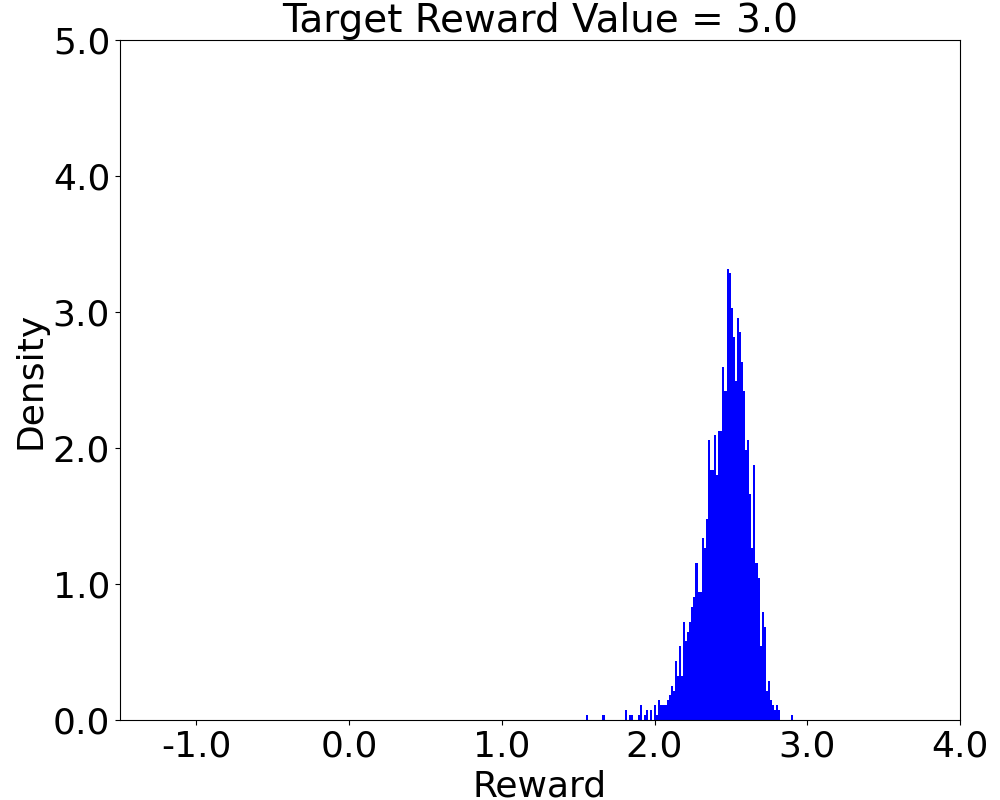}
    \includegraphics[width=0.24\linewidth]{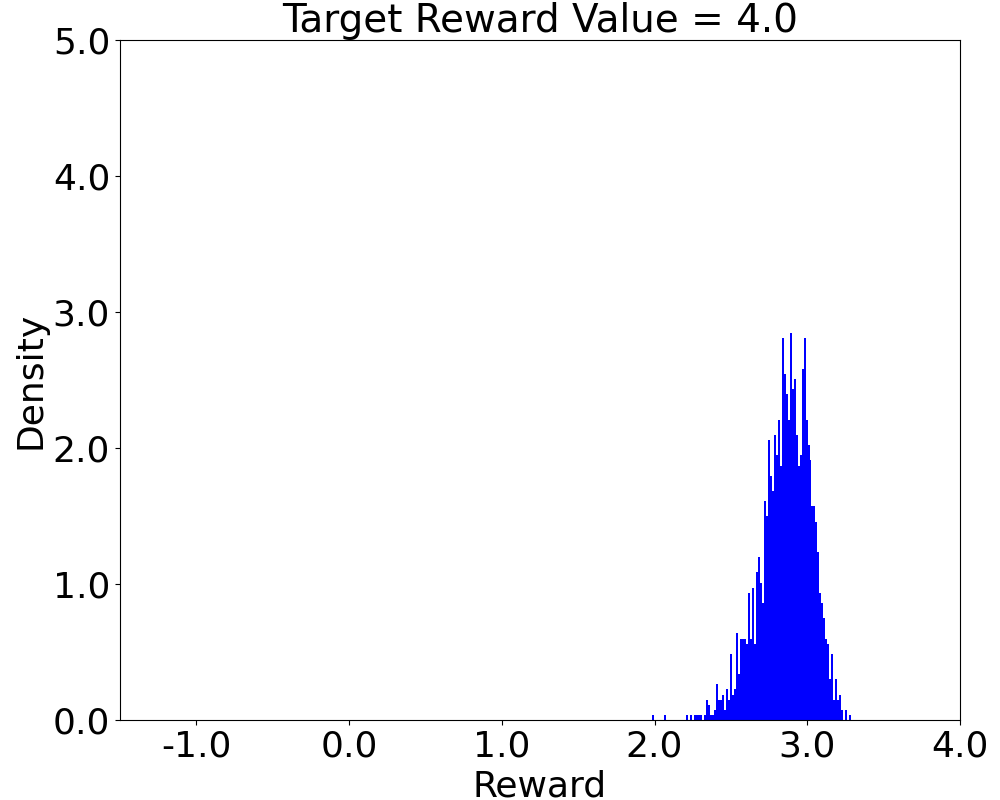}
    \caption{\textbf{Shifting reward distribution of the generated population.}}
    \label{fig:simu_dists}
    \vspace{-12pt}
\end{figure}

Figure \ref{fig:simu_curves} shows the average reward of the generated samples under different target reward values. We also plot the distribution shift and off-support deviation in terms of the $l_2$-norm distance from the support. For small target reward values, the generated average reward almost scales linearly with the target value, which is consistent with the theory as the distribution shift remains small for these target values. The generated reward begins to decrease as we further increase the target reward value, and the reason is twofold. Firstly, the off-support deviation of the generated samples becomes large in this case, which prevents the generated reward from further going up. Secondly, the distribution shift increases rapidly as we further increase the target value. In Figure \ref{fig:simu_dists}, we show the distribution of the rewards in the generated samples. As we increase the target reward value, the generated rewards become less concentrated and are shifted to the left of the target value, which is also due to the distribution shift and off-support deviation. 

\subsection{Directed Text-to-Image Generation}\label{sec:experiment:2}

Next, we empirically verify our theory through directed text-to-image generation. Instead of training a diffusion model from scratch, we use Stable Diffusion v1.5 \citep{rombach2022high}, pre-trained on LAION dataset \citep{schuhmann2022laion}. Stable Diffusion operates on the latent space of its Variational Auto-Encoder and can incorporate text conditions. We show that by training a reward model we can further guide the Stable Diffusion model to generate images of desired properties.

\textbf{Ground-truth Reward Model.} We start from an ImageNet~\citep{deng2009imagenet} pre-trained ResNet-18~\citep{he2016deep} model and replace the final prediction layer with a randomly initialized linear layer of scalar outputs. Then we use this model as the ground-truth reward model. To investigate the meaning of this randomly-generated reward model, we generate random samples and manually inspect the images with high rewards and low rewards. The ground-truth reward model seems to favor colorful and vivid natural scenes against monochrome and dull images; see Appendix~\ref{appendix:exp} for sample images.  

\textbf{Labelled Dataset.} We use the ground-truth reward model to compute a scalar output for each instance in the CIFAR-10~\citep{krizhevsky2009learning}  training dataset and perturb the output by adding a Gaussian noise from $\mathcal{N}(0,0.01)$. We use the images and the corresponding outputs as the training dataset. 

\textbf{Reward-network Training.}
To avoid adding additional input to the diffusion model and tuning the new parameters, we introduce a new network $\mu_\theta$ and approximate $p_t(y|x_t)$ by ${\sf N}(\mu_{\theta}(x_t), \sigma^2)$. For simplicity, we set $\sigma^2$ as a tunable hyperparameter. We share network parameters for different noise levels $t$, so our $\mu_\theta$ has no additional input of $t$. We train $\mu_\theta$ by minimizing the expected KL divergence between $p_t(y|x_t)$ and ${\sf N}(\mu_\theta(x_t), \sigma^2)$:
\[
   \EE_{t} \EE_{x_t}  \Big[ \mathrm{KL}(p_t(y|x_t) \mid {\sf N}(\mu_\theta(x_t), \sigma^2))  \Big]
    =   \EE_{t} \EE_{(x_t, y) \sim p_t}  \frac{\|y- \mu_\theta(x_t)\|_2^2}{2\sigma^2} + \mathrm{Constant}.
\]
Equivalently, we train the reward model $\mu_\theta$ to predict the noisy reward $y$ from the noisy inputs $x_t$. Also, notice that the minimizers of the objective do not depend on the choice of $\sigma^2$.

\textbf{Reward-network-based Directed Diffusion.} To perform reward-directed conditional diffusion, observe that $\nabla_x \log p_t(x|y) = \nabla_x \log p_t(x) + \nabla_x \log p_t(y|x)$, and $ p_t(y|x) \propto \exp \Big ( -\frac{\| y-\mu_\theta(x)\|_2^2 }{2\sigma^2} \Big)$.
Therefore, 
\[
\nabla_x \log p_t(y|x) = -1/\sigma^2  \cdot  \nabla_x \Big[ \frac12 \| y-\mu_\theta(x)\|_2^2 \Big].
\] 
In our implementation, we compute the gradient by back-propagation through $\mu_\theta$ and incorporate this gradient guidance into each denoising step of the DDIM sampler \citep{song2020denoising} following \citep{dhariwal2021diffusion} (equation (14)).
We see that $1/\sigma^2$ corresponds to the weights of the gradient with respect to unconditioned score. In the sequel, we refer to $1/\sigma^2$ as the ``guidance level'', and $y$ as the ``target value''.

\begin{figure}[htb!]
    \centering
    \begin{subfigure}[h]{0.8\textwidth}
        \includegraphics[width = 1\textwidth]{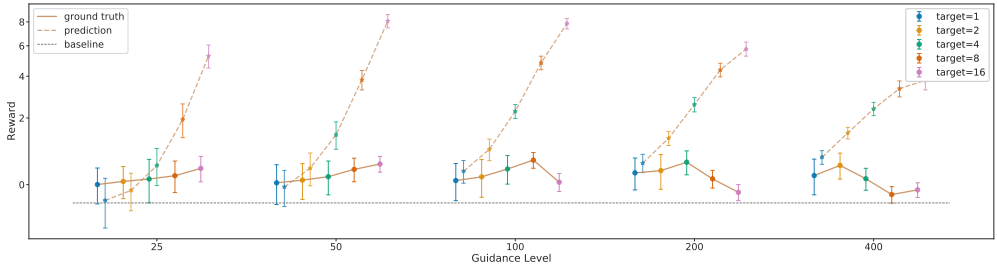}
    \end{subfigure} 
    \caption{\textbf{The predicted rewards and the ground-truth rewards of the generated images}.  At each guidance level, increasing the target $y$ successfully directs the generation towards higher predicted rewards, but also increases the error induced by the distribution shift. The reported baseline is the expected ground-truth reward for undirected generations.}
\label{fig:exp}
\end{figure}

\textbf{Quantitative Results.}  We vary $1/\sigma^2$ in $\{25, 50, 100, 200, 400\}$ and $y$ in $\{1,2,4,8,16\}$. For each combination, we generate 100 images with the text prompt ``A nice photo'' and calculate the mean and the standard variation of the predicted rewards and the ground-truth rewards. 
The results are plotted in  Figure~\ref{fig:exp}. 
From the plot, we see similar effects of increasing the target value $y$ at different guidance levels $1/\sigma^2$. A larger target value  puts more weight on the guidance signals $\nabla_x \mu_\theta(x)$, which successfully drives the generated images towards higher predicted rewards, but suffers more from the distribution-shift effects between the training distribution and the reward-conditioned distribution, which renders larger gaps between the predicted rewards and the ground-truth rewards. To optimally choose a target value, we must trade off between the two counteractive effects.

\begin{figure}[htb!]
    \centering
    \begin{subfigure}[h]{0.15\textwidth}
        \includegraphics[width = \textwidth]{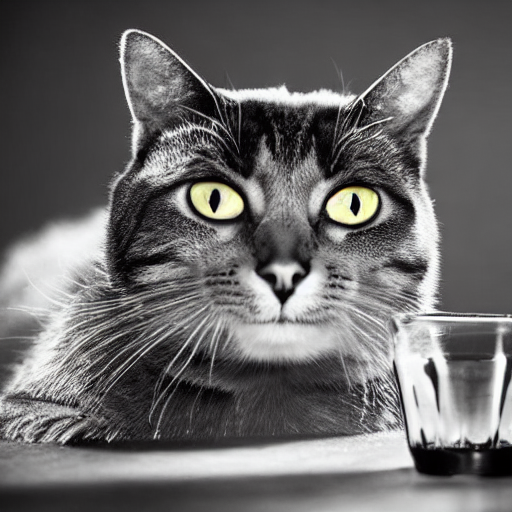}
    \end{subfigure}
    \begin{subfigure}[h]{0.15\textwidth}
        \includegraphics[width = \textwidth]{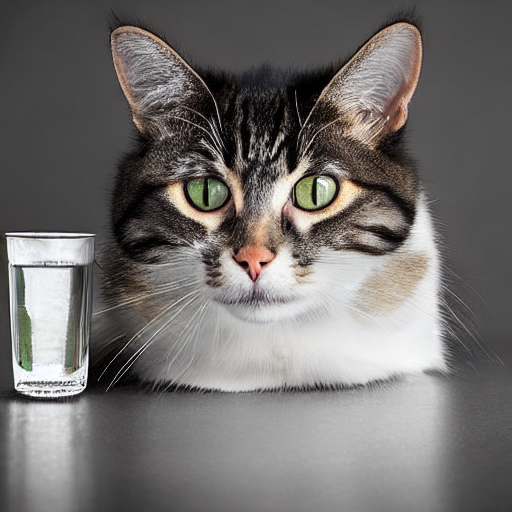}
    \end{subfigure} 
    \begin{subfigure}[h]{0.15\textwidth}
        \includegraphics[width = \textwidth]{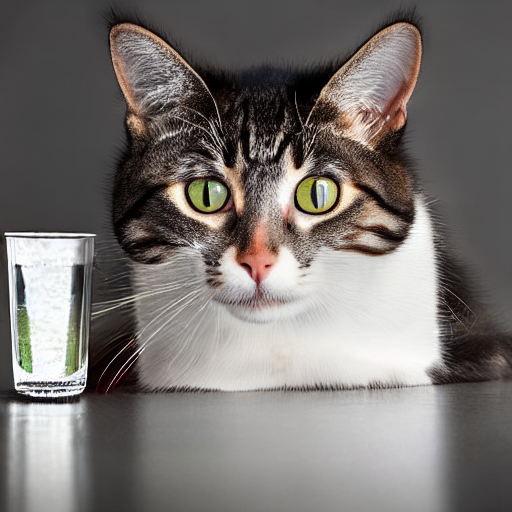}
    \end{subfigure}
    \begin{subfigure}[h]{0.15\textwidth}
        \includegraphics[width = \textwidth]{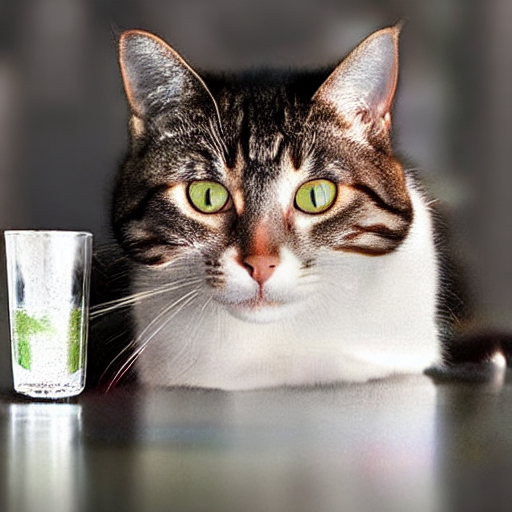}
    \end{subfigure}
    \begin{subfigure}[h]{0.15\textwidth}
        \includegraphics[width = \textwidth]{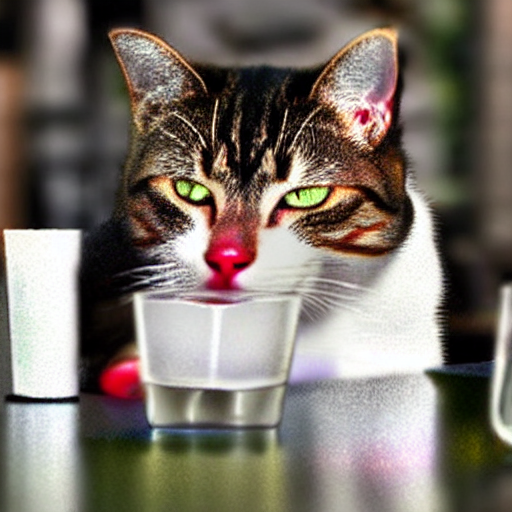}
    \end{subfigure}
    \begin{subfigure}[h]{0.15\textwidth}
        \includegraphics[width = \textwidth]{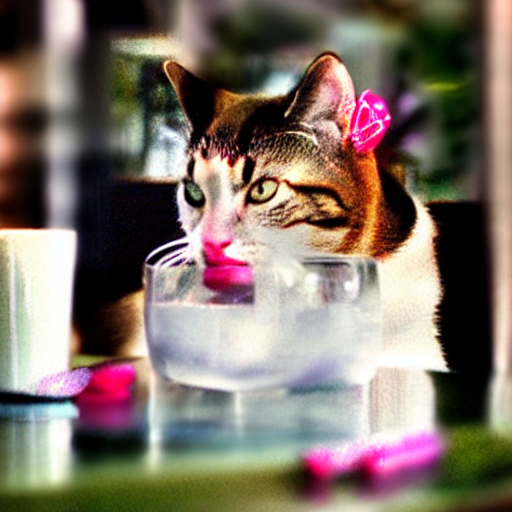}
    \end{subfigure}

    \vfill 

        \begin{subfigure}[h]{0.15\textwidth}
        \includegraphics[width = \textwidth]{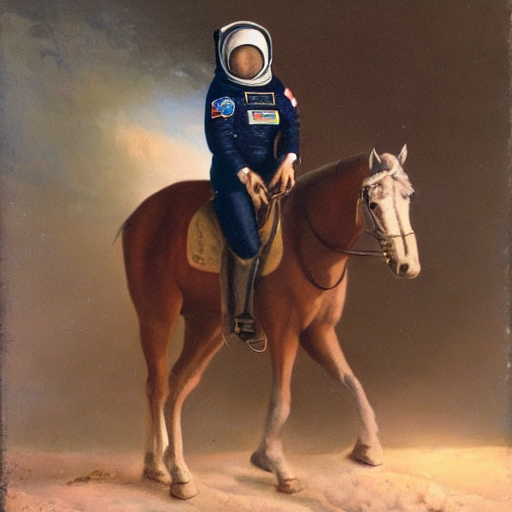}
    \end{subfigure}
    \begin{subfigure}[h]{0.15\textwidth}
        \includegraphics[width = \textwidth]{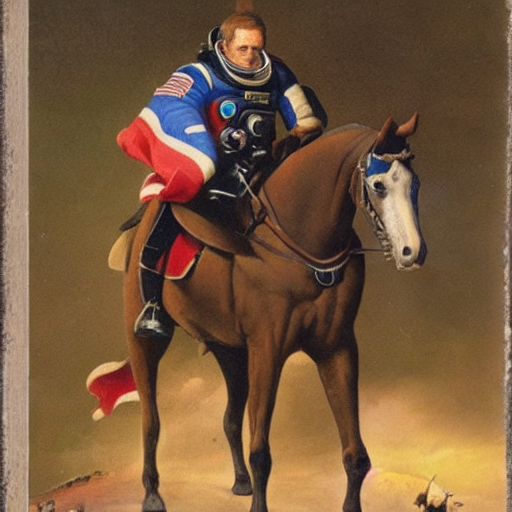}
    \end{subfigure} 
    \begin{subfigure}[h]{0.15\textwidth}
        \includegraphics[width = \textwidth]{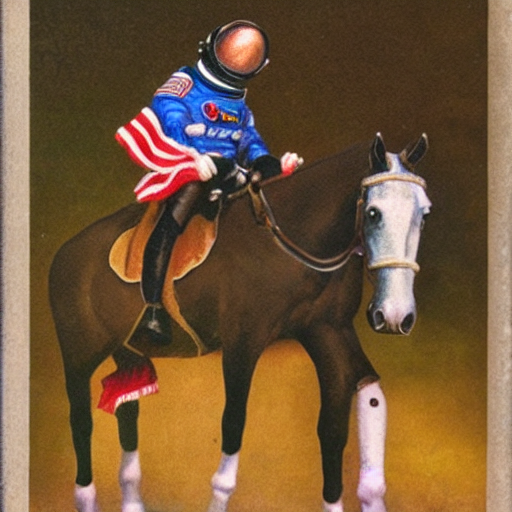}
    \end{subfigure}
    \begin{subfigure}[h]{0.15\textwidth}
        \includegraphics[width = \textwidth]{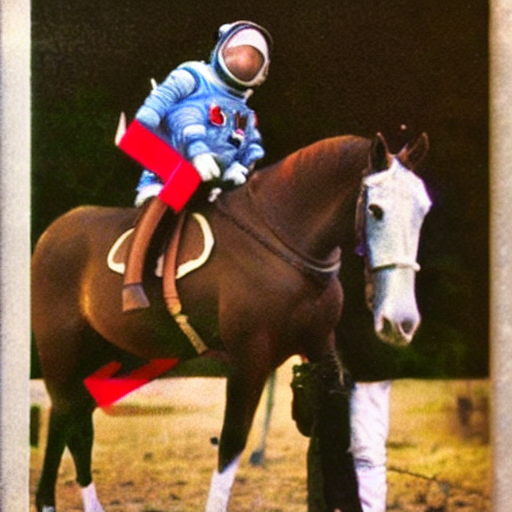}
    \end{subfigure}
    \begin{subfigure}[h]{0.15\textwidth}
        \includegraphics[width = \textwidth]{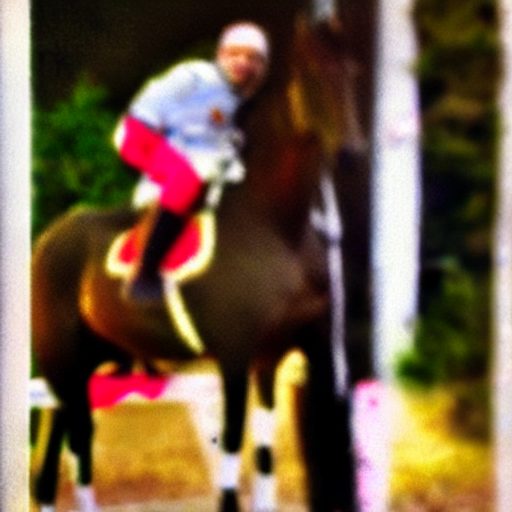}
    \end{subfigure}
    \begin{subfigure}[h]{0.15\textwidth}
        \includegraphics[width = \textwidth]{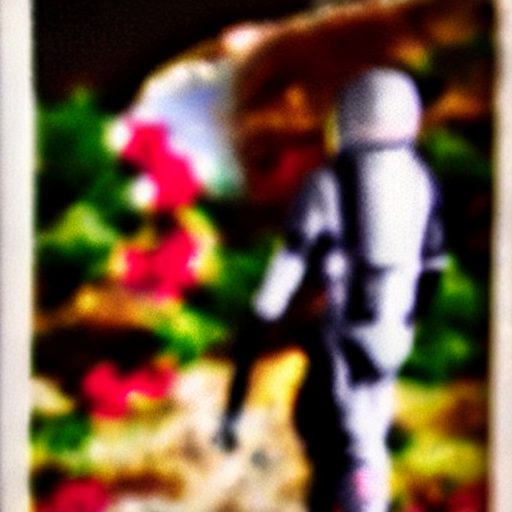}
    \end{subfigure}
     \vspace*{3mm}
    \includegraphics[width = 0.95\textwidth]{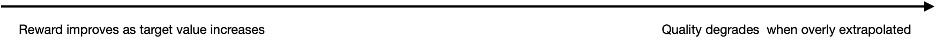}
    \caption{\textbf{The effects of the reward-directed diffusion.} Increasing the target value directs the images to be more colorful and vivid at the cost of degradation of the image qualities. Leftmost: without reward conditioning. Second-to-Last: target value $y = 2, 4, 6, 8, 10$. The guidance level $1/\sigma^2$ is fixed to $100$. The text prompts are ``A cat with a glass of water'', ``An astronaut on the horseback''.}
\label{fig:demo}
\end{figure}

\textbf{Qualitative Results.} To qualitatively test the effects of the reward conditioning, we generate a set of images with increasing target values $y$ under different text prompts and investigate the visual properties of the produced images. 
We isolate the effect of reward conditioning by fixing all the randomness during the generation processes, so the generated images have similar semantic layouts. 
After hyper-parameter tuning, we find that setting $1/\sigma^2=100$ and $y \in \{2,4,6,8,10\}$ achieves good results across different text prompts and random seeds. We pick out typical examples and summarized the results in Figure~\ref{fig:demo}, which demonstrates that as we increase the target value, the generated images become more colorful at the expense of degradations of the image qualities. 

\subsection{Offline High-Reward Trajectory Generation in Reinforcement Learning}
\begin{figure}[htb!]
    \centering
    \includegraphics[width=\textwidth]{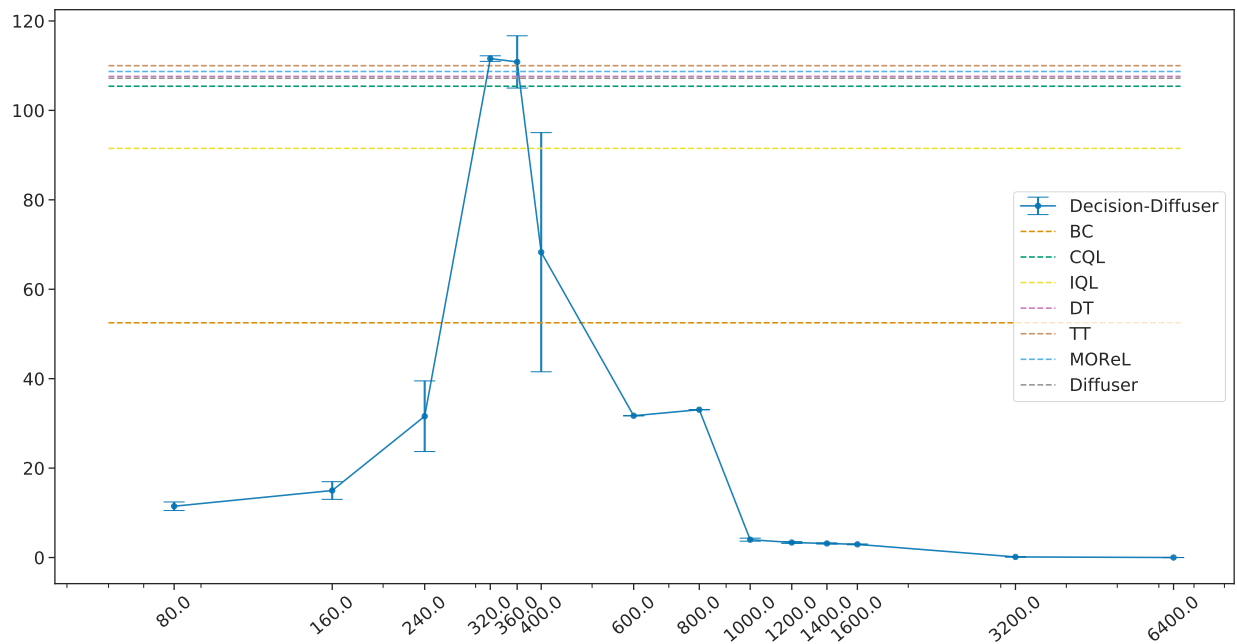} 
    \caption{{Total Reward v.s. Target Reward for Decision Diffuser in the (Med-Expert, Hopper) setting}.
}
    \label{fig:dd} 
\end{figure}

Lastly, we demonstrate using our proposed reward-directed conditional diffusion model for direct high-reward trajectory generation in reinforcement learning problems. We replicate the implementation of Decision Diffuser~\citep{ajay2023is} under the (Med-Expert, Hopper) setting. In Decision Diffuser, the RL trajectory and the final reward are jointly modeled with a conditioned diffusion model. The policy is given by first performing reward-directed conditional diffusion to sample a trajectory of high reward and then extracting action sequences from the trajectory using a trained inverse dynamics model. We plot the mean and standard deviation of the total returns (averaged across 10 independent episodes) v.s. the target rewards in Figure~\ref{fig:dd}. The theoretical maximum of the reward is 400. Therefore, trajectories with rewards greater than 400 are never seen during training. We observe that when we increase the target reward beyond 320, the actual total reward decreases. This is supported by our theory that as we increase the reward guidance signal, the condition effect becomes stronger but the distribution-shift effect also becomes stronger. We also observe that the reward-directed conditional generation yields comparable and even marginally better performance compared to state-of-the-art methods, such as trajectory transformer \citep{janner2021offline} and MoReL \citep{kidambi2020morel}.

%% file: appendix/appdx.tex
\input{sections/tem_proofs}

\section{Supporting Lemmas and Proofs}
\label{spt_lemmas}
\subsection{Supporting Lemmas}
\begin{lemma}
\label{lmm:VU_A}
The estimated subspace $V$ satisfies
\begin{equation}
    \|VU - A\|_{F} = \cO \left(d^{\frac{3}{2}} \sqrt{\subangle{V}{A}}\right)
\end{equation}
for some orthogonal matrix $U \in \RR^{d \times d}$.
\end{lemma}

Proof of Lemma \ref{lmm:VU_A} is in Appendix \ref{sec:proof_VU_A}.

\begin{lemma}
\label{lmm:exp_by_tv}
Suppose $P_1$ and $P_2$ are two distributions over $\mathbb{R}^d$ and $m$ is a function defined on $\mathbb{R}^d$, then $\left| \EE_{x \sim P_1}[m(z)] - \EE_{z \sim P_2}[m(z)] \right|$can be bounded in terms of  $\dtv(P_1, P_2)$, specifically when $P_1$ and $P_2$ are Gaussians and $m(z) = \|z\|^2_2$:
\begin{equation}
    \EE_{x \sim P_1}[\|z\|^2_2] = \cO \left(\EE_{z \sim P_2}[\|z\|^2_2] (1+ \dtv(P_1, P_2))\right).
\end{equation}

When $P_1$ and $P_2$ are Gaussians and $m(z) = \|z\|_2$:
\begin{equation}
    \left| \EE_{z \sim P_1}[\|z\|_2] - \EE_{z \sim P_2}[\|z\|_2] \right| = \cO \left( \left(\sqrt{ \EE_{z \sim P_1} [\|z\|_2^2]} +  \sqrt{ \EE_{z \sim P_2} [\|z\|_2^2]} \right) \cdot \dtv(P_1, P_2) \right).
\end{equation}
\end{lemma}

Proof of Lemma \ref{lmm:exp_by_tv} is in Appendix \ref{sec:exp_by_tv}.

\begin{lemma}
\label{lmm:exp_x_norm}
We compute $\EE_{z \sim P^{LD}(a) } [\left\| z\right\|_2^2], \EE_{z \sim P^{LD}_{t_0}(a)} [ \|z\|_2^2] , \EE_{x \sim \hat{P}_a} [\|x\|_2^2], \EE_{z \sim (U^{\top}V^{\top})_{\#}\hat{P}_a} [\|z\|_2^2]$ in this Lemma.
\begin{equation}
    \EE_{z \sim P^{LD}(a) } [\left\| z\right\|_2^2] = \frac{\hat{\beta}^{\top} \Sigma^2 \hat{\beta}}{\left( \|\hat{\beta}\|_{\Sigma}^2 + \nu^2 \right)^2} a^{2}  + \operatorname{trace}(\Sigma - \Sigma \hat{\beta} \left( \hat{\beta}^{\top} \Sigma \hat{\beta} + \nu^2 \right)^{-1} \hat{\beta}^{\top} \Sigma).
\end{equation}
Let $M(a):= \EE_{z \sim P^{LD}(a) } [\left\| z\right\|^2_2]$, which has an upper bound $M(a) = O \left(  \frac{a^2}{\| {\beta}^* \|_{\Sigma}} + d \right)$.
\begin{align}
    &\EE_{z \sim P^{LD}_{t_0}(a)} [ \|z\|^2_2] \leq  M(a) + t_0 d.\\
    \label{equ:x_norm}&\EE_{x \sim \hat{P}_a} \left[ \|x\|^2_2 \right] \leq \cO \left( ct_0D + M(a) \cdot  (1+ TV(\hat{P}_a) \right).
\end{align}
\end{lemma}

Proof of Lemma \ref{lmm:exp_x_norm} is in Appendix \ref{sec:proof_exp_x_norm}.

\subsection{Proof of Lemma \ref{lmm:diff_results}}
\label{pf:diff_results}
The first two assertions \eqref{equ:orth_dtb} and \eqref{equ:subspace_rcv} are consequences of \citet[Theorem 3, item 1 and 3]{chen2023score}. To show \eqref{equ:tv_Pq}, we first have the conditional score matching error under distribution shift being
\begin{align*}
\cT(P(x, \hat{y} = a), P_{x\hat{y}}; \bar{\cS}) \cdot \epsilon_{diff}^2,
\end{align*}
where $\cT(P(x, \hat{y} = a), P_{x\hat{y}}; \bar{\cS})$ accounts for the distribution shift as in the parametric case (Lemma~\ref{lmm:E2}). Then we apply \citet[Theorem 3, item 2]{chen2023score} to conclude
\begin{align*}
TV(\hat{P}_a) = \tilde{\cO}\left(\sqrt{\frac{\cT(P(x, \hat{y} = a), P_{x\hat{y}}; \bar{\cS})}{c_0}} \cdot \epsilon_{diff} \right).
\end{align*}
The proof is complete.

\subsection{Proof of Lemma \ref{lmm:reg_err}}
\label{pf:reg_err}
Given
\begin{align*}
    \cE_1 &= \EE_{\hat{P}_a} \left| x^{\top} (\theta^* - \hat{\theta})\right| \leq \EE_{\hat{P}_a} \|x\|_{V_{\lambda}^{-1}} \cdot \|\theta^* - \hat{\theta}\|_{V_{\lambda}},
\end{align*}
then things to prove are 
\begin{align}
    \label{reg:d_shift}
    \EE_{\hat{P}_a} \|x\|_{V_{\lambda}^{-1}} &= \sqrt{\operatorname{trace}( V_{\lambda}^{-1} \Sigma_{\hat{P}_a})};\\
    \label{reg:e_loss}\|\theta^* - \hat{\theta}\|_{V_{\lambda}} &\leq \cO \left( \sqrt{d \log n_2 } \right),
\end{align}
where the second inequality is to be proven with high probability w.r.t the randomness in $\cD_{label}$.
For \eqref{reg:d_shift}, $\EE_{\hat{P}_a} \|x\|_{V_{\lambda}^{-1}} \leq \sqrt{\EE_{\hat{P}_a} x^{\top} V_{\lambda}^{-1} x} = \sqrt{\EE_{\hat{P}_a} \operatorname{trace}( V_{\lambda}^{-1} x x^{\top}) } = \sqrt{ \operatorname{trace}( V_{\lambda}^{-1}\EE_{\hat{P}_a} x x^{\top}) }.$

For \eqref{reg:e_loss}, what's new to prove compared to a classic bandit derivation is its $d$ dependency instead of $D$, due to the linear subspace structure in $x$. From the closed form solution of $\hat \theta$, we have
\begin{equation}
    \hat{\theta} - \theta^* = V_{\lambda}^{-1} X^{\top} \eta - \lambda V_{\lambda}^{-1} \theta^*.
\end{equation}
Therefore,
\begin{equation}
    \|\theta^* - \hat{\theta}\|_{V_{\lambda}} \leq \|X^{\top} \eta \|_{V_{\lambda}^{-1}}  + \lambda\| \theta^*\|_{V_{\lambda}^{-1}},  
\end{equation}
where $\lambda\| \theta^*\|_{V_{\lambda}^{-1}} \leq \sqrt{\lambda} \|\theta^*\|_2 \leq \sqrt{\lambda}$ and 
\begin{align*}
    \|X^{\top} \eta \|_{V_{\lambda}^{-1}}^2 &=\eta^{\top} X \left( X^{\top} X + \lambda I_D \right)^{-1} X^{\top} \eta \\
    &= \eta^{\top} X X^{\top} \left( X X^{\top}  + \lambda I_{n_2} \right)^{-1}  \eta.
\end{align*}
Let $Z^{\top} = (z_1, \cdots, z_i, \cdots, z_{n_2})$ s.t. $A z_i = x_i$, then it holds that $X = Z A^{\top}$, and $X X^{\top} = Z A^{\top} A Z^{\top} = Z Z^{\top} $ , thus
\begin{align*}
    \|X^{\top} \eta \|_{V_{\lambda}^{-1}}^2
    &= \eta^{\top} X X^{\top} \left( X X^{\top}  + \lambda I_{n_2} \right)^{-1}  \eta\\
    &= \eta^{\top} Z Z^{\top} \left( Z Z^{\top}  + \lambda I_{n_2} \right)^{-1}  \eta\\
    &= \eta^{\top} Z  \left(  Z^{\top} Z  + \lambda I_{d} \right)^{-1} Z^{\top} \eta\\
    &= \|Z^{\top} \eta \|_{\left(  Z^{\top} Z  + \lambda I_{d} \right)^{-1}}.
\end{align*}
With probability $1-\delta$, $\|z_i\|^2 \leq d+ \sqrt{d \log\left( \frac{2 n_2}{\delta}\right)}:= L^2, \forall i \in [n_2]$. Then applying \citet[Theorem 1]{abbasi2011improved} gives rise to
\begin{equation*} 
    \|Z^{\top} \eta \|_{\left(  Z^{\top} Z  + \lambda I_{d} \right)^{-1}} \leq \sqrt{2 \log (2 / \delta)+d \log (1+n_2 L^2 /(\lambda d))}
\end{equation*}
with probability $1-\delta/2$. Combine things together and plugging in $\lambda = 1$, $L^2 = d+ \sqrt{d \log\left( \frac{2 n_2}{\delta}\right)}$, we have with high probability 
$$
\|\theta^* - \hat{\theta}\|_{V_{\lambda}}  = \cO \left( \sqrt{d \log \left(n_2 \sqrt{\log(n_2)} \right)}\right) = \cO \left( \sqrt{d \log n_2 + \frac{1}{2} d \log( \log n_2)} \right) = \cO \left( \sqrt{d \log n_2 } \right).
$$

\subsection{Proof of Lemma \ref{lmm:distribution_sft}}
\label{pf:distribution_sft}
Recall the definition of $\hat{\Sigma}_{\lambda}$ and $\Sigma_{P_a}$ that
\begin{align*}
    \hat{\Sigma}_{\lambda} &= \frac{1}{n_2} X^{\top}X + \frac{\lambda}{n_2} I_D,\\
    \Sigma_{P_a} &= \mathbb{E}_{x \sim P_a} \left[ x x^{\top}\right],
\end{align*}
where $X$ are stack matrix of data supported on $\mathcal{A}$ and $P_a$ is also supported on $\cA$, $\cA$ is the subspace encoded by matrix $A$. The following lemma shows it is equivalent to measure $\operatorname{trace}( \hat{\Sigma}_{\lambda}^{-1} \Sigma_{P_a})$ on $\cA$ subspace. 

\begin{lemma}
\label{lmm:XtoZ}
    For any P.S.D. matrices $\Sigma_1,\Sigma_2\in\mathbb{R}^{d\times d}$ and $A\in\mathbb{R}^{D\times d}$ such that $A^\top A=I_d$, we have
    \begin{align*}
        \mathrm{Tr}\left((\lambda I_D+A\Sigma_1 A^\top)^{-1}A\Sigma_2A^\top\right)=\mathrm{Tr}\left(\left(\lambda I_d+\Sigma_1\right)^{-1}\Sigma_2\right).
    \end{align*}
\end{lemma}

The lemma above allows us to abuse notations $\hat{\Sigma}_{\lambda}$ and $\Sigma_{P_a}$ in the following way while keeping the same $\operatorname{trace}( \hat{\Sigma}_{\lambda}^{-1} \Sigma_{P_a})$ value:
\begin{align*}
    \hat{\Sigma}_{\lambda} &= \frac{1}{n_2} Z^{\top}Z + \frac{\lambda}{n_2} I_d,\\
    \Sigma_{P_a} &= \mathbb{E}_{z \sim P^{LD}(a)} \left[ z z^{\top}\right],
\end{align*}
where $Z^{\top} = (z_1, \cdots, z_i, \cdots, z_{n_2})$ s.t. $A z_i = x_i$ and recall notataion $P^{LD}(a) = P_z\left(z \mid \hat{f}(Az) \right)$. 

Given $z \sim {\sf N} (\mu, \Sigma)$, as a proof artifact, let $\hat f (x) = \hat{\theta}^{\top} x + \xi, \xi \sim {\sf N} (0, \nu^2)$ where we will let $\nu \to 0$ in the end, then let $\hat{\beta} = A^{\top} \hat{\theta} \in \mathbb{R}^{d}$, $(z, \hat f(Az))$  has a joint distribution
\begin{equation}
    (z, \hat f) \sim {\sf N} \left(
    \begin{bmatrix}
        \mu \\
        \hat{\beta}^{\top} \mu
    \end{bmatrix},  \begin{bmatrix}
        \Sigma & \Sigma \hat{\beta}  \\
        \hat{\beta}^{\top} \Sigma & \hat{\beta}^{\top} \Sigma \hat{\beta} + \nu^2
    \end{bmatrix}\right).
\end{equation}
Then we have the conditional distribution $z \mid \hat{f}(Az) = a$ following
\begin{equation}
\label{equ:z|hat_f}
    P_z \left( z \mid \hat{f}(Az) = a \right) = {\sf N} \left( \mu+ \Sigma \hat{\beta} \left( \hat{\beta}^{\top} \Sigma \hat{\beta} + \nu^2 \right)^{-1} (a - \hat{\beta}^{\top} \mu), \Gamma \right)
\end{equation}
with $\Gamma:= \Sigma - \Sigma \hat{\beta} \left( \hat{\beta}^{\top} \Sigma \hat{\beta} + \nu^2 \right)^{-1} \hat{\beta}^{\top} \Sigma$.

When $\mu = 0$, we compute $\operatorname{trace}( \hat{\Sigma}_{\lambda}^{-1} \Sigma_{P_a})$ as

\begin{align*}
    \operatorname{trace}( \hat{\Sigma}_{\lambda}^{-1} \Sigma_{P_a}) &= \operatorname{trace} \left( \hat{\Sigma}_{\lambda}^{-1}  \frac{\Sigma \hat{\beta}   \hat{\beta}^{\top} \Sigma}{\left( \|\hat{\beta}\|_{\Sigma}^2 + \nu^2 \right)^2} a^{2} \right) +  \operatorname{trace} \left(\hat{\Sigma}_{\lambda}^{-1} \Gamma \right)\\
    &= \operatorname{trace} \left(   \frac{\hat{\beta}^{\top} \Sigma \hat{\Sigma}_{\lambda}^{-1} \Sigma \hat{\beta}    }{\left( \|\hat{\beta}\|_{\Sigma}^2 + \nu^2 \right)^2} a^{2} \right) +  \operatorname{trace} \left(\hat{\Sigma}_{\lambda}^{-1} \Sigma \right) - \operatorname{trace} \left(\hat{\Sigma}_{\lambda}^{-1}  \frac{\Sigma \hat{\beta} \hat{\beta}^{\top} \Sigma}{\|\hat{\beta}\|_{\Sigma}^2 + \nu^2}\right)\\
    & = \operatorname{trace} \left(   \frac{\Sigma^{1/2}\hat{\beta}\hat{\beta}^{\top} \Sigma^{1/2} \Sigma^{1/2} \hat{\Sigma}_{\lambda}^{-1} \Sigma^{1/2}  }{\left( \|\hat{\beta}\|_{\Sigma}^2 + \nu^2 \right)^2} a^{2} \right) \\
    &\leq    \frac{\|\Sigma \hat{\Sigma}_{\lambda}^{-1} \Sigma\|_{op} \cdot \|\hat{\beta}\|_{\Sigma}^2 }{\left( \|\hat{\beta}\|_{\Sigma}^2 + \nu^2 \right)^2} \cdot a^{2} +  \operatorname{trace} \left(\Sigma^{\frac{1}{2}} \hat{\Sigma}_{\lambda}^{-1} \Sigma^{\frac{1}{2}} \right)
\end{align*}

By Lemma 3 in \cite{chen2020towards}, it holds that
\begin{equation}
    \|\Sigma^{\frac{1}{2}} \hat{\Sigma}_{\lambda}^{-1} \Sigma^{\frac{1}{2}} - I_{d}\|_{2} \leq O \left( \frac{1}{ \sqrt{\lambda_{\min}n_2 } }\right).
\end{equation}

Therefore,
\begin{align*}
    \operatorname{trace}( \hat{\Sigma}_{\lambda}^{-1} \Sigma_{P_a}) 
    &\leq  \frac{ 1 + \frac{1}{ \sqrt{\lambda_{\min}n_2 }} }{ \|\hat{\beta}\|_{\Sigma}^2} \cdot a^{2} + O \left( d \left( 1 + \frac{1}{ \sqrt{\lambda_{\min}n_2 } }\right) \right).
\end{align*}

Then, what left is to bound $\| \hat{\beta} \|_{\Sigma} = \| \hat{\theta} \|_{A \Sigma A^{\top}} \geq \| \theta^* \|_{A \Sigma A^{\top}} - \| \hat{\theta} - \theta^* \|_{A \Sigma A^{\top}}$ by triangle inequality. On one hand,
\begin{equation}
    \| \theta^* \|_{A \Sigma A^{\top}} = \| \beta^* \|_{\Sigma}.
\end{equation}
On the other hand,
\begin{align*}
    \| \hat{\theta} - \theta^*\|_{A \Sigma A^{\top}} &= \cO \left( \| \hat{\theta} - \theta^*\|_{\hat{\Sigma}_{\lambda}} \right)\\
    &= \cO \left( \frac{\| \hat{\theta} - \theta^*\|_{V_{\lambda}} }{\sqrt{n_2}}\right) \\
    &=  \cO \left( \sqrt{\frac{d \log(n_2)}{n_2}}\right).
\end{align*}
with high probability. Thus when $n_2 = \Omega (\frac{d}{\|\beta^*\|^2_{\Sigma}})$ 
\begin{equation*}
    \|\hat{\beta}\|_{\Sigma} \geq \frac{1}{2}\| \beta^* \|_{\Sigma}.
\end{equation*}

Therefore
\begin{equation}
    \operatorname{trace}( \hat{\Sigma}_{\lambda}^{-1} \Sigma_{P_a}) 
    \leq  \cO \left( \frac{ 1 + \frac{1}{ \sqrt{\lambda_{\min} n_2 }} }{ \| \beta^* \|_{\Sigma}} \cdot a^{2} + d \left( 1 + \frac{1}{ \sqrt{\lambda_{\min}n_2 } }\right) \right)\\
    = \cO \left( \frac{ a^2 }{ \| \beta^* \|_{\Sigma}} + d \right).
\end{equation}
when $n_2 = \Omega ( \max \{\frac{1}{\lambda_{\min}}, \frac{d}{\|\beta^*\|^2_{\Sigma}} \})$.

\subsection{Proof of Lemma \ref{thm:rew-hf-est}}\label{pf:mle}
\re{Denote the loss function in Equation \eqref{eq:loss-func} by $l(\theta)$.
By simple algebra we have \begin{align*}
        &\nabla l(\theta) = \frac{1}{n_2}\sum_{i=1}^{n_2} \bigg\{ \frac{\exp(\theta^\top x_i^{(1)}) \cdot ( x_i^{(1)} - u_i) + \exp(\theta^\top x_i^{(2)})\cdot (x_i^{(2)} - h_i)}{\exp(\theta^\top x_i^{(1)}) + \exp(\theta^\top x_i^{(2)})}\bigg\} ,\\
        & \nabla^2 l(\theta) = \frac{1}{n_2}\sum_{i=1}^{n_2} \bigg\{\frac{\exp((x_i^{(1)} + x_i^{(2)})^\top \theta)}{(\exp(\theta^\top x_i^{(1)})+ \exp(\theta^\top x_i^{(2)}))^2}\cdot (x_i^{(1)} - x_i^{(2)})(x_i^{(1)} - x_i^{(2)})^\top\bigg\},
    \end{align*}
    holds for all $\theta\in \RR^d$. By Assumption \ref{assumption:gaussian_design} and Hoeffding's inequality, we have  $$\PP\big(\max_{i\in[n_2]}\big\{| \theta^{*\top} x_i^{(1)}|, |\theta^{*\top} x_i^{(2)}|\big\} \leq \log(2n_2 / \delta) \big) \geq 1-\delta.
    $$
    In the following proof, we will condition on this event. We have \begin{align*}
    \nabla^2 l(\theta^*) &= \frac{1}{n_2} \sum_{i=1}^{n_2}\bigg(\frac{1}{1+ \exp((x_i^{(1)} - x_i^{(2)})^\top \theta^*)}\cdot\frac{1}{1+ \exp(( x_i^{(2)} - x_i^{(1)})^\top \theta^*)} \bigg)\cdot (x_i^{(1)} - x_i^{(2)})(x_i^{(1)} - x_i^{(2)})^\top\\
    &\succeq \frac{1}{n_2} \cdot \frac{1}{(1+\exp(\sqrt{\log 2n_2/\delta}))^2} \sum_{i=1}^n (x_i^{(1)} - x_i^{(2)})(x_i^{(1)} - x_i^{(2)})^\top.
    \end{align*}
    In the following we adapt the notation of $\Sigma_x  = \frac{1}{n} \sum_{i=1}^n (x_i^{(1)} - x_i^{(2)})(x_i^{(1)} - x_i^{(2)})^\top$, then $\nabla^2 l(\theta^*) \succeq 1/\kappa\cdot \Sigma_x$. To bound $\EE_{x\sim P_a}(|\hat{f}(x) - f^*(x)|)$, we have the following inequality: \begin{align*}
        \EE_{x\sim P_a}[|\hat{f}(x) - f^*(x)|] &= \EE_{x\sim P_a}[|x^\top(\hat{\theta} - \theta^*)|]\\
        &= \EE_{z\sim( A^\top)_\sharp P_a}[|z^\top (A^\top \widehat{ \theta} - \beta)|]\\
        &\leq \underbrace{\EE_{ z\sim( A^\top)_\sharp P_a} [\|z\|_{(\Sigma_z + \lambda I_d)^{-1}}]}_{\operatorname{(i)}} \cdot \underbrace{\|\Delta_z\|_{(\Sigma_z + \lambda I_d)^{-1}}}_{\operatorname{(ii)}}
    \end{align*}
    Here $\Delta_z = A^\top \widehat{\theta} - \beta$. We also denote $\Delta_x = \widehat{\theta} - \theta^*$. To bound (i) note that by Jensen's inequality, we have $$
    \EE_{ z\sim( A^\top)_\sharp P_a} [\|z\|_{(\Sigma_z + \lambda I_d)^{-1}}] \leq \sqrt{\EE_{ z\sim( A^\top)_\sharp P_a}\big[\|z\|^2_{(\Sigma_z + \lambda I_d)^{-1}}\big]} = \sqrt{\operatorname{Tr}\big(\EE_{z\sim( A^\top)_\sharp P_a}[zz^\top]\big) (\Sigma_z + \lambda I_d)^{-1}}.
    $$
    Next we bound (ii). It is easy to prove that $$
    l(\widehat{\theta}) - l(\theta^*) - \nabla l(\theta^*)^\top (\widehat{\theta} - \theta^*) \geq \frac{1}{2\kappa}\|\Delta_x\|_{\Sigma_x}^2.
    $$
By definition of MLE, we have $l(\widehat{\theta}) - l(\theta) \leq 0$. Therefore \begin{align}\label{eq:weight-convex}
    \frac{1}{2\kappa} \|\Delta_x\|_{\Sigma_x}^2 &\leq |\nabla l(\theta^*)^\top (\widehat{\theta} - \theta^*)|\nonumber\\
    & \leq |\nabla l(\theta^*)^\top (A A^\top \widehat{\theta} - \theta^*)| + |\nabla l(\theta^*)^\top(I - A A^\top) (\widehat{\theta} - \theta^*)|.
\end{align}
Recall that $$\nabla l(\theta^*) =  \frac{1}{n_2}\sum_{i=1}^{n_2}  \frac{\exp(\theta^\top x_i^{(1)}) \cdot (x_i^{(1)} -u_i) + \exp(\theta^\top x_i^{(2)})\cdot (x_i^{(2)} -u_i)}{\exp(\theta^\top x_i^{(1)}) + \exp(\theta^\top x_i^{(2)})}, $$ 
since $x_i^{(1)}, x_i^{(2)} $ are in the image set of $A$,  there exists $\alpha\in \RR^d$ such that $\nabla l(\theta^*) = A\alpha$, and we have $$
\nabla l(\theta^*)^\top (I - AA^\top) = \alpha^\top A^\top (I - AA^\top) = 0.
$$
Therefore we have \begin{align}\label{eq:tech-1}
    \frac{1}{2\kappa} \|\Delta_x\|_{\Sigma_x}^2 &\leq |\nabla l(\theta^*)^\top (AA^\top \widehat{\theta} - \theta^* )|\nonumber\\
    & = |(A^\top \nabla l(\theta^*) )^\top \Delta_z |\nonumber\\
    &\leq \|A^\top \nabla l(\theta^*)\|_{(\Sigma_z + \lambda I_d)^{-1}}\cdot \|\Delta_z\|_{\Sigma_z + \lambda I_d}
\end{align}
By $A A^\top (x_i^{(1)} - x_i^{(2)}) = (x_i^{(1)} - x_i^{(2)}) $,  we have \begin{align*}
\|\Delta_x\|_{\Sigma_x}^2 = \Delta_x^\top A \Sigma_z A^\top \Delta_x = \Delta_z^\top \Sigma_z \Delta_z = \|\Delta_z\|_{\Sigma_z}^2,\end{align*}
plug this in \eqref{eq:tech-1}, we have 
\begin{align*}
    \frac{1}{2\kappa} \|\Delta_z\|_{\Sigma_z+\lambda I_d}^2 & \leq \frac{1}{2\kappa} (\|\Delta_z\|_{\Sigma_z}^2 + \lambda \|\Delta_z\|_2^2)\\
    &\leq \|A^\top \nabla l(\theta^*)\|_{(\Sigma_z + \lambda I_d)^{-1}}\cdot\|\Delta_z\|_{\Sigma_z+ \lambda I_d} + \frac{\lambda}{2\kappa} \|\Delta_z\|_2^2 ,
\end{align*}
By  $\|\widehat{\theta}\|_2\leq 1$, we have $ \|\Delta_z\|_2 = \|A^\top \widehat{\theta} - \beta\|_2 = \|\widehat{\theta} - \theta^*\|_2 \leq 2$, and therefore
\begin{align}\label{eq:tech-2}
\|\Delta_z\|_{\Sigma_z+\lambda I_d}^2 \leq 2\kappa \cdot \|A^\top \nabla l(\theta^*)\|_{(\Sigma_z + \lambda I_d)^{-1}}\cdot\|\Delta_z\|_{\Sigma_z+ \lambda I_d}+ {2\lambda}.
\end{align}
To derive an upper bound for $\|\Delta_z\|_{\Sigma_z + \lambda I_d}$, we need to upper bound $\|A^\top \nabla l(\theta^*)\|_{(\Sigma_z + \lambda I_d)^{-1}}$. By Bernstein’s inequality for sub-Gaussian random variables in quadratic form (e.g. see Theorem 2.1 in \cite{hsu2011tail}), we have $$\|A^\top \nabla l(\theta^*)\|_{(\Sigma_z + \lambda I_d)^{-1}} \leq c\cdot\sqrt{\frac{d+\log1/\delta}{n}}$$ for some constant $c$ with probability at least $1-\delta$.
Then we have $$
\|\Delta_z\|_{\Sigma_z+\lambda I_d}^2 \leq 2c\cdot\kappa \cdot \sqrt{\frac{d+\log1/\delta}{n}}\cdot\|\Delta_z\|_{\Sigma_z+ \lambda I_d}+ {2\lambda},
$$
solving which gives $$
\operatorname{(ii)} \leq O\bigg(\sqrt{\kappa^2\cdot \frac{d+\log1/\delta}{n}+ \lambda}\bigg).
$$
Combining (i) and (ii), we prove that $$
\cE_1 = \EE_{x\sim P_a} [|\widehat{f}(x) - f^*(x)|] \leq O\bigg(\sqrt{\kappa^2\cdot \frac{d+\log1/\delta}{n}+ \lambda} \cdot \sqrt{ \operatorname{Tr}(\EE_{P_a}[z z^\top] (\Sigma_z + \lambda I_d)^{-1} )}\bigg).
$$
By Lemma \ref{lmm:XtoZ}, we have \begin{align*}
    \cE_1 \leq O\bigg(\sqrt{\kappa^2\cdot \frac{d+\log1/\delta}{n}+ \lambda} \cdot \sqrt{ \operatorname{Tr}(\EE_{P_a}[xx^\top] (\Sigma_x + \lambda I_D)^{-1} )}\bigg),
\end{align*}
recall that $\tilde{\Sigma}_\lambda := \frac{1}{n_2}(\sum_{i=1}^{n_2} (x_i^{(1)} - x_i^{(2)})(x_i^{(1)} - x_i^{(2)})^\top + \lambda I)$, reset $\lambda = \lambda / n_2$ and we conclude the proof.
}

\subsection{Proof of lemma \ref{lmm:E2}}
\label{pf:E2}
Recall the definition of $g(x)$ that 
\begin{equation*}
    g(x) = {\theta^{*}}^{\top} AA^{\top} x,
\end{equation*}
note that $g(x) = {\theta^{*}}^{\top} x$ when $x$ is supported on $\cA$. Thus,
\begin{align*}
    &\left| \EE_{x \sim P_a} [g(x)] - \EE_{x \sim \hat{P}_a} [g(x)]\right|\\ = &\left| \EE_{x \sim P_a} [{\theta^{*}}^{\top} x] - \EE_{x \sim \hat{P}_a} [{\theta^{*}}^{\top} AA^{\top} x]\right|\\
    \leq & \left| \EE_{x \sim P_a} [{\theta^{*}}^{\top} x] - \EE_{x \sim \hat{P}_a} [{\theta^{*}}^{\top} VV^{\top} x]\right| + \underbrace{\left| \EE_{x \sim \hat{P}_a} [{\theta^{*}}^{\top} VV^{\top} x] - \EE_{x \sim \hat{P}_a} [{\theta^{*}}^{\top} AA^{\top} x]\right|}_{e_1},
\end{align*}
where 
\begin{align}
    \nonumber e_1 &= \left| \EE_{x \sim \hat{P}_a} [{\theta^{*}}^{\top} \left( VV^{\top} - AA^{\top} \right) x] \right|\\
    \nonumber & \leq  \EE_{x \sim \hat{P}_a} \left[ \left( \left\|(VV^{\top} - AA^{\top} \right) x\right\| \right] \\
    \nonumber &\leq  \| VV^{\top} - AA^{\top} \|_{F} \cdot \sqrt{\EE_{x \sim \hat{P}_a} \left[ \left\| x\right\|^2_2 \right]}.
\end{align}

Use notation $ P^{LD}(a) = P (z \mid \hat{f} (Az) = a)$, $ P^{LD}_{t_0}(a) = P_{t_0} (z \mid \hat{f} (Az) = a)$
\begin{align*}
    &\left| \EE_{x \sim P_a} [{\theta^{*}}^{\top} x] - \EE_{x \sim \hat{P}_a} [{\theta^{*}}^{\top} VV^{\top} x]\right|\\
    =& \left| \EE_{z \sim P(z \mid \hat f(Az) = a)} [{\theta^{*}}^{\top} A z] - \EE_{z \sim (U^{\top}V^{\top})_{\#}\hat{P}_a} [{\theta^{*}}^{\top} VU z]\right| \\
    \leq & \left| \EE_{z \sim P^{LD}_{t_0}(a)} [{\theta^{*}}^{\top} A z] - \EE_{z \sim (U^{\top}V^{\top})_{\#}\hat{P}_a} [{\theta^{*}}^{\top} VU z]\right| + \underbrace{\left| \EE_{z \sim P^{LD}_{t_0}(a)} [{\theta^{*}}^{\top} A z] - \EE_{z \sim P^{LD}(a)} [{\theta^{*}}^{\top} A z] \right|}_{e_2},
\end{align*}
here
\begin{align}
     \nonumber e_2 &=  \left|\alpha(t_0) \EE_{z \sim P^{LD}(a) } [{\theta^{*}}^{\top} A z] + h(t_0)\EE_{u \sim \sf N(0, I_d) } [{\theta^{*}}^{\top} A u] - \EE_{z \sim P^{LD}(a)} [{\theta^{*}}^{\top} A z] \right|\\
     \nonumber &\leq (1-\alpha(t_0)) \left| \EE_{z \sim P^{LD}(a) } [{\theta^{*}}^{\top} A z]\right|\\
    \nonumber&\leq t_0 \cdot \EE_{z \sim P^{LD}(a) } [\left\| z\right\|_2].
\end{align}
Then what is left to bound is 
\begin{align*}
   &\left| \EE_{z \sim P^{LD}_{t_0}(a)} [{\theta^{*}}^{\top} A z] - \EE_{z \sim (U^{\top}V^{\top})_{\#}\hat{P}_a} [{\theta^{*}}^{\top} VU z]\right|\\
   \leq & \underbrace{\left| \EE_{z \sim P^{LD}_{t_0}(a)} [{\theta^{*}}^{\top} VU z] - \EE_{z \sim (U^{\top}V^{\top})_{\#}\hat{P}_a} [{\theta^{*}}^{\top} VU z]\right|}_{e_3} +\underbrace{\left| \EE_{z \sim P^{LD}_{t_0}(a)} [{\theta^{*}}^{\top} VU z] - \EE_{z \sim P^{LD}_{t_0}(a)} [{\theta^{*}}^{\top} A z]\right|}_{e_4}.
\end{align*}
Then for term $e_3$, by Lemma \ref{lmm:exp_by_tv}, we get
\begin{align*}
    e_3 &\leq \left| \EE_{z \sim P^{LD}_{t_0}(a)} [\|z\|_2] - \EE_{z \sim (U^{\top}V^{\top})_{\#}\hat{P}_a} [\|z\|_2]\right|\\
    &= \cO \left(TV(\hat{P}_a) \cdot \left(\sqrt{ \EE_{z \sim P^{LD}_{t_0}(a)}[\|z\|^2_2]} +  \sqrt{ \EE_{x \sim \hat{P}_a} [ \| x\|^2_2 ]} \right)  \right),
\end{align*}
where we use $\EE_{z \sim (U^{\top}V^{\top})_{\#}\hat{P}_a} [\|z\|_2] \leq \EE_{x \sim \hat{P}_a} [ \| x\|_2 ]$

For $e_4$, we have
\begin{align}
    \nonumber e_4 &= \left| \EE_{z \sim P^{LD}_{t_0}(a)} [{\theta^{*}}^{\top} (VU-A) z] \right|\\
    \nonumber &= \alpha(t_0) \left| \EE_{z \sim P (a)} [{\theta^{*}}^{\top} (VU-A) z] \right|\\
    \nonumber &\leq \|VU-A\|_{F} \cdot \EE_{z \sim P^{LD}(a) } [\left\| z\right\|_2].
\end{align}
Therefore, by combining things together, we have
\begin{align*}
    \cE_2 \leq& e_1 + e_2 + e_3 + e_4\\
    \leq& \| VV^{\top} - AA^{\top} \|_{F} \cdot \sqrt{\EE_{x \sim \hat{P}_a} [ \| x\|^2_2 ]} + (\|VU-A\|_{F}+t_0) \cdot \sqrt{M(a)} \\
    &+\cO \left(TV(\hat{P}_a) \cdot \left(\sqrt{M(a) + t_0 d} + \sqrt{\EE_{x \sim \hat{P}_a} [ \| x\|^2_2 ]} \right)\right).
\end{align*}

By Lemma~\ref{lmm:diff_results} and Lemma~\ref{lmm:VU_A}, we have
\begin{align*}
    &TV(\hat{P}_a) = \tilde{\cO}\left(\sqrt{\frac{\cT(P(x, \hat{y} = a), P_{x\hat{y}}; \bar{\cS})}{\lambda_{\min}}} \cdot \epsilon_{diff} \right),\\ &\norm{VV^\top - AA^\top}_{\rm F} = \tilde{\cO}\left(\frac{\sqrt{t_0}}{\sqrt{\lambda_{\min}}} \cdot \epsilon_{diff} \right),\\
    & \norm{VU - A}_{\rm F} = \cO(d^{\frac{3}{2}}\norm{VV^\top - AA^\top}_{\rm F}).
\end{align*}

And by Lemma \ref{lmm:exp_x_norm}
\begin{equation*}
    \EE_{x \sim \hat{P}_a} \left[ \|x\|^2_2 \right] = \cO \left( ct_0D + M(a) \cdot  (1+ TV(\hat{P}_a) \right).
\end{equation*}

Therefore. leading term in $\cE_2$ is
\begin{equation*}
    \cE_2 = \cO\left((TV(\hat{P}_a) + t_0)\sqrt{M(a)} \right).
\end{equation*}
By plugging in score matching error $\epsilon^2_{diff} = \tilde \cO \left(\frac{1}{t_0}\sqrt{ \frac{Dd^2 + D^2d} {n_1} } \right)$, we have
\begin{equation*}
    TV(\hat{P}_a) = \tilde{\cO}\left(\sqrt{\frac{\cT(P(x, \hat{y} = a), P_{x\hat{y}}; \bar{\cS})}{\lambda_{\min}}} \cdot  \left(\frac{Dd^2 + D^2d} {n_1} \right)^{\frac{1}{4}} \cdot \frac{1}{\sqrt{t_0}} \right).
\end{equation*}
When $t_0 = \left((Dd^2 + D^2d) / n_1\right)^{1/6}$, it admits the best trade off in $\cE_2$ and $\cE_2$ is bounded by
\begin{equation*}
    \cE_2 = \tilde \cO\left(\sqrt{\frac{\cT(P(x, \hat{y} = a), P_{x\hat{y}}; \bar{\cS})}{\lambda_{\min}}} \cdot  \left(\frac{Dd^2 + D^2d} {n_1} \right)^{\frac{1}{6}} \cdot a \right).
\end{equation*}

\subsection{Proof of Lemma \ref{lmm:VU_A}}
\label{sec:proof_VU_A}
    From Lemma 17 in \cite{chen2023score}, we have
    $$\|U - V^{\top}A\|_{F} = \cO(\|V V^{\top} - A A^{\top}\|_{F}).$$
    Then it suffices to bound 
    $$\left| \|VU - A\|^2_{F} - \|U - V^{\top}A\|^2_{F} \right|,$$ where
    \begin{align*}
        &\|VU - A\|^2_{F} = 2d - \operatorname{trace} \left( U^{\top} V^{\top} A + A^{\top} VU  \right)\\
        &\|U - V^{\top}A\|^2_{F}  = d + \operatorname{trace} \left( A^{\top} V V^{\top}A \right) - \operatorname{trace} \left( U^{\top} V^{\top} A + A^{\top} VU  \right).
    \end{align*}
    Thus $$ \left| \|VU - A\|^2_{F} - \|U - V^{\top}A\|^2_{F} \right| = \left| d - \operatorname{trace} \left( A^{\top} V V^{\top}A \right) \right| = \left| \operatorname{trace} \left( A A^{\top} (V V^{\top} - A A^{\top})\right) \right|,$$ which is because $\operatorname{trace} \left( A^{\top} V V^{\top}A \right)$ is calcualted as
    \begin{align*}
        \operatorname{trace} \left( A^{\top} V V^{\top}A \right) &= \operatorname{trace} \left( A A^{\top} V V^{\top}\right)\\
        &= \operatorname{trace} \left( A A^{\top} A A^{\top}\right) +  \operatorname{trace} \left( A A^{\top} (V V^{\top} - A A^{\top})\right)\\
        &= d +  \operatorname{trace} \left( A A^{\top} (V V^{\top} - A A^{\top})\right).
    \end{align*}
    Then we will bound $\left| \operatorname{trace} \left( A A^{\top} (V V^{\top} - A A^{\top})\right) \right|$ by $\|V V^{\top} - A A^{\top}\|_{F}$, 
    \begin{align*}
        \left| \operatorname{trace} \left( A A^{\top} (V V^{\top} - A A^{\top})\right) \right| 
        &\leq  \operatorname{trace} \left( A A^{\top}\right) \operatorname{trace} \left( \left|V V^{\top} - A A^{\top} \right| \right) \\
        &\leq d \cdot \operatorname{trace} \left( \left|V V^{\top} - A A^{\top} \right| \right)\\
        &\leq d \cdot \sqrt{2d \left\|V V^{\top} - A A^{\top} \right\|^2_{F}}.
    \end{align*}
    Thus, $\|VU - A\|_{F} = \cO \left( d^{\frac{3}{2}} \sqrt{\subangle{V}{A}}\right).$

\subsection{Proof of Lemma \ref{lmm:exp_by_tv}}
\label{sec:exp_by_tv}
When $P_1$ and $P_2$ are Gaussian, $m(z) = \|z\|^2_2$, 
\begin{align*}
     & \quad \left| \EE_{z \sim P_1}[m(z)] - \EE_{z \sim P_2}[m(z)] \right| \\
     &= \left|\int m(z) \left( p_1(z) - p_2(z) \right) \diff z \right|\\
     &\leq \left|\int_{\|z\|_2 \leq R} \|z\|^2_2 \left( p_1(z) - p_2(z) \right) \diff z  \right| + \int_{\|z\|_2 > R} \|z\|^2_2 p_1(z)\diff z +  \int_{\|z\|_2 > R}  \|z\|^2_2 p_2(z) \diff z \\
     &\leq R^2 \dtv(P_1, P_2) + \int_{\|z\|_2 > R} \|z\|^2_2 p_1(z) \diff z +  \int_{\|z\|_2 > R}  \|z\|^2_2 p_2(z) \diff z. 
\end{align*}
Since $P_1$ and $P_2$ are Gaussains, $\int_{\|z\|_2 > R} \|z\|^2_2 p_1(z)\diff z$  and $\int_{\|z\|_2 > R}  \|z\|^2_2 p_2(z) \diff z$ are bounded by some constant $C_1$ when $R^2 \geq C_2 \max\{\EE_{z \sim P_1} [\|z\|_2^2], \EE_{z \sim P_2} [\|z\|_2^2]\}$ as suggested by Lemma 16 in \cite{chen2023score}.

Therefore,
\begin{align*}
    \EE_{z \sim P_1}[\|z\|^2_2] &\leq \EE_{z \sim P_2}[\|z\|^2_2]  + C_2 \max\{\EE_{P_1} [\|z\|_2^2], \EE_{P_2} [\|z\|_2^2]\} \cdot \dtv(P_1, P_2) + 2C_1\\
    &\leq \EE_{z \sim P_2}[\|z\|^2_2]  + C_2 (\EE_{z \sim P_1} [\|z\|_2^2] +  \EE_{z \sim P_2} [\|z\|_2^2]) \cdot \dtv(P_1, P_2) + 2C_1.
\end{align*}
Then
\begin{equation*}
    \EE_{z \sim P_1}[\|z\|^2_2] = \cO \left(\EE_{z \sim P_2}[\|z\|^2_2]  + \EE_{z \sim P_2} [\|z\|_2^2] \cdot \dtv(P_1, P_2) \right)
\end{equation*}
since $\dtv(P_1, P_2)$ decays with $n_1$.

Similarly, when $m(z) = \|z\|_2$
\begin{align*}
     & \quad \left| \EE_{z \sim P_1}[m(z)] - \EE_{z \sim P_2}[m(z)] \right| \\
     &= \left|\int m(z) \left( p_1(z) - p_2(z) \right) \diff z \right|\\
     & \leq \left|\int_{\|z\|_2 \leq R} \|z\|_2 \left( p_1(z) - p_2(z) \right) \diff z  \right| + \int_{\|z\|_2 > R} \|z\|_2 p_1(z)\diff z +  \int_{\|z\|_2 > R}  \|z\|_2 p_2(z) \diff z \\
     &\leq R \dtv(P_1, P_2) + \sqrt{\int_{\|z\|_2 > R} \|z\|^2_2 p_1(z) \diff z} +  \sqrt{\int_{\|z\|_2 > R}  \|z\|^2_2 p_2(z) \diff z},
\end{align*}
where $\int_{\|z\|_2 > R} \|z\|^2_2 p_1(z)\diff z$  and $\int_{\|z\|_2 > R}  \|z\|^2_2 p_2(z) \diff z$ are simultaneously bounded by a constant $C_1$, if the radius $R^2 \geq C_2 \max\{\EE_{z \sim P_1} [\|z\|_2^2], \EE_{z \sim P_2} [\|z\|_2^2]\}$ as suggested by \citet[Lemma 16]{chen2023score}.

Therefore,
\begin{align*}
    \left| \EE_{z \sim P_1}[\|z\|_2] - \EE_{z \sim P_2}[\|z\|_2] \right| &\leq \sqrt{C_2 \max\{\EE_{P_1} [\|z\|_2^2], \EE_{P_2} [\|z\|_2^2]\}} \cdot \dtv(P_1, P_2) + 2C_1\\
    &\leq \left(\sqrt{C_2 \EE_{z \sim P_1} [\|z\|_2^2]} +  \sqrt{ C_2 \EE_{z \sim P_2} [\|z\|_2^2]} \right) \cdot \dtv(P_1, P_2) + 2C_1\\
    & = \cO \left( \left(\sqrt{ \EE_{z \sim P_1} [\|z\|_2^2]} +  \sqrt{ \EE_{z \sim P_2} [\|z\|_2^2]} \right) \cdot \dtv(P_1, P_2) \right).
\end{align*}
\endproof
\subsection{Proof of Lemma \ref{lmm:exp_x_norm}}
\label{sec:proof_exp_x_norm}
Recall from \eqref{equ:z|hat_f} that
\begin{equation*}
    P^{LD}(a) = P_z(z \mid \hat{f} (Az) = a) = \sf N \left( \mu(a), \Gamma \right)
\end{equation*}
with $\mu(a):= \Sigma \hat{\beta} \left( \hat{\beta}^{\top} \Sigma \hat{\beta} + \nu^2 \right)^{-1} a$, $\Gamma:= \Sigma - \Sigma \hat{\beta} \left( \hat{\beta}^{\top} \Sigma \hat{\beta} + \nu^2 \right)^{-1} \hat{\beta}^{\top} \Sigma$.

\begin{align*}
    \EE_{z \sim P^{LD}(a)} \left[  \|z\|^2_2 \right] &= \mu(a)^{\top} \mu(a) + \operatorname{trace}(\Gamma)\\
    &= \frac{\hat{\beta}^{\top} \Sigma^2 \hat{\beta}}{\left( \|\hat{\beta}\|_{\Sigma}^2 + \nu^2 \right)^2} a^{2}  + \operatorname{trace}(\Sigma - \Sigma \hat{\beta} \left( \hat{\beta}^{\top} \Sigma \hat{\beta} + \nu^2 \right)^{-1} \hat{\beta}^{\top} \Sigma)\\
    &=: M(a).
\end{align*}
\begin{align*}
    M(a) &= \cO \left( \frac{\hat{\beta}^{\top} \Sigma^2 \hat{\beta}}{\left( \|\hat{\beta}\|_{\Sigma}^2 \right)^2} a^{2}  + \operatorname{trace} (\Sigma) \right)\\
    &= \cO \left( \frac{a^2}{\|\hat{\beta}\|_{\Sigma}} + d \right),
\end{align*}
and by Lemma \ref{lmm:distribution_sft}
\begin{equation*}
        \| \hat{\beta} \|_{\Sigma} \leq  \frac{1}{2} \| {\beta}^* \|_{\Sigma}.
    \end{equation*}
Thus $\EE_{z \sim P^{LD}(a)} \left[  \|z\|^2_2 \right] = M(a), M(a) = \cO\left( \frac{a^2}{\|{\beta}^*\|_{\Sigma}} + d \right)$.

Thus after adding diffusion noise at $t_0$, we have
for $\alpha(t)=e^{-t / 2}$ and $h(t)=$ $1-e^{-t}$:
\begin{align*}
    \EE_{z \sim P^{LD}_{t_0}(a)} \left[ \|z\|^2_2 \right] &= \EE_{z_0 \sim P^{LD}(a)} \EE_{z \sim \sf N \left( \alpha(t_0) \cdot z_0, h(t_0) \cdot I_d \right)} \left[ \|z\|^2_2 \right] \\
    &= \EE_{z_0 \sim P^{LD}(a)} \left[ \alpha^2(t_0) \|z_0\|^2_2 + d \cdot h(t_0)\right]\\\
    &= \alpha^2(t_0) \cdot \EE_{z_0 \sim P^{LD}(a)} \left[  \|z_0\|^2_2 \right] + d \cdot h(t_0)\\
    &= e^{-t_0} \cdot \EE_{z_0 \sim P^{LD}(a)} \left[  \|z_0\|^2_2 \right] + (1 - e^{-t_0}) \cdot d.
\end{align*}
Thus $ \EE_{z \sim P^{LD}_{t_0}(a)} \left[ \|z\|^2_2 \right] \leq M(a) + t_0 d$.
    
    By orthogonal decomposition we have
\begin{align*}
    \EE_{x \sim \hat{P}_a} \left[ \|x\|^2_2 \right] &\leq \EE_{x \sim \hat{P}_a} \left[ \|(I_D - VV^{\top}) x\|^2_2 \right] + \EE_{x \sim \hat{P}_a} \left[ \| VV^{\top} x\|^2_2 \right]\\
    &= \EE_{x \sim \hat{P}_a} \left[ \|(I_D - VV^{\top}) x\|^2_2 \right] + \EE_{x \sim \hat{P}_a} \left[ \| U^{\top} V^{\top} x\|^2_2 \right],
\end{align*}
where $\EE_{x \sim \hat{P}_a} \left[ \|(I_D - VV^{\top}) x\|^2_2 \right]$ is bounded by \eqref{equ:bkw_Gaussian_l2} and the distribution of $U^{\top}V^{\top} x$, which is $(U^{\top}V^{\top})_{\#}\hat{P}_a$, is close to $\mathbb{P}^{LD}_{t_0} (a)$ up to $TV(\hat{P}_a)$, which is defined in Definition~\ref{def:tv}. Then by Lemma~\ref{lmm:exp_by_tv}, we have
\begin{align*}
    \EE_{x \sim \hat{P}_a} \left[ \| U^{\top} V^{\top} x\|^2_2 \right]
    &= \cO \left(\EE_{z \sim P^{LD}_{t_0}(a)} \left[ \|z\|^2_2 \right] (1+ TV(\hat{P}_a) \right).
\end{align*}
Thus $\EE_{x \sim \hat{P}_a} \left[ \|x\|^2_2 \right] = \cO \left( ct_0D + (M(a) + t_0 d) \cdot  (1+ TV(\hat{P}_a) \right)$.

\subsection{Proof of Lemma \ref{lmm:XtoZ}}
    Firstly, one can verify the following two equations by direct calculation: 
    \begin{align*}
        (\lambda I_D+A\Sigma_1 A^\top)^{-1}&=\frac{1}{\lambda}\left(I_D-A(\lambda I_d+\Sigma_1)^{-1}\Sigma_1A^\top\right),\\
        (\lambda I_d+\Sigma_1)^{-1}&=\frac{1}{\lambda}\left(I_d-(\lambda I_d+\Sigma_1)^{-1}\Sigma_1\right).
    \end{align*}
    Then we have
    \begin{align*}
        (\lambda I_D+A\Sigma_1 A^\top)^{-1}A\Sigma_2 A^\top=&\frac{1}{\lambda}\left(I_D-A(\lambda I_d+\Sigma_1)^{-1}\Sigma_1A^\top\right)A\Sigma_2A^\top\\
        =&\frac{1}{\lambda}\left(A\Sigma_2A^\top-A(\lambda I_d+\Sigma_1)^{-1}\Sigma_1\Sigma_2A^\top\right).
    \end{align*}
    Therefore, 
    \begin{align*}
        \mathrm{Tr}\left((\lambda I_D+A\Sigma_1A^\top)^{-1}A\Sigma_2A^\top\right)=&\mathrm{Tr}\left(\frac{1}{\lambda}\left(A\Sigma_2A^\top-A(\lambda I_d+\Sigma_1)^{-1}\Sigma_1\Sigma_2A^\top\right)\right)\\
        =&\mathrm{Tr}\left(\frac{1}{\lambda}\left(\Sigma_2-(\lambda I_d+\Sigma_1)^{-1}\Sigma_1\Sigma_2\right)\right)\\
        =&\mathrm{Tr}\left(\frac{1}{\lambda}\left(I_d-(\lambda I_d+\Sigma_1)^{-1}\Sigma_1\right)\Sigma_2\right)\\
        =&\mathrm{Tr}\left(\left(\lambda I_d+\Sigma_1\right)^{-1}\Sigma_2\right),
    \end{align*}
    which has finished the proof.

\section{Omitted Proofs in Section~\ref{sec:nonparametric}}
\label{pf:nonparametric_all}

\subsection{Conditional Score Decomposition and Score Matching Error}\label{pf:nonparametric_score}
\begin{lemma}\label{lemma:score_error}
Under Assumption \ref{assumption:subspace}, \ref{assumption:tail_non} and \ref{asmp:lipschitz}, with high probability
\begin{equation*}
    \frac{1}{T-t_0}\int_{t_0}^T \norm{\hat{s}(\cdot, t) - \nabla \log p_t(\cdot)}_{L^2(P_t)}^2 \diff t \leq \epsilon_{diff}^2(n_1),
\end{equation*}
with $\epsilon_{diff}^2(n_1) = \tilde{\cO}\left(\frac{1}{t_0} \left(n_1^{-\frac{2 - 2\delta(n_1)}{d+6}} + Dn_1^{-\frac{d+4}{d+6}}\right)\right)$ for $\delta(n_1) = \frac{d\log\log n_1}{\log n_1}$.
\end{lemma}

\proof{}
\citet[Theorem 1]{chen2023score} is easily adapted here to prove Lemma~\ref{lemma:score_error} with the input dimension $d+1$ and the Lipschitzness in Assumption~\ref{asmp:lipschitz}. Network size of $\cS$ is implied by \citet[Theorem 1]{chen2023score} with $\epsilon = n_1^{-\frac{1}{d+6}}$ accounting for the additional dimension of reward $\hat{y}$ and then the score matching error follows.
\endproof
\subsection{Proof of Theorem~\ref{thm:nonparametric}}\label{pf:nonparametric}
\textbf{Additional Notations:} Similar as before, use $P_t^{LD}(z)$ to denote the low-dimensional distribution on $z$ corrupted by diffusion noise. Formally, $p_t^{LD}(z) =  \int  \phi_t(z'|z)p_z(z) \diff z$ with $\phi_t( \cdot | z)$ being the density of ${\sf N}(\alpha(t)z, h(t)I_d)$. $P^{LD}_{t_0} (z \mid \hat{f} (Az) = a)$ the corresponding conditional distribution on $\hat{f}(Az)= a$ at $t_0$, with shorthand as $P^{LD}_{t_0}(a)$. Also give $P_{z}(z \mid \hat{f} (Az) = a)$ a shorthand as $P^{LD}(a)$.

Now we are ready to begin our proof. By the same argument as in Appendix \ref{sec:dcp}, we have
\begin{align*}
    \subopt(\hat{P}_a ; y^* = a) 
    &\leq \underbrace{\EE_{x \sim P_a} \left[ \left|f^*(x) - \hat{f}(x)\right| \right]}_{\cE_1} + \underbrace{\left| \EE_{x \sim P_a} [g^*(\px)] - \EE_{x \sim \hat{P}_a} [g^*(\px)]\right|}_{\cE_2} +  \underbrace{\EE_{x \sim \hat{P}_a} [h^*(\ox)]}_{\cE_3}.
\end{align*}
In the following, we will bound terms $\cE_1,\cE_2$  respectively. Since $\cE_3$ is essentially the off-support error, its magnitude is  bounded by Theorem \ref{thm:fidelity}.
\subsubsection{$\cE_1$: Nonparamtric Regression Induced Error.}

Since $P_z$ has a light tail due to Assumption~\ref{assumption:tail_non}, by union bound and \citet[Lemma 16]{chen2023score}, we have
\begin{align*}
\PP(\exists~x_i~\text{with}~\norm{x_i}_2 > R~\text{for}~i = 1, \dots, n_2) \leq n_2 \frac{C_1d2^{-d/2+1}}{C_2\Gamma(d/2 + 1)} R^{d-2} \exp(-C_2 R^2 / 2),
\end{align*}
where $C_1, C_2$ are constants and $\Gamma(\cdot)$ is the Gamma function. Choosing $R = \cO(\sqrt{d\log d + \log \frac{n}{\delta}})$ ensures $\PP(\exists~x_i~\text{with}~\norm{x_i}_2 > R~\text{for}~i = 1, \dots, n_2) < \delta$. On the event $\cE = \{\norm{x_i}_2 \leq R ~\text{for all}~ i = 1, \dots, n_2\}$, denoting $\delta(n_2) = \frac{d \log \log n_2}{\log n_2}$, we have
\begin{align*}
\norm{f^* - \hat{f}}^2_{L^2} = \tilde{\cO}\left(n_2^{-\frac{2(\alpha - \delta(n_2))}{d + 2\alpha}} \right)
\end{align*}
by \citet[Theorem 7]{nakada2020adaptive} with a new covering number of $\cS$, when $n_2$ is sufficiently large. The corresponding network architecture follows \citet[Theorem 2]{chen2022nonparametric}.

We remark that linear subspace is a special case of low Minkowski dimension. Moreover, $\delta(n_2)$ is asymptotically negligible and accounts for the truncation radius $R$ of $x_i$'s (see also \citet[Theorem 2 and 3]{chen2023score}). The covering number of $\cS$ is $\tilde{\cO}\left(d^{d/2} n^{-\frac{d}{\alpha}} (\log n_2)^{d/2} + Dd\right)$ as appear in \citet[Proof of Theorem 2]{chen2023score}.
Therefore
\begin{align*}
    \EE_{x \sim P_a} \left[ \left|f^*(x) - \hat{f}(x)\right| \right] &\leq \sqrt{\EE_{x \sim P_a} \left[ \left|f^*(x) - \hat{f}(x)\right|^2 \right]}\\
    &\leq \sqrt{\cT(P(x | \hat{y} = a), P_{x}; \bar{\cF}) \cdot \norm{f^* - \hat{f}}^2_{L^2}}\\
    &= \sqrt{\cT(P(x | \hat{y} = a), P_{x}; \bar{\cF})}  \cdot \tilde{\cO}\left( n_2^{-\frac{\alpha - \delta(n_2)}{2\alpha + d}} + D / n_2 \right).
\end{align*}

\subsubsection{$\cE_2$: Diffusion Induced On-support Error}

Suppose $L_2$ score matching error is $\epsilon^2_{diff} (n_1)$, i.e.
\begin{align*}
\frac{1}{T - t_0} \int_{t_0}^T \EE_{x, \hat f} \norm{\nabla_x \log p_t(x, \hat f) - s_{\hat w}(x, \hat f, t)}_2^2 \diff t \leq \epsilon^2_{diff} (n_1),
\end{align*}

We revoke Definition~\ref{def:tv} measuring the distance between $\hat{P}_a$ to $P_a$ that
$$TV(\hat{P}_a):= \dtv \left(P^{LD}_{t_0} (z \mid \hat{f} (Az) = a), (U^{\top} V^{\top})_{\#}\hat{P}_a \right).$$ Lemma~\ref{lmm:diff_results} applies to nonparametric setting, so we have
 \begin{align}
        &(I_D - VV^{\top}) x \sim {\sf N}(0, \Lambda), \quad \Lambda \prec c t_0 I_D,\\
       &\subangle{V}{A} = \tilde{\cO}\left(\frac{t_0 }{c_0} \cdot \epsilon^2_{diff}(n_1) \right).
    \end{align}
In addition,
    \begin{equation}
        TV(\hat{P}_a) = \tilde{\cO}\left(\sqrt{\frac{\cT(P(x, \hat{y} = a), P_{x\hat{y}}; \bar{\cS})}{c_0}} \cdot \epsilon_{diff}(n_1) \right).
    \end{equation}

$\cE_2$ will be bounded by
\begin{align*}
    \cE_2 &= \left| \EE_{x \sim P_a} [g^*(x)] - \EE_{x \sim \hat{P}_a} [g^*(x)]\right|\\
    &\leq \left| \EE_{x \sim P_a} [g^*(A A^{\top} x)] - \EE_{x \sim \hat{P}_a} [g^*(VV^{\top} x)]\right| + \left| \EE_{x \sim \hat{P}_a} [g^*(VV^{\top} x) - g^*(AA^{\top} x)] \right|, 
\end{align*}
where for $\left| \EE_{x \sim \hat{P}_a} [g^*(VV^{\top} x) - g^*(AA^{\top} x)] \right|$, we have
\begin{equation}
    \left| \EE_{x \sim \hat{P}_a} [g^*(VV^{\top} x) - g^*(AA^{\top} x)] \right| \leq \EE_{x \sim \hat{P}_a} [\|VV^{\top} x - AA^{\top} x\|_2] \leq \|VV^{\top} - AA^{\top}\|_{F} \cdot \EE_{x \sim \hat{P}_a} [\|x\|_2].
\end{equation}
For the other term $\left| \EE_{x \sim P_a} [g^*(A A^{\top} x)] - \EE_{x \sim \hat{P}_a} [g^*(VV^{\top} x)]\right|$, we will bound it with $TV(\hat{P}_a)$.
\begin{align*}
    &\left| \EE_{x \sim P_a} [g^*(A A^{\top} x)] - \EE_{x \sim \hat{P}_a} [g^*(VV^{\top} x)]\right| \\
     \leq& \left| \EE_{z \sim \mathbb{P}_{t_0} (a)} [g^*(Az)] - \EE_{z \sim (V^{\top})_{\#}\hat{P}_a}  [g^*(V z)]\right| + \left| \EE_{z \sim \mathbb{P} (a)} [g^*(A z)] - \EE_{z \sim \mathbb{P}_{t_0} (a)} [g^*(A z)]\right|\\
\end{align*}
Since any $z \sim \mathbb{P}_{t_0} (a)$ can be represented by $\alpha(t_0) z+ \sqrt{h(t_0)} u$, where $z \sim \mathbb{P} (a), u \sim {\sf N}(0, I_d)$, then
\begin{align*}
    &\EE_{z \sim \mathbb{P}_{t_0} (a)} [g^*(A z)]\\
    &= \EE_{z \sim \mathbb{P} (a), u \sim {\sf N}(0, I_d)} [g^*(\alpha(t)Az + \sqrt{h(t)}A u))]\\
    &\leq \EE_{z \sim \mathbb{P} (a)} [g^*(\alpha(t_0)Az))] + \sqrt{h(t_0)} \EE_{ u \sim {\sf N}(0, I_d)} [\|A u\|_2]\\
    &\leq \EE_{z \sim \mathbb{P} (a)} [g^*(Az))] + (1-\alpha(t_0)) \EE_{z \sim \mathbb{P} (a)} [\|Az\|_2] + \sqrt{h(t_0)} \EE_{ u \sim \sf N(0, I_d)} [\|A u\|_2],
\end{align*}
thus 
\begin{equation*}
    \left| \EE_{z \sim \mathbb{P} (a)} [g^*(A z)] - \EE_{z \sim \mathbb{P}_{t_0} (a)} [g^*(A z)]\right| \leq t_0 \cdot \EE_{z \sim \mathbb{P} ( a)} [\|z\|_2] + d ,
\end{equation*}
where we further use $1 - \alpha(t_0) = 1-e^{- t_0/2} \leq t_0/2$, $h(t_0) \leq 1$.

As for $\left| \EE_{z \sim \mathbb{P}_{t_0} (a)} [g^*(A z)] - \EE_{z \sim (V^{\top})_{\#}\hat{P}_a}  [g^*(V z)]\right|$, we have
\begin{align*}
    &\left| \EE_{z \sim \mathbb{P}_{t_0} (a)} [g^*(A z)] - \EE_{z \sim (U^{\top}V^{\top})_{\#}\hat{P}_a}  [g^*(V U z)]\right|\\
    = & \left| \EE_{z \sim \mathbb{P}_{t_0} (a)} [g^*(VU z)] - \EE_{z \sim (U^{\top} V^{\top})_{\#}\hat{P}_a}  [g^*(V U z)]\right| + \left| \EE_{z \sim \mathbb{P}_{t_0} (a)} [g^*(A z)] - \EE_{z \sim \mathbb{P}_{t_0} (a)}  [g^*(V U z)]\right|,
\end{align*}
where 
\begin{equation*}
    \left| \EE_{z \sim \mathbb{P}_{t_0} (a)} [g^*(A z)] - \EE_{z \sim \mathbb{P}_{t_0} (a)}  [g^*(V U z)]\right| \leq \|A - VU\|_{F} \cdot \EE_{z \sim \mathbb{P}_{t_0}(a)} [\|z\|_2],
\end{equation*}
and 
\begin{equation*}
    \left| \EE_{z \sim \mathbb{P}_{t_0} (a)} [g^*(VU z)] - \EE_{z \sim (U^{\top} V^{\top})_{\#}\hat{P}_a}  [g^*(V U z)]\right| \leq TV(\hat{P}_a) \cdot \|g^*\|_{\infty}.
\end{equation*}

Combining things up, we have
\begin{align*}
    \cE_2 \leq& \|VV^{\top} - AA^{\top}\|_{F} \cdot \EE_{x \sim \hat{P}_a} [\|x\|_2] + \|A - VU\|_{F} \cdot \EE_{z \sim \mathbb{P}_{t_0}(a)} [\|z\|_2]\\
    &+ t_0 \cdot \EE_{z \sim \mathbb{P} ( a)} [\|z\|_2] + d  +TV(\hat{P}_a) \cdot \|g^*\|_{\infty}.
\end{align*}
Similar to parametric case, Let $M(a): = \EE_{z \sim \mathbb{P} ( a)} [\|z\|^2_2]$, then
$$\EE_{z \sim \mathbb{P}_{t_0}(a)} [\|z\|^2_2] \leq M(a) + t_0 d,$$ expect for in nonparametric case, we can not compute $M(a)$ out as it is not Gaussian.
But still, with higher-order terms in $n_1^{-1}$ hided, we have
\begin{align*}
    \cE_2 &= \cO \left( TV(\hat{P}_a) \cdot \|g^*\|_{\infty} + t_0 M(a)\right)\\
    &= \tilde \cO \left( \sqrt{\frac{\cT(P(x, \hat{y} = a), P_{x\hat{y}}; \bar{\cS})}{c_0}} \cdot \epsilon_{diff}(n_1) \cdot \|g^*\|_{\infty} + t_0 M(a)\right).
\end{align*}

\section{Parametric Conditional Score Estimation: Proof of Lemma \ref{lmm:scr_mtc_err}}
\label{sec:score_matching_err}
We first derive a decomposition of the conditional score function similar to \cite{chen2023score}. We have
\begin{align*}
p_t(x, y) & = \int p_t(x, y | z) p_z(z) \diff z \\
& = \int p_t(x | z) p(y | z) p_z(z) \diff z \\
& = C \int\exp\left(-\frac{1}{2h(t)}\norm{x - \alpha(t)Az}_2^2\right) \exp\left(-\frac{1}{\sigma^2_y} \left(\theta^\top z - y\right)^2\right)p_z(z) \diff z \\
& \overset{(i)}{=} C \exp\left(-\frac{1}{2h(t)}\norm{(I_D - AA^\top)x}_2^2 \right)  \cdot \int \exp\left(-\frac{1}{2h(t)}\norm{A^\top x - \alpha(t)z}_2^2\right) \exp\left(-\frac{1}{\sigma^2_y} \left(\theta^\top z - y\right)^2\right) p_z(z) \diff z,
\end{align*}
where equality $(i)$ follows from the fact $AA^\top x \perp (I_D - AA^\top) x$ and $C$ is the normalizing constant of Gaussian densities. Taking logarithm and then derivative with respect to $x$ on $p_t(x, y)$, we obtain
\begin{align*}
\nabla_x \log p_t(x, y) = \frac{\alpha(t)}{h(t)} \frac{A \int z \exp\left(-\frac{1}{2h(t)}\norm{A^\top x - \alpha(t)z}_2^2\right) \exp\left(-\frac{1}{\sigma^2_y} \left(\theta^\top z - y\right)^2\right) p_z(z) \diff z}{\int \exp\left(-\frac{1}{2h(t)}\norm{A^\top x - \alpha(t)z}_2^2\right) \exp\left(-\frac{1}{\sigma^2_y} \left(\theta^\top z - y\right)^2\right) p_z(z) \diff z} - \frac{1}{h(t)} x.
\end{align*}
Note that the first term in the right-hand side above only depends on $A^\top x$ and $y$. Therefore, we can compactly write $\nabla_x \log p_t(x, y)$ as
\begin{align}\label{eq:conditional_score_decomp}
\nabla_x \log p_t(x, y) = \frac{1}{h(t)} A u(A^\top x, y, t) - \frac{1}{h(t)} x,
\end{align}
where mapping $u$ represents $$\frac{\alpha(t) \int z \exp\left(-\frac{1}{2h(t)}\norm{A^\top x - \alpha(t)z}_2^2\right) \exp\left(-\frac{1}{\sigma^2_y} \left(\theta^\top z - y\right)^2\right) p_z(z) \diff z}{\int \exp\left(-\frac{1}{2h(t)}\norm{A^\top x - \alpha(t)z}_2^2\right) \exp\left(-\frac{1}{\sigma^2_y} \left(\theta^\top z - y\right)^2\right) p_z(z) \diff z}.$$
We observe that \eqref{eq:conditional_score_decomp} motivates our choice of the neural network architecture $\cS$ in \eqref{equ:function_class}. In particular, $\psi$ attempts to estimate $u$ and matrix $V$ attempts to estimate $A$.

In the Gaussian design case (Assumption~\ref{assumption:gaussian_design}), we instantiate $p_z(z)$ to the Gaussian density $(2\pi |\Sigma|)^{-d/2} \exp\left(-\frac{1}{2} z^\top \Sigma^{-1} z\right)$. Some algebra on the Gaussian integral gives rise to
\begin{align}\label{eq:gaussian_score}
\nabla_x \log p_t(x, y) & = \frac{\alpha(t)}{h(t)} A B_t \mu_t(x, y) - \frac{1}{h(t)} (I_D - AA^\top) x - \frac{1}{h(t)} AA^\top x \nonumber \\
& = \frac{\alpha(t)}{h(t)} A B_t \left(\alpha(t) A^\top x + \frac{h(t)}{\nu^2} y \theta \right) - \frac{1}{h(t)} x,
\end{align}
where we have denoted
\begin{align*}
\mu_t(x, y) = \alpha(t) A^\top x + \frac{h(t)}{\nu^2} y \theta \quad \text{and} \quad B_t = \left(\alpha^2(t)I_d + \frac{h(t)}{\nu^2} \theta \theta^\top + h(t) \Sigma^{-1}\right)^{-1}.
\end{align*}

\subsection{Score Estimation Error} Recall that we estimate the conditional score function via minimizing the denoising score matching loss in Proposition~\ref{prop:equivalent_score_matching}. To ease the presentation, we denote
\begin{align*}
\ell(x, y; s) = \frac{1}{T-t_0} \int_{t_0}^T \EE_{x' | x} \norm{\nabla_{x'} \log \phi_t(x' | x) - s(x', y, t)}_2^2 \diff t
\end{align*}
as the loss function for a pair of clean data $(x, y)$ and a conditional score function $s$. Further, we denote the population loss as
\begin{align*}
\cL(s) = \EE_{x, y} [\ell(x, y; s)],
\end{align*}
whose empirical counterpart is denoted as $\hat{\cL}(s) = \frac{1}{n_1} \sum_{i=1}^{n_1} \ell(x_i, y_i; s)$.

To bound the score estimation error, we begin with an oracle inequality. Denote $\cL^{\rm trunc}(s)$ as a truncated loss function defined as
\begin{align*}
\cL^{\rm trunc}(s) = \EE [\ell(x, y; s) \mathds{1}\{\norm{x}_2 \leq R, |y| \leq R\}],
\end{align*}
where $R > 0$ is a truncation radius chosen as $\cO(\sqrt{d \log d + \log K + \log \frac{n_1}{\delta}})$. Here $K$ is a uniform upper bound of $s(x, y, t) \mathds{1}\{\norm{x}_2 \leq R, |y| \leq R\}$ for $s \in \cS$, i.e., $\sup_{s \in \cS} \norm{s(x, y, t) \mathds{1}\{\norm{x}_2 \leq R, |y| \leq R\}}_2 \leq K$. To this end, we have
\begin{align*}
\cL(\hat{s}) & = \cL(\hat{s}) - \hat{\cL}(\hat{s}) + \hat{\cL}(\hat{s}) \\
& = \cL(\hat{s}) - \hat{\cL}(\hat{s}) + \inf_{s \in \cS} \hat{\cL}(s) \\
& \overset{(i)}{=} \cL(\hat{s}) - \hat{\cL}(\hat{s}) \\
& \leq \cL(\hat{s}) - \cL^{\rm trunc}(\hat{s}) + \cL^{\rm trunc}(\hat{s}) - \hat{\cL}^{\rm trunc}(\hat{s}) \\
& \leq \underbrace{\sup_{s} \cL^{\rm trunc}(s) - \hat{\cL}^{\rm trunc}(s)}_{(A)} + \underbrace{\sup_{s} \cL(s) - \cL^{\rm trunc}(s)}_{(B)},
\end{align*}
where equality $(i)$ holds since $\cS$ contains the ground truth score function. We bound term $(A)$ by a PAC-learning concentration argument. Using the same argument in \citet[Theorem 2, term $(A)$]{chen2023score}, we have
\begin{align*}
\sup_{s \in \cS} \ell^{\rm trunc}(x, y; s) = \cO\left(\frac{1}{t_0(T-t_0)} (K^2 + R^2) \right).
\end{align*}
Applying the standard metric entropy and symmetrization technique, we can show
\begin{align*}
(A) = \cO\left( \hat{\mathfrak{R}}(\cS) + \left(\frac{K^2 + R^2}{t_0(T-t_0)} \right)\sqrt{\frac{\log \frac{2}{\delta}}{2n_1}}\right),
\end{align*}
where $\hat{\mathfrak{R}}$ is the empirical Rademacher complexity of $\cS$. Unfamiliar readers can refer to Theorem 3.3 in ``Foundations of Machine Learning'', second edition for details. The remaining step is to bound the Rademacher complexity by Dudley's entropy integral. Indeed, we have
\begin{align*}
\hat{\mathfrak{R}}(\cS) \leq \inf_{\epsilon} \frac{4\epsilon}{\sqrt{n_1}} + \frac{12}{n_1} \int_{\epsilon}^{K^2\sqrt{n_1}} \sqrt{\cN(\cS, \epsilon, \norm{\cdot}_2)} \diff \epsilon.
\end{align*}
We emphasize that the log covering number considers $x, y$ in the truncated region. Taking $\epsilon = \frac{1}{n_1}$ gives rise to
\begin{align*}
(A) = \cO\left(\left(\frac{K^2 + R^2}{t_0(T-t_0)}\right) \sqrt{\frac{\cN(\cS, 1/n_1) \log \frac{1}{\delta}}{n_1}}\right).
\end{align*}
Here $K$ is instance-dependent and majorly depends on $d$. In the Gaussian design case, we can verify that $K$ is $\cO(\sqrt{d})$. To this end, we deduce $(A) = \tilde{\cO}\left(\frac{1}{t_0} \sqrt{d^2\frac{\cN(\cS, 1/n_1) \log \frac{1}{\delta}}{n_1}}\right)$. In practice, $d$ is often much smaller than $D$ (see for example \cite{pope2021intrinsic}, where ImageNet has intrinsic dimension no more than $43$ in contrast to image resolution of $224 \times 224 \times 3$). In this way, we can upper bound $d^2$ by $D$, yet $d^2$ is often a tighter upper bound.

For term $(B)$, we invoke the same upper bound in \citet[Theorem 2, term $(B)$]{chen2023score} to obtain
\begin{align*}
(B) = \cO\left(\frac{1}{n_1 t_0(T-t_0)}\right),
\end{align*}
which is negligible compared to $(A)$. Therefore, summing up $(A)$ and $(B)$, we deduce
\begin{align*}
\epsilon_{diff}^2 = \cO\left(\frac{1}{t_0} \sqrt{\frac{\cN(\cS, 1/n_1) (d^2 \vee D) \log \frac{1}{\delta}}{n_1}}\right).
\end{align*}

\subsection{Gaussian Design} We only need to find the covering number under the Gaussian design case. Using \eqref{eq:gaussian_score}, we can construct a covering from coverings on matrices $V$ and $\Sigma^{-1}$. Suppose $V_1, V_2$ are two matrices with $\norm{V_1 - V_2}_2 \leq \eta_V$ for some $\eta > 0$. Meanwhile, let $\Sigma^{-1}_1, \Sigma^{-1}_2$ be two covariance matrices with $\norm{\Sigma^{-1}_1 
- \Sigma^{-1}_2}_2 \leq \eta_{\Sigma}$. Then we bound
\begin{align*}
& \quad \sup_{\norm{x}_2 \leq R, |y|\leq R} \norm{s_{V_1, \Sigma^{-1}_1}(s, y, t) - s_{V_2, \Sigma^{-1}_2}(x, y, t)}_2 \\
& \leq \frac{1}{h(t)} \sup_{\norm{x}_2 \leq R, |y|\leq R} \Big[\big\lVert V_1 \psi_{\Sigma^{-1}_1}(V_1^\top x, y, t) - V_1 \psi_{\Sigma^{-1}_1}(V_2^\top x, y, t)\big\rVert_2 \\
& \quad + \underbrace{\big\lVert V_1 \psi_{\Sigma^{-1}_1}(V_2^\top x, y, t) -  V_1 \psi_{\Sigma^{-1}_2}(V_2^\top x, y, t) \big\rVert_2}_{(\spadesuit)} + \big\lVert V_1 \psi_{\Sigma^{-1}_2}(V_2^\top x, y, t) - V_2 \psi_{\Sigma^{-1}_2}(V_2^\top x, y, t) \big\rVert_2\Big] \\
& \leq \frac{1}{h(t)} \left(2R \eta_V + 2 \nu^{-2} R \eta_{\Sigma} \right),
\end{align*}
where for bounding $(\spadesuit)$, we invoke the identity $\norm{(I+A)^{-1} - (I+B)^{-1}}_2 \leq \norm{B - A}_2$. Further taking supremum over $t \in [t_0, T]$ leads to
\begin{align*}
\sup_{\norm{x}_2 \leq R, |y|\leq R} \norm{s_{V_1, \Sigma^{-1}_1}(s, y, t) - s_{V_2, \Sigma^{-1}_2}(x, y, t)}_2 \leq \frac{1}{t_0} \left(2R \eta_V + 2 \nu^{-2} R \eta_{\Sigma} \right)
\end{align*}
for any $t \in [t_0, T]$. Therefore, the inequality above suggests that coverings on $V$ and $\Sigma^{-1}$ form a covering on $\cS$. The covering numbers of $V$ and $\Sigma^{-1}$ can be directly obtained by a volume ratio argument; we have
\begin{align*}
\cN(V, \eta_V, \norm{\cdot}_2) \leq Dd \log \left(1 + \frac{2\sqrt{d}}{\eta_V}\right) \quad \text{and} \quad \cN(\Sigma^{-1}, \eta_{\Sigma}, \norm{\cdot}_2) \leq d^2 \log \left(1 + \frac{2\sqrt{d}}{\lambda_{\min} \eta_{\Sigma}}\right).
\end{align*}
Thus, the log covering number of $\cS$ is
\begin{align*}
\cN(\cS, \eta, \norm{\cdot}_2) & = \cN(V, t_0 \eta_V/2R, \norm{\cdot}_2) + \cN(\Sigma^{-1}, t_0\nu^2\eta_{\Sigma}/2R, \norm{\cdot}_2) \\
& \leq (Dd + d^2) \log \left(1 + \frac{d D}{t_0 \lambda_{\min} \eta} \right),
\end{align*}
where we have plugged $\nu^2 = 1/D$ into the last inequality. Setting $\eta = 1/n_1$ and substituting into $\epsilon_{diff}^2$ yield the desired result.

We remark that the analysis here does not try to optimize the error bounds, but aims to provide a provable guarantee for conditional score estimation using finite samples. We foresee that sharper analysis via Bernstein-type concentration may result in a better dependence on $n_1$. Nonetheless, the optimal dependence should not beat a $1/n_1$-rate.

%% file: sections/tem_proofs.tex
\section{Conditional Score Matching and Generation}
\label{pf_sec:2}

\subsection{Proof of Proposition~\ref{prop:equivalent_score_matching}}\label{pf:equivalent_score_matching}
For any $t \geq 0$, it hold that $\nabla_{x_t} \log p_t(x_t \mid y) = \nabla_{x_t} \log p_t(x_t, y)$ since the gradient is taken w.r.t. $x_t$ only. Then plugging in this equation and expanding the norm square on the LHS gives
\begin{align*}
\EE_{(x_t, y) \sim P_t} \left[\norm{\nabla_{x_t} \log p_t(x_t, y) - s(x_t, y, t)}_2^2\right]  = \EE_{(x_t, y) \sim P_t} \big[\norm{s(x_t, y, t)}_2^2  - 2 \langle \nabla_{x_t} \log p_t(x_t, y), s(x_t, y, t)\rangle \big] + C.
\end{align*}
Then it suffices to prove 
\begin{equation*}
    \EE_{(x_t, y) \sim P_t} \left[ \langle \nabla_{x_t} \log p_t(x_t, y), s(x_t, y, t)\rangle \right] = \EE_{(x,y) \sim P_{x \hat{y}}} \EE_{x^{\prime} \sim {\sf N}(\alpha(t)x, h(t)I)} \left[\langle \nabla_{x^{\prime}}\phi_t(x^{\prime} \mid x), s(x^{\prime}, y, t)\rangle\right]
\end{equation*}
Using integration by parts to rewrite the inner product we have
\begin{align*}
    \EE_{(x_t, y) \sim P_t} \left[\langle \nabla_{x_t} \log p_t(x_t, y), s(x_t, y, t)\rangle \right] &= \int p_t(x_t,y) \langle \nabla_{x_t} \log p_t(x_t, y), s(x_t, y, t)\rangle dx_t dy\\
    &= \int \langle \nabla_{x_t} p_t(x_t, y), s(x_t, y, t)\rangle dx_t dy\\
    &= -\int p_t(x_t,y) \operatorname{div}(s(x_t, y, t)) dx_t dy,
\end{align*}
where denote by $\phi_t(x' | x)$ the density of ${\sf N}(\alpha(t)x, h(t)I_D)$ with $\alpha(t) = \exp(- t/2)$ and $h(t) = 1 - \exp(-t)$, then
\begin{align*}
    -\int p_t(x_t,y) \operatorname{div}(s(x_t, y, t)) dx_t dy &= -\EE_{(x,y) \sim P_{x\hat{y}}} \int   \phi_t(x^{\prime} \mid x) \operatorname{div}(s(x^{\prime}, y, t)) dx^{\prime}\\
    &= \EE_{(x,y) \sim P_{x\hat{y}}} \int   \langle \nabla_{x^{\prime}}\phi_t(x^{\prime} \mid x), s(x^{\prime}, y, t)\rangle  dx^{\prime}\\
    &= \EE_{(x,y) \sim P_{x \hat{y}}} \EE_{x^{\prime} \sim {\sf N}(\alpha(t)x, h(t)I)} \left[\langle \nabla_{x^{\prime}}\phi_t(x^{\prime} \mid x), s(x^{\prime}, y, t)\rangle \right].
\end{align*}

\section{Omitted Proofs in Section~\ref{sec:linear}}
\label{pf_sec:3}

\textbf{Additional Notations:} We follow the notations in the main paper along with some additional ones. Use $P_t^{LD}(z)$ to denote the low-dimensional distribution on $z$ corrupted by diffusion noise. Formally, $p_t^{LD}(z) =  \int  \phi_t(z'|z)p_z(z) \diff z$ with $\phi_t( \cdot | z)$ being the density of ${\sf N}(\alpha(t)z, h(t)I_d)$. $P^{LD}_{t_0} (z \mid \hat{f} (Az) = a)$ the corresponding conditional distribution on $\hat{f}(Az)= a$ at $t_0$, with shorthand as $P^{LD}_{t_0}(a)$. Also give $P_{z}(z \mid \hat{f} (Az) = a)$ a shorthand as $P^{LD}(a)$. In our theorems, $\cO$ hides constant factors and higher order terms in $n_1^{-1}$ and $n_2^{-1}$ and , $\tilde \cO$ further hides logarithmic terms and can also hide factors in $d$.

\subsection{Parametric Conditional Score Matching Error }\label{pf:parametric_score}

Theorems presented in Section~\ref{sec:linear} are established upon the conditional score estimation error, which has been studied in \cite{chen2023score} for general distributions, but in Lemma~\ref{lmm:scr_mtc_err} we provide a new one specific to our setting where the true score is linear in input $(x_t, \hat{y})$ due to the Gaussian design. Despite the linearity of score in Gaussian case, we emphasize matching score in \eqref{equ:scr_mtc} is not simply linear regression as $\cS$ consists of an encoder-decoder structure for estimating matrix $A$ to reduce dimension (see \S\ref{sec:score_matching_err} for $\cS$ construction and more proof details). 

In the following lemma, we first present a general result for the case where the true score is within $\cS$, which is  constructed as a parametric function class. Then the score matching error is bounded in terms of $\cN(\cS, 1/n_1)$, the $\log$ covering number of $\cS$, recall $n_1$ is the size of $\cD_{\rm unlabel}$. Instantiating this general result, we derive score matching error for Gaussian case by upper bounding $\cN(\cS, 1/n_1)$ in this special case.

\begin{lemma}
\label{lmm:scr_mtc_err}
    Under Assumption \ref{assumption:subspace}, if $\nabla \log p_t(x \mid y) \in \cS$, where 
    \begin{align}
        \tag{\eqref{equ:function_class} revisited}
    \cS = \bigg\{\sbb_{V, \psi}(x, y, t) = \frac{1}{h(t)} (V \cdot \psi (V^\top x, y, t) - x) & :~ V \in \RR^{D \times d}, ~\psi \in \Psi:\RR^{d+1} \times [t_0, T] \to \RR^d~\bigg\},
    \end{align}
    with $\Psi$ parametric. Then for $\delta \geq 0$, with probability $1-\delta$, the square score matching error is bounded by $\epsilon^2_{diff} = \cO \left(\frac{1}{t_0}\sqrt{ \frac{\cN(\cS, 1/n_1) (d^2 \vee D) \log \frac{1}{\delta}} {n_1} } \right)$, i.e.,
\begin{equation}
\label{equ:score_error}
    \frac{1}{T - t_0} \int_{t_0}^T \EE_{(x_t, y) \sim P_t} \left[\norm{\nabla \log p_t(x_t | y) - \hat{s}(x_t, y, t)}_2^2\right] \diff t \leq \epsilon_{diff}^2,
\end{equation}
Proof of this lemma can be found in Appendix \ref{sec:score_matching_err}.
Recall $P_t$ comes from $P_{x\hat{y}}$ by noising $x$ at $t$ in the forward process. Under Assumption \ref{assumption:gaussian_design} and given $\hat f (x) = \hat{\theta}^{\top} x$ and $\hat{y} = \hat{f}(x)+\xi, \xi \sim {\sf N}(0, \nu^2)$, the score function $\nabla \log p_t(x \mid \hat y)$ to approximate is linear in $x$ and $\hat y$. When approximated by $S$ with $\Psi$ linear,
$\cN(\cS, 1/n_1) = \cO((d^2 + Dd)\log (D d n_1))$.
\end{lemma}

To provide fidelity and reward guarantees of $\hat{P}_a$: the generated distribution of $x$ given condition $\hat{y} = a$, we will need the following lemma. It provides a subspace recovery guarantee between $V$(score matching output) and $A$(ground truth), as well as a distance measure between distributions $P_a$ and $\hat{P}_a$, given score matching error $\epsilon_{diff}$.

Note $P_a$ and $\hat{P}_a$ are over $x$, which admits an underlying low-dimensional structure $x = Az$. Thus we measure distance between $P_a$ and $\hat{P}_a$ by defining 
\begin{definition}
\label{def:tv}
   We define $TV(\hat{P}_a):= \dtv \left(P^{LD}_{t_0} (z \mid \hat{f} (Az) = a), (U^{\top} V^{\top})_{\#}\hat{P}_a \right)$ with notations:
    \begin{itemize}
    \item $\dtv(\cdot, \cdot)$ is the TV distance between two distribution.
    \item $f_{\sharp} P$ denotes a push-forward measure, i.e., for any measurable $\Omega$, $(f_{\sharp}P)(\Omega) = P(f^{-1}(\Omega))$
    \item $(V^{\top})_{\#}\hat{P}_a$ pushes generated $\hat{P}_a$ forward to the low dimensional subspace using learned subspace matrix $V$. $U$ is an orthonormal matrix of dimension $d$.
    \item $P^{LD}_{t_0} (z \mid \hat{f} (Az) = a)$ is close to $(A^{\top})_{\#}P_a$, with $t_0$ taking account for the early stopping in backward process.
\end{itemize}
\end{definition}

We note that there is a distribution shift between the training and the generated data, which has a profound impact on the generative performance. We quantify the influence of distribution shift by the following class restricted divergence measure.

\begin{definition}
\label{def:dst_sft}
Distribution shift between two arbitrary distributions $P_1$ and $P_2$ restricted under function class $\cL$ is defined as 
\begin{align*}
\textstyle \cT(P_1, P_2; \cL) = \sup_{l \in \cL} \EE_{x \sim P_1}[l(x)] / \EE_{x \sim P_2}[l(x)] \quad \text{with arbitrary two distributions~} P_1, P_2.
\end{align*}
\end{definition}

Definition \ref{def:dst_sft} is well perceived in bandit and RL literature \citep{munos2008finite, liu2018breaking, chen2019information, fan2020theoretical}. 

\begin{lemma} 
\label{lmm:diff_results}
Given the square score matching error \eqref{equ:score_error} upper bounded by $\epsilon_{diff}^2$,
and when $P_z$ satisfying Assumption \ref{asmp:tail} with $c_0 I_d \preceq \EE_{z \sim P_z} \left[ z z^{\top}\right]$, it guarantees on for $x \sim \hat{P}_a$ and $\subangle{V}{A}:= \norm{VV^\top - AA^\top}^2_{\rm F}$ that
    \begin{align}
        \label{equ:orth_dtb} &(I_D - VV^{\top}) x \sim {\sf N}(0, \Lambda), \quad \Lambda \prec c t_0 I_D,\\
        \label{equ:subspace_rcv}&\subangle{V}{A} = \tilde{\cO}\left(\frac{t_0 }{c_0} \cdot \epsilon^2_{diff} \right).
    \end{align}
In addition,
    \begin{equation}
    \label{equ:tv_Pq}
        TV(\hat{P}_a) = \tilde{\cO}\left(\sqrt{\frac{\cT(P(x, \hat{y} = a), P_{x\hat{y}}; \bar{\cS})}{c_0}} \cdot \epsilon_{diff} \right).
    \end{equation}
\end{lemma}
with $\bar{\cS} = \left\{\frac{1}{T - t_0} \int_{t_0}^T \EE_{x_t \mid x} \norm{\nabla \log p_t(x_t \mid y) - s(x_t,  y, t)}_2^2 \diff t : s \in \cS\right\}$. $TV(\hat{P}_a)$ and $\cT(P(x, \hat{y} = a), P_{x\hat{y}}; \bar{\cS})$ are defined in Definition~\ref{def:tv} and~\ref{def:dst_sft}.
    Proof of Lemma \ref{lmm:diff_results} is in Appendix \ref{pf:diff_results}.

\subsection{Proof of Theorem~\ref{thm:fidelity} and Theorem \ref{thm:opr}}\label{pf:fidelity-opr}
\textbf{Proof of Theorem \ref{thm:fidelity}: bounding $\subangle{V}{A}$}. By Lemma 3 of \cite{chen2023score}, we have
\begin{equation*}
    \subangle{V}{A} = {\cO}\left(\frac{t_0 }{c_0} \cdot \epsilon^2_{diff} \right)
\end{equation*}
when the latent $z$ satisfying Assumption \ref{asmp:tail} and $c_0 I_d \preceq \EE_{z \sim P_z} \left[ z z^{\top}\right]$. Therefore, by \eqref{equ:score_error}, we have with high probability that
\begin{equation*}
    \subangle{V}{A} = \tilde{\cO} \left( \frac{1}{c_0} \sqrt{\frac{\cN(\cS, 1/n_1) (D\vee d^2)} {n_1}} \right).
\end{equation*}
When Assumption \ref{assumption:gaussian_design} holds, plugging in $c_0 = \lambda_{\min}$ and $\cN(\cS, 1/n_1) = \cO((d^2 + Dd)\log (D d n_1))$, it gives
\begin{equation*}
    \subangle{V}{A} = \tilde{\cO} \left( \frac{1}{\lambda_{\min}} \sqrt{\frac{(D \vee d^2)d^2 + (D \vee d^2)Dd} {n_1}} \right),
\end{equation*}
where $\tilde \cO$ hides logarithmic terms. When $D > d^2$, which is often the case in practical applications, we have
\begin{align*}
\subangle{V}{A} = \tilde{\cO}\left(\frac{1}{\lambda_{\min}} \sqrt{\frac{Dd^2 + D^2d} {n_1}} \right).
\end{align*}

\noindent \textbf{Proof of Theorem \ref{thm:opr}: bounding $\EE_{x \sim \hat{P}_a} [\|\ox\|_2]$}. 
By the definition of $\ox$ that $\ox = (I_D - AA^{\top})x$,
\begin{equation*}
        \EE_{x \sim \hat{P}_a} [\|x^{\perp}\|_2] = \EE_{x \sim \hat{P}_a} [\|(I_D - AA^{\top})x\|_2] \leq \sqrt{\EE_{x \sim \hat{P}_a} [\|(I_D - AA^{\top})x\|_2^2]}.
    \end{equation*}   
Score matching returns $V$ as an approximation of $A$, then
\begin{align*}
    \|(I_D - AA^{\top})x\|_2 &\leq \|(I_D - VV^{\top})x\|_2 + \|(VV^{\top} - AA^{\top})x\|_2,\\
    \EE_{x \sim \hat{P}_a} [\|(I_D - AA^{\top})x\|^2_2] &\leq 2 \EE_{x \sim \hat{P}_a} [\|(I_D - VV^{\top})x\|^2_2] + 2 \EE_{x \sim \hat{P}_a} [\|(VV^{\top} - AA^{\top})x\|^2_2],
\end{align*}
where by \eqref{equ:orth_dtb} in Lemma \ref{lmm:diff_results} we have
\begin{equation*}
    (I_D - VV^{\top})x \sim \sf N(0, \Lambda), \quad \Lambda \prec c t_0 I
\end{equation*}
for some constant $c \geq 0$. Thus 
\begin{equation}
\label{equ:bkw_Gaussian_l2}
    \EE_{x \sim \hat{P}_a} \left[ \|(I_D - VV^{\top})x\|^2_2 \right] = \operatorname{Tr}(\Lambda) \leq c t_0 D.
\end{equation}
On the other hand,
\begin{equation*}
    \|(VV^{\top} - AA^{\top})x\|^2_2 \leq \|VV^{\top} - AA^{\top}\|^2_{op} \|x\|^2_2 \leq \|VV^{\top} - AA^{\top}\|^2_{F} \|x\|^2_2,
\end{equation*}
where $\|VV^{\top} - AA^{\top}\|^2_{F}$ has an upper bound as in \eqref{equ:subspace_rcv} and $\EE_{x \sim \hat{P}_a} \left[ \|x\|^2_2 \right] )$ is bounded in Lemma \ref{lmm:exp_x_norm} by
\begin{equation*}
    \EE_{x \sim \hat{P}_a} \left[ \|x\|^2_2 \right] = \cO \left( ct_0D + M(a) \cdot  (1+ TV(\hat{P}_a) \right).
\end{equation*}
with $M(a) = O \left(  \frac{a^2}{\| {\beta}^* \|_{\Sigma}} + d \right)$.

Therefore, to combine things together, we have
\begin{align*}
     \EE_{x \sim \hat{P}_a} [\|x^{\perp}\|_2] &\leq \sqrt{2 \EE_{x \sim \hat{P}_a} [\|(I_D - VV^{\top})x\|^2_2] + 2 \EE_{x \sim \hat{P}_a} [\|(VV^{\top} - AA^{\top})x\|^2_2]} \\
    &\leq c^{\prime} \sqrt{t_0 D} + 2 \sqrt{\subangle{V}{A}}\cdot \sqrt{\EE_{x \sim \hat{P}_a} \left[ \|x\|^2_2 \right]}\\
    &= \cO \left(\sqrt{t_0 D} + \sqrt{\subangle{V}{A}}\cdot \sqrt{M(a)}\right) .
\end{align*}
$\cO$ hides multiplicative constant and $\sqrt{\subangle{V}{A} t_0D}$, $\sqrt{\subangle{V}{A} M(a) TV(\hat{P}_a)}$, which are terms with higher power of $n_1^{-1}$ than the leading term.

\subsection{Proof of Theorem~\ref{thm:parametric} and \ref{thm:rlhf-parametric}}\label{pf:parametric}
Proof of Theorem~\ref{thm:parametric} and \ref{thm:rlhf-parametric}  are provided in this section. This section breaks down into three parts: \textbf{ Suboptimality Decomposition}, \textbf{bounding $\cE_1$ Relating to Offline Bandits for different datasets }, \textbf{Bounding $\cE_2$ and the Distribution Shift in Diffusion}.

\subsubsection{$\subopt(\hat{P}_a ; y^* = a)$ Decomposition.}
\label{sec:dcp}
    Recall notations $\hat{P}_a:= \hat{P}(\cdot | \hat{y} = a)$ (generated distribution) and $P_a:= P(\cdot | \hat{y} = a)$ (target distribution) and $f^*(x) = g^*(\px) + h^*(\ox)$.
$\EE_{x \sim \hat{P}_a} [f^{\star}(x)]$ can be decomposed into 3 terms:
\begin{align*}
    \EE_{x \sim \hat{P}_a} [f^{\star}(x)] \geq&  \EE_{x \sim P_a} [f^*(x)] - \left|\EE_{x \sim \hat{P}_a} [f^*(x)] - \EE_{x \sim P_a} [f^*(x)]\right|\\
    \geq & \EE_{x \sim P_a} [\hat{f}(x)] -  \EE_{x \sim P_a} \left[ \left|\hat{f}(x) - f^*(x)\right| \right] - \left|\EE_{x \sim \hat{P}_a} [f^*(x)] - \EE_{x \sim P_a} [f^*(x)]\right|\\
    \geq& \EE_{x \sim P_a} [\hat{f}(x)] - \underbrace{\EE_{x \sim P_a} \left[ \left|\hat{f}(x) - g^*(x)\right| \right]}_{\cE_1}- \underbrace{\left| \EE_{x \sim P_a} [g^*(\px)] - \EE_{x \sim \hat{P}_a} [g^*(\px)]\right|}_{\cE_2} -  \underbrace{\EE_{x \sim \hat{P}_a} [h^*(\ox)]}_{\cE_3} ,
\end{align*}
where $\EE_{x \sim P_a} [\hat{f}(x)] = \EE_{a \sim q} [a]$ and we use $x = \px$, $f^*(x) = g^*(x)$ when $x \sim P_a$.
Therefore
\begin{align*}
    \subopt(\hat{P}_a ; y^* = a) &= a - \EE_{x \sim \hat{P}_a} [f^{\star}(x)] \\
    &\leq \underbrace{\EE_{x \sim P_a} \left[ \left|(\hat{\theta} - \theta^*)^{\top} x\right| \right]}_{\cE_1} + \underbrace{\left| \EE_{x \sim P_a} [g^*(\px)] - \EE_{x \sim \hat{P}_a} [g^*(\px)]\right|}_{\cE_2}  +  \underbrace{\EE_{x \sim \hat{P}_a} [h^*(\ox)]}_{\cE_3}.
\end{align*}

$\cE_1$ comes from regression: prediction/generalization error onto $P_a$, which is independent from any error of distribution estimation that occurs in diffusion. $\cE_2$ and $\cE_3$ do not measure regression-predicted $\hat{f}$, thus they are independent from the prediction error in $\hat{f}$ for pseudo-labeling. $\cE_2$ measures the disparity between $\hat{P}_a$ and $P_a$ on the subspace support and $\cE_3$ measures the off-subspace component in generated $\hat{P}_a$.

\subsubsection{Bounding $\cE_1$ under the Labeled Dataset Setup.}
For all $x_i \in \cD_{\rm label}, y_i = f^*(x_i) + \epsilon_i = g(x_i) + \epsilon_i$. Thus, trained on $\cD_{\rm label}$ the prediction model $\hat{f}$ is essentially approximating $g$. By estimating $\theta^*$ with ridge regression on $\cD_{\rm label}$, we have $\hat f (x)= \hat{\theta}^{\top} x$ with
\begin{equation}
     \hat{\theta} = \left(X^{\top} X+\lambda I\right)^{-1} X^{\top}\left(X \theta^*+\eta\right),
\end{equation}
where $X^{\top} = (x_1, \cdots, x_i, \cdots, x_{n_2})$ and $\eta = (\epsilon_1, \cdots, \epsilon_i, \cdots, \epsilon_{n_2})$.

\begin{lemma}
\label{lmm:reg_err}
Under Assumption \ref{assumption:subspace} and \ref{assumption:linear_reward} and given $\epsilon_i \sim \sf N (0, \sigma^2)$, define $V_{\lambda}:= X^{\top} X+\lambda I$, $\hat{\Sigma}_{\lambda}:= \frac{1}{n_2} V_{\lambda} $ and $\Sigma_{P_a}:= \EE_{x \sim P_a} x x^{\top}$ the covariance matrix (uncentered) of $P_a$, and take $\lambda = 1$, then with high probability
\begin{equation}
    \cE_1 \leq \sqrt{\operatorname{Tr}( \hat{\Sigma}_{\lambda}^{-1} \Sigma_{P_a})} \cdot \frac{\cO \left( \sqrt{d \log n_2 } \right)}{\sqrt{n_2}}.
\end{equation}
\end{lemma}
Proof of Lemma \ref{lmm:reg_err} is in Appendix \ref{pf:reg_err}.

\begin{lemma}
\label{lmm:distribution_sft}
    Under Assumption \ref{assumption:subspace}, \ref{assumption:linear_reward} and \ref{assumption:gaussian_design}, when $\lambda = 1$, $P_a$ has a shift from the empirical marginal of $x$ in dataset by
    \begin{equation}
        \operatorname{Tr}( \hat{\Sigma}_{\lambda}^{-1} \Sigma_{P_a}) \leq \cO \left( \frac{ a^2 }{ \| \beta^* \|_{\Sigma}} + d \right).
    \end{equation}
    when $n_2 = \Omega ( \max \{\frac{1}{\lambda_{\min}}, \frac{d}{\|\beta^*\|^2_{\Sigma}} \})$.
\end{lemma}

Proof of Lemma \ref{lmm:distribution_sft} is in Appendix \ref{pf:distribution_sft}.

\subsubsection{Bounding $\cE_1$ under the Human Preference Setup}\label{proof:hf}
\re{Recall that in the human preference setup, we use the  MLE to learn the underlying parameter $\hat{\theta}$. For all $x_i^{(j)} \in \dlabel$, $j\in\{1,2\}$, we have $ f(x_i^{(j)}) = g(x_i^{(j)})$. Therefore, by Assumption \ref{assumption:linear_reward}, we have the following objective: \begin{align}\label{eq:loss-func}
    \hat{\theta} =  \operatorname{argmin}_{\theta \in \Theta} -\frac{1}{n} \sum_{i=1}^n \bigg\{ u_i^\top \theta - \log\bigg(\exp( \theta^\top x_i^{(1)} ) + \exp(\theta^\top x_i^{(2)})\bigg) \bigg\},
\end{align}
We are now ready to present our results.
\begin{lemma}\label{thm:rew-hf-est}
    With probability at least $1-\delta$, we have \begin{align*}
        \cE_1 \leq O\bigg(\sqrt{\kappa^2\cdot \frac{d+\log1/\delta}{n}+ \lambda} \cdot \sqrt{ \operatorname{Tr}(\EE_{P_a}[xx^\top] (\tilde{\Sigma_\lambda})^{-1} )}\bigg)
    \end{align*}
    here $\kappa = \big(1+\exp(\sqrt{\log n / \delta})\big)^2$.
\end{lemma}
    Proof of Lemma \ref{thm:rew-hf-est} is in Appendix \ref{pf:mle}.}
\subsubsection{Bounding $\cE_2$ and the Distribution Shift in Diffusion}

\begin{lemma}
\label{lmm:E2}
Under Assumption \ref{assumption:subspace}, \ref{assumption:linear_reward} and 
\ref{assumption:gaussian_design}, when $t_0 = \left((Dd^2 + D^2d) / n_1\right)^{1/6}$
\begin{align*}
    \cE_2 = \tilde \cO\left(\sqrt{\frac{\cT(P(x, \hat{y} = a), P_{x\hat{y}}; \bar{\cS})}{\lambda_{\min}}} \cdot  \left(\frac{Dd^2 + D^2d} {n_1} \right)^{\frac{1}{6}} \cdot a \right).
\end{align*}
\end{lemma}

Proof of Lemma \ref{lmm:E2} can be found in Appendix \ref{pf:E2}.

Note that $\cT(P(x, \hat{y} = a), P_{x\hat{y}}; \bar{\cS})$ depends on $a$ and measures the distribution shift between the desired distribution $P(x, \hat{y} = a)$ and the data distribution $P_{x \hat{y}}$. To understand this distribution's dependency on $a$, it what follows we give $\cT(P(x, \hat{y} = a), P_{x\hat{y}}; \bar{\cS})$ a  shorthand as $\dshift^2(a)$ and give it an upper bound in one special case of the problem.

\paragraph{Distribution Shift} In the special case of covariance $\Sigma$ of $z$ is known and $\norm{A - V}_2^2 = \cO\left(\norm{AA^\top - VV^\top}_{\rm F}^2\right)$, we showcase a bound on the distribution shift in $\cE_2$, as promised in the discussion following Theorem~\ref{thm:parametric}. We have
\begin{align*}
\dshift^2(a) = \frac{\EE_{P_{x , \hat{y} = a}} [\ell(x, y; \hat{s})]}{\EE_{P_{x\hat{y}}} [\ell(x, y; \hat{s})]},
\end{align*}
where $\ell(x, y; \hat{s}) = \frac{1}{T-t_0} \int_{t_0}^T \EE_{x' | x} \norm{\nabla_{x'} \log \phi_t(x' | x) - \hat{s}(x', y, t)}_2^2 \diff t$. By Proposition~\ref{prop:equivalent_score_matching}, it suffices to bound
\begin{align*}
\dshift^2(a) = \frac{\EE_{P_{x , \hat{y} = a}} [\int_{t_0}^T \norm{\nabla \log p_t(x, y) - \hat{s}(x, y, t)}_2^2 \diff t]}{\EE_{P_{x\hat{y}}} [\int_{t_0}^T \norm{\nabla \log p_t(x, y) - \hat{s}(x, y, t)}_2^2 \diff t]}.
\end{align*}
We expand the difference $\norm{\nabla \log p_t(x, y) - \hat{s}(x, y, t)}_2^2$ by
\begin{align*}
\norm{\nabla \log p_t(x, y) - \hat{s}(x, y, t)}_2^2 & \leq \frac{2}{h^2(t)} \Big[\norm{(A - V)B_t (A^\top x + \nu^{-2} y\theta)}_2^2 + \norm{VB_t (A - V)^\top x}_2^2\Big] \\
& \leq \frac{2}{h^2(t)} \Big[\norm{A - V}_2^2 \norm{B_t (A^\top x + \nu^{-2} y\theta)}_2^2 + \norm{A - V}_2^2 \norm{x}_2^2\Big] \\
& \leq \frac{2}{h^2(t)} \norm{A - V}_2^2 (3\norm{x}_2^2 + y^2),
\end{align*}
where we recall $B_t$ is defined in \eqref{eq:gaussian_score} and in the last inequality, we use $(a + b)^2 \leq 2a^2 + 2b^2$. In the case of covariance matrix $\Sigma$ is known, i.e., $B_t$ is known, we also consider matrix $V$ directly matches $A$ without rotation. Then by \citet[Lemma 3 and 17]{chen2023score}, we have $\norm{A - V}_2^2 = \cO\left(\norm{AA^\top - VV^\top}_{\rm F}^2\right) = \cO\left(t_0/c_0\EE_{P_{x\hat{y}}}[\ell(x, y; \hat{s})]\right)$. To this end, we only need to find $\EE_{P_{x | \hat{y} = a}}[\norm{x}_2^2]$. Since we consider on-support $x$, which can be represented as $x = Az$, we have $\norm{x}_2 = \norm{z}_2$. Thus, we only need to find the conditional distribution of $z | \hat{y} = a$. Fortunately, we know $(z, \hat{y})$ is jointly Gaussian, with mean $0$ and covariance
\begin{align*}
\begin{bmatrix}
\Sigma & \Sigma \hat{\beta} \\
\hat{\beta}^\top \Sigma & \hat{\beta}^\top \Sigma \hat{\beta} + \nu^2
\end{bmatrix}.
\end{align*}
Consequently, the conditional distribution of $z | \hat{y} = a$ is still Gaussian, with mean $\Sigma\hat{\beta} a / (\hat{\beta}^\top \Sigma \hat{\beta} + \nu^2)$ and covariance $\Sigma - \Sigma \hat{\beta} \hat{\beta}^\top \Sigma / (\hat{\beta}^\top \Sigma \hat{\beta} + \nu^2)$. Hence, we have
\begin{align*}
\EE_{P_{z | \hat{y} = a}}[\norm{z}_2^2] = \frac{1}{(\hat{\beta}^\top \Sigma \hat{\beta} + \nu^2)^2} \left((a^2-\hat{\beta}^\top \Sigma \hat{\beta} - \nu^2) \hat{\beta}^\top \Sigma^2 \hat{\beta}\right) + {\rm Tr}(\Sigma) = \cO\left(a^2 \vee d \right).
\end{align*}
We integrate over $t$ for the numerator in $\dshift(a)$ to obtain $\EE_{P_{x | \hat{y} = a}} [\int_{t_0}^T \norm{\nabla \log p_t(x, y) - \hat{s}(x, y, t)}_2^2 \diff t = \cO\left((a^2 \vee d) \frac{1}{c_0} \EE_{P_{x\hat{y}}}[\ell(x, y; \hat{s})]\right)$. Note the cancellation between the numerator and denominator, we conclude
\begin{align*}
\dshift(a) = \cO\left(\frac{1}{c_0} (a \vee \sqrt{d}) \right).
\end{align*}
As $d$ is a natural upper bound of $\sqrt{d}$ and viewing $c_0$ as a constant, we have $\dshift(a) = \cO\left(a \vee d \right)$ as desired.

%% file: appendix/experiment.tex
\section{Additional Experimental Results}\label{appendix:exp}
\subsection{Simulation}
We generate the latent sample $z$ from standard normal distribution $z\sim\sf{N}(0, I_d)$ and set $x=Az$ for a randomly generated orthonormal matrix $A\in\mathbb{R}^{D\times d}$. The dimensions are set to be $d=16, D=64$. The reward function is set to be $f(x)=(\theta^*)^\top x_\parallel - 5\Vert x_\perp\Vert^2_2$, where $\theta^*$ is defined by $A\beta^*$. We generate $\beta^*$ by uniformly sampling from the unit sphere. 

When estimating $\hat{\theta}$, we set $\lambda=1.0$. The score matching network is based on the UNet implementation from \url{https://github.com/lucidrains/denoising-diffusion-pytorch}, where we modified the class embedding so it accepts continuous input. The predictor is trained using $8192$ samples and the score function is trained using $65536$ samples. When training the score function, we choose Adam as the optimizer with learning rate $8\times 10^{-5}$. We train the score function for $10$ epochs, each epoch doing a full iteration over the whole training dataset with batch size $32$. 

For evaluation, the statistics is computed using $2048$ samples generated from the diffusion model. The curve in the figures is computed by averaging over $5$ runs. 

\subsection{Directed Text-to-Image Generation}
\begin{figure}[htb!]
    \centering
    \begin{subfigure}[h]{0.24\textwidth}
        \includegraphics[width = \textwidth]{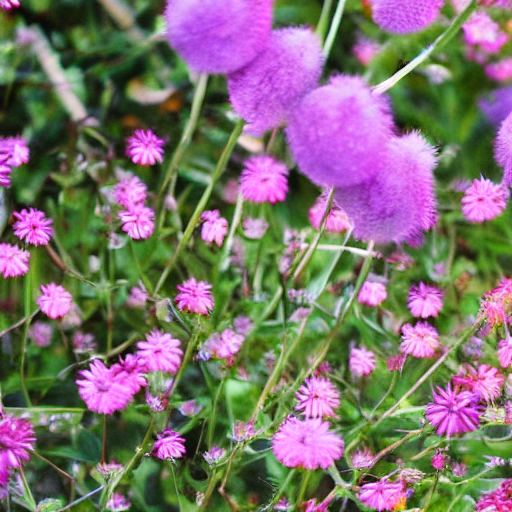}
        \caption{A positive sample}
        \label{fig:a}
    \end{subfigure} 
    \begin{subfigure}[h]{0.24\textwidth}
        \includegraphics[width = \textwidth]{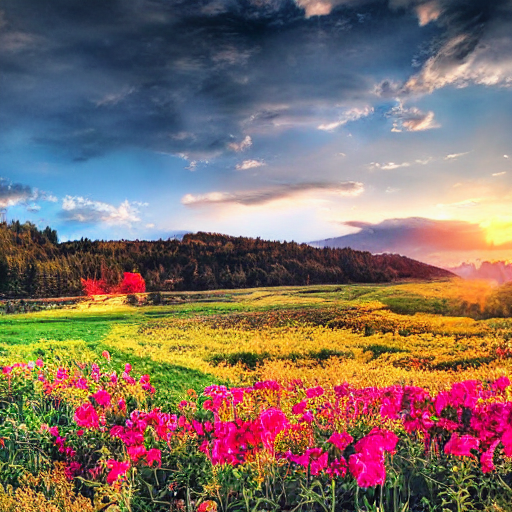}
        \caption{A positive sample}
        \label{fig:b}
    \end{subfigure}
    \hspace{1pt}
    \begin{subfigure}[h]{0.24\textwidth}
        \includegraphics[width = \textwidth]{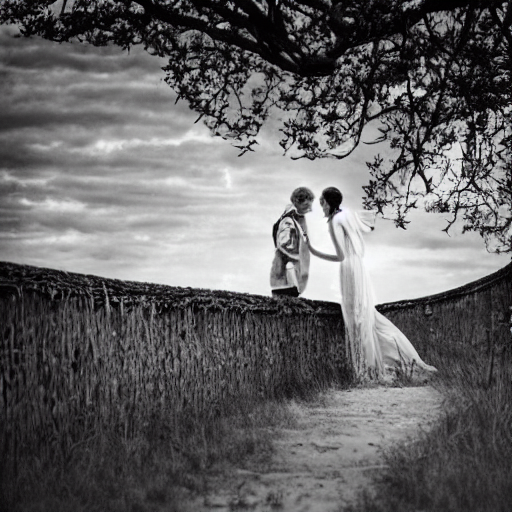}
        \caption{A negative sample}
        \label{fig:c}
    \end{subfigure} 
    \begin{subfigure}[h]{0.24\textwidth}
        \includegraphics[width = \textwidth]{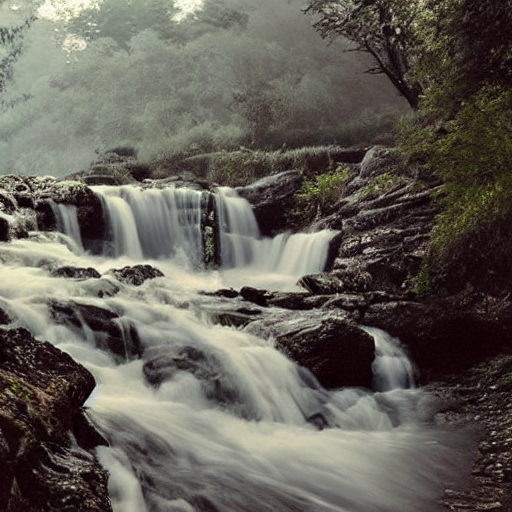}
        \caption{A negative sample}
        \label{fig:d}
    \end{subfigure}
    \caption{Random samples with high rewards and low rewards. }
\label{fig:s}
\end{figure}

\textbf{Samples of high rewards and low rewards from the ground-truth reward model.} In Section~\ref{sec:experiment:2}, the ground-truth reward model is built by replacing the final prediction layer of the ImageNet pre-trained ResNet-18 model with a randomly initialized linear layer of scalar outputs. To investigate the meaning of this randomly-generated reward model, we generate images using Stable Diffusion and filter out images with rewards $\geq 0.4$ (positive samples) and rewards $\leq -0.4$ (negative samples) and pick two typical images for each; see Figure~\ref{fig:s}.  We note that in real-world use cases, the ground-truth rewards are often measured and annotated by human labors according to the demands.

\textbf{Training Details.} In our implementation, as the Stable Diffusion model operates on the latent space of its VAE, we build a 3-layer ConvNet with residual connections and batch normalizations on top of the VAE latent space. We train the network using Adam optimizer with learning rate $0.001$ for 100 epochs.

%% file: appendix/nonparametric.tex
\newpage